\documentclass{article}

\usepackage[square,sort,comma,numbers]{natbib}
\usepackage[final]{nips_2017}
\usepackage{color}
\usepackage{defs_151115}
\usepackage[utf8]{inputenc} % allow utf-8 input
\usepackage[T1]{fontenc}    % use 8-bit T1 fonts
\usepackage{hyperref}       % hyperlinks
\usepackage{url}            % simple URL typesetting
\usepackage{booktabs}       % professional-quality tables
\usepackage{amsfonts}       % blackboard math symbols
\usepackage{nicefrac}       % compact symbols for 1/2, etc.
\usepackage{microtype}      % microtypography
\usepackage{amsmath}
\usepackage{amsthm}
\usepackage{amssymb}
\usepackage{float}
\usepackage{bm}
\usepackage{wrapfig}
\usepackage{subcaption}
\usepackage{adjustbox}
\usepackage{enumitem}
\usepackage{lipsum}
\usepackage{multirow}
\usepackage{comment}
\usepackage{paralist}
\usepackage{graphicx}
\usepackage[normalem]{ulem}
\graphicspath{ {./figs/} }

\theoremstyle{definition}

\newcommand{\mysubfigurecaption}[1]{{\begin{scriptsize}#1\end{scriptsize}}}

\title{An empirical analysis of the optimization of deep network loss surfaces}

\author{
Daniel Jiwoong Im$^{1,2}$\thanks{Work done during an internship at Janelia Research Campus} \\
  $^{1}$AIFounded Inc.\\
  \texttt{daniel.im@aifounded.com} \\
  \And
  Michael Tao\\
  University of Toronto\\
  \texttt{mtao@dgp.toronto.edu} \\
  \AND
  Kristin Branson \\
  $^{2}$Janelia Research Campus, HHMI\\
  \texttt{bransonk@janelia.hhmi.org} \\
}

\begin{document}

\maketitle

\begin{abstract}
    The success of deep neural networks hinges on our ability to accurately and efficiently optimize high-dimensional, non-convex functions. In this paper, we empirically investigate the loss functions of state-of-the-art networks, and how commonly-used stochastic gradient descent variants optimize these loss functions. To do this, we visualize the loss function by projecting them down to low-dimensional spaces chosen based on the convergence points of different optimization algorithms. Our observations suggest that optimization algorithms encounter and choose different descent directions at many saddle points to find different final weights. Based on consistency we observe across re-runs of the same stochastic optimization algorithm, we hypothesize that each optimization algorithm makes characteristic choices at these saddle points. 
\end{abstract}

\section{Introduction}
\label{sec:intro}

%Our main results are build upon empirically investigating high dimensional
%non-convex loss functions such as deep neural networks.
%Our experimental outcomes provide rich inituition on properties of 
%different optimization methods and the structure of the loss surface.

Deep neural networks are trained by optimizing extremely high-dimensional loss functions with respect to the networks' weights. These loss functions measure the error of the network's predictions based on these weights compared to training data. These loss functions are non-convex and are known to have many local minima. They are usually minimized using first-order gradient descent
%~\citep{Robbins1951, polyak1964} 
algorithms such as stochastic gradient descent (SGD)~\citep{Bottou1991}. The success of deep learning critically depends on how well we can minimize these loss functions, both in terms of the quality of the convergence points and the time it takes to find them. Understanding the geometry of these loss functions and how optimization algorithms traverse them is thus of vital importance. 

%\mtao{This doesn't make sense with the exacerbated
%Of course the ability to find those minima is naught if the computational time required to find those local minima is prohibitively large.
%This issue is exacerbated as the size of modern datasets become larger and the tasks at hand become increasingly sophisticated.
%Intuition on the behavior of loss functions can also allow for practitioners to expediate the tedious training component of building networks.
%It is for these reasons that we posit that analysis of loss functions is of utmost value.

%There are a few fundamental facts about these loss functions that let us know that training neutral networks do not fit in the classical category of being an easy problem.
%Not only are these functions of a high dimension and non-convex; they're highly nonlinear.
%Even evaluating these functions is computationally burdensome so one must apply stochastic methods to decrease the evaluation cost.
%A more direct and applied sort of analysis is of course to look at the behavior of different optimization procedures is to discern the advantages and disadvantages of each method.
%The most basic, important, and yet ellusive questions are certainly ``which one's the fastest?'' and ``how can I make them faster?''.

Several works have theoretically analyzed and characterized the shape of deep network loss functions. However, to make these analyses tractible, they have relied on simplifying assumptions. Some have characterized the critical points of deep {\em linear} neural networks~\citep{Baldi1989a,Saxe2014} and high-dimensional, random Gaussian error functions~\citep{Parisi2007, Fyodorov2007, Bray2007,Dauphin2014}. Others~\citep{Choromanska2015a,Choromanska2015b,Kawaguchi2016} characterize the critical points of fully-connected, nonlinear deep networks, but make simplifying assumptions about the distributions and independence of various variables in the network. These works have used a variety of techniques to conclude that, given some simplifying assumptions, the loss functions of deep network  have no or few bad local minima, but may instead have many saddle points. Many of these works have used empirical analyses to show that some properties of the simplified and real networks are similar. 

If bad local minima are indeed rare, as suggested by theoretical analyses as well as the practical success of deep learning, then optimization algorithms do not need to take precautions to avoid them. Instead, they must only bypass saddle points and find \textit{any} local minimum quickly. The speed of gradient-descent-based algorithms are generally measured on strictly convex, in particular, quadratic functions~\citep{polyak1964, Broyden1970, nesterov1983, Martens2010, Erdogdu2015}. If different algorithms converge to the same local minimum from a common initialization, then such analyses might directly apply to the performance of these algorithms on deep networks. Conversely, if different optimization algorithms converge to different local minima that have different characteristics, then it may be necessary to evaluate the algorithms' performance on real loss functions. 

In this work, we empirically investigated the geometry of the real loss functions for state-of-the-art networks and data sets, and how commonly used optimization algorithms interact with these real loss surfaces. To our knowledge, this is the first work to jointly analyze deep network loss functions and optimization algorithms. To do this, we extended the methodology of Goodfellow et al.~\citep{Goodfellow2015}, and examined the loss function in a low-dimensional, projected space chosen to investigate properties of and the relationship between the convergence points of the different optimization algorithms. Our empirical results support the following novel conclusions:
\begin{compactitem}
\item Different optimization algorithms find different solutions\footnote{When we say solutions we specifically mean the result of running a reasonably-parameterized optimization method for a reasonable number of iterations}%local minima 
%\mtao{converge to diffrent points?} kb: when we plot beyond alpha = 1, we see that they are local minima within the projected space
    within the projected space. This is true even when starting from the same initialization with the same mini-batch and dropout settings. Most surprisingly, this remained true when we switched from one optimization algorithm to another after the training error has nearly plateaued, suggesting that there are a plethora of saddle points even near the convergence points.
\item Despite corresponding to different final points, the loss surfaces for the same algorithm from different initializations are remarkably consistent and characteristic of the optimization algorithm. The shapes of the loss functions near the final solution differ across algorithms, and we trace this back to different algorithms selecting weight vectors with a consistent norm. Switching from one optimization algorithm to another late in training results in a final point characteristic of the second optimization algorithm. 
% \mtao{shape -> character because things like distance are not``shape'' properties}
% kb: we show that shape is consistent across reruns, but we don't actually show distance traveled for different runs, we just show distance traveled when switching. tried to make this explicit.
\item Batch normalization is key to obtaining this consistency in the projected loss surface. Without it, we see much more variability across re-runs from different initializations.
\end{compactitem}

\section{Experimental setup}
\label{sec:setup}

\subsection{Network architectures and data sets}
\label{sec:setup:networksanddatasets}

We conducted experiments on three different neural network architectures.
They are all high-dimensional, deep networks and are currently used in many machine vision and learning tasks.
Most importantly, their loss functions are highly non-convex.

The Network-in-Network (NIN)~\citep{Lin2014} and Visual Geometry Group (VGG) network~\citep{Simonyan2015} are feed-forward convolutional networks developed for image classification, and have excellent performance on the Imagenet~\citep{Russakovsky2014} and CIFAR10~\citep{Krizhevsky2009} data sets. 
%The Long Short-Term Memory (LSTM) network ~\citep{Hochreiter1997} is a recurrent neural network that has been successful in tasks that take variable-length sequences as input and/or produce variable-length sequences as output, such as speech recognition~\citep{Graves2013} and image caption generation~\citep{Vinyals2015}. 
Finally, we tested a two-layer fully-connected neural network (FC2). 

In our experiments, we tested NIN and VGG on the CIFAR10 image classification data set. We tested FC2 on the MNIST digit recognition task~\citep{MNIST}. %We tested the LSTM on the Penn Treebank (PTB) next-word prediction data set~\cite{PennTreebank}. 
% \mtao{Isn't this info mostly reflected in the previous paragraph?}
% kb: edited, now networks in first paragraph, data sets in second. 

Details of the network parameters and training data sets can be found in Section~\ref{sec:supp:networkanddatadetails}. 

\subsection{Optimization methods}
\label{sec:setup:optimization}

We analyzed the performance of five stochastic gradient-descent optimization methods commonly used for training deep neural networks: (vanilla) Stochastic Gradient Descent (SGD)~\citep{Robbins1951}, Stochastic Gradient Descent with Momentum (SGDM), RMSprop~\citep{Tieleman2012}, Adadelta~\citep{Zeiler2011}, and Adam~\citep{Kingma2015}. These are all first-order gradient descent algorithms that estimate the gradients based on randomly-grouped minibatches of training examples. One of the major differences between these algorithms is the step-sizes chosen for each iteration. SGD and SGDM utilize fixed step-sizes, while RMSprop, Adadelta, and Adam use adaptive step-sizes based off of previous iterations. Details are provided in Section~\ref{sec:background}.

%Runge Kutta integerators were commonly used to solve 
%determininstic system. Nonetheless, there exist
%generalization of Runge Kutta methods for stochastic 
%system as well \citep{Kloeden1992}.

%The probabilistic numerical methods are numerical 
%methods that are applied to solving integrals and 
%oridinary differential equations that 
%takes in inputs from probability distribution
%\citep{Schober2014a, Schober2014b, Mahsereci2016}.
%There has been little work on probablistic numerical
%methods, hence, this area is considered to be open 
%domain of research.

In addition to these five existing optimization methods, we compared to a new gradient descent method we developed based on the family of Runge-Kutta integrators~\citep{Butcher1963}. In our experiments, we tested a second-order Runge-Kutta integrator in combination with SGD (RK2) and in combination with Adam (Adam\&RK2). Details are provided in Section~\ref{sec:rk}).

%\begin{table*}[htp]
%    \includegraphics[width=\linewidth]{AlgorithmMinimumComparisonTable_v2.pdf}
%    \caption{Visualization of the loss surface near and between local minima found by different optimization methods.
%    Each box corresponds to a pair of optimization methods. In the lower triangle, we plot the projection of the loss surface at weight vectors between the initial weight and the learned weights found by the two optimization methods. Color as well as height of the surface indicate the loss function value. In the upper triangle, we plot the functional difference between the network corresponding to the learned weights for the first algorithm and networks corresponding to weights linearly interpolated between the first and second algorithm's learned weights.
%    (Best viewed in zoom)}
%    \label{fig:loss_surface_zoo}
%\end{table*}

\subsection{Analysis methods}
\label{sec:setup:methods}

Several of our empirical analyses are based on the technique of Goodfellow et al.~\citep{Goodfellow2015}, in which properties of the loss function are examined by projecting it onto a single, carefully chosen dimension. The projection is chosen based on important weight configurations and they plot the value of the loss function along the line segment between two weight configurations. They perform two such analyses: one in which they interpolate between the initialization weights and the final learned weights, and one in which they interpolate between two sets of final weights\footnote{Batch normalization parameters beta and gamma (from \citep{Ioffe2015}) are interpolated, and batch mean and standard deivation were calculated using training data.}, each learned from different initializations. Based on their observations, they conclude that local minima do not need to be overcome by SGDM, which is in agreement with previous work suggesting there are few bad local minima. 

%\begin{figure}[htb]
\begin{wrapfigure}{r}{0.5\textwidth}
   \vspace{-.5cm}
  \centering
  \begin{minipage}{0.2455\textwidth}
    \centering
    \includegraphics[width=\textwidth]{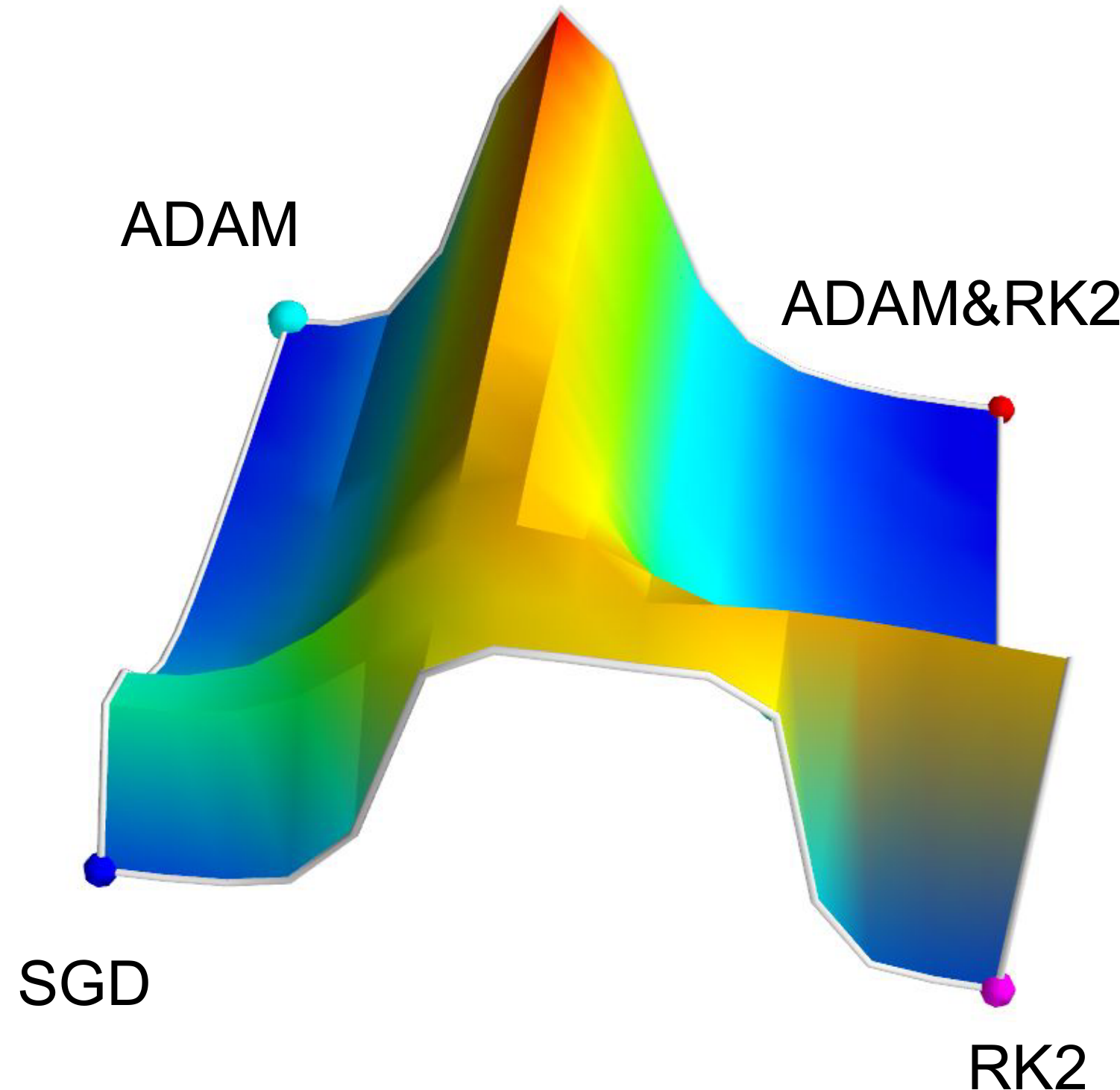}
    \vspace{-.2cm}
    \subcaption{NIN, CIFAR10}
    %\label{fig:nin_inter_render_f2f}
  \end{minipage}
  \hspace{.2cm}
  \begin{minipage}{0.2155\textwidth}
    \centering
    \includegraphics[width=\textwidth]{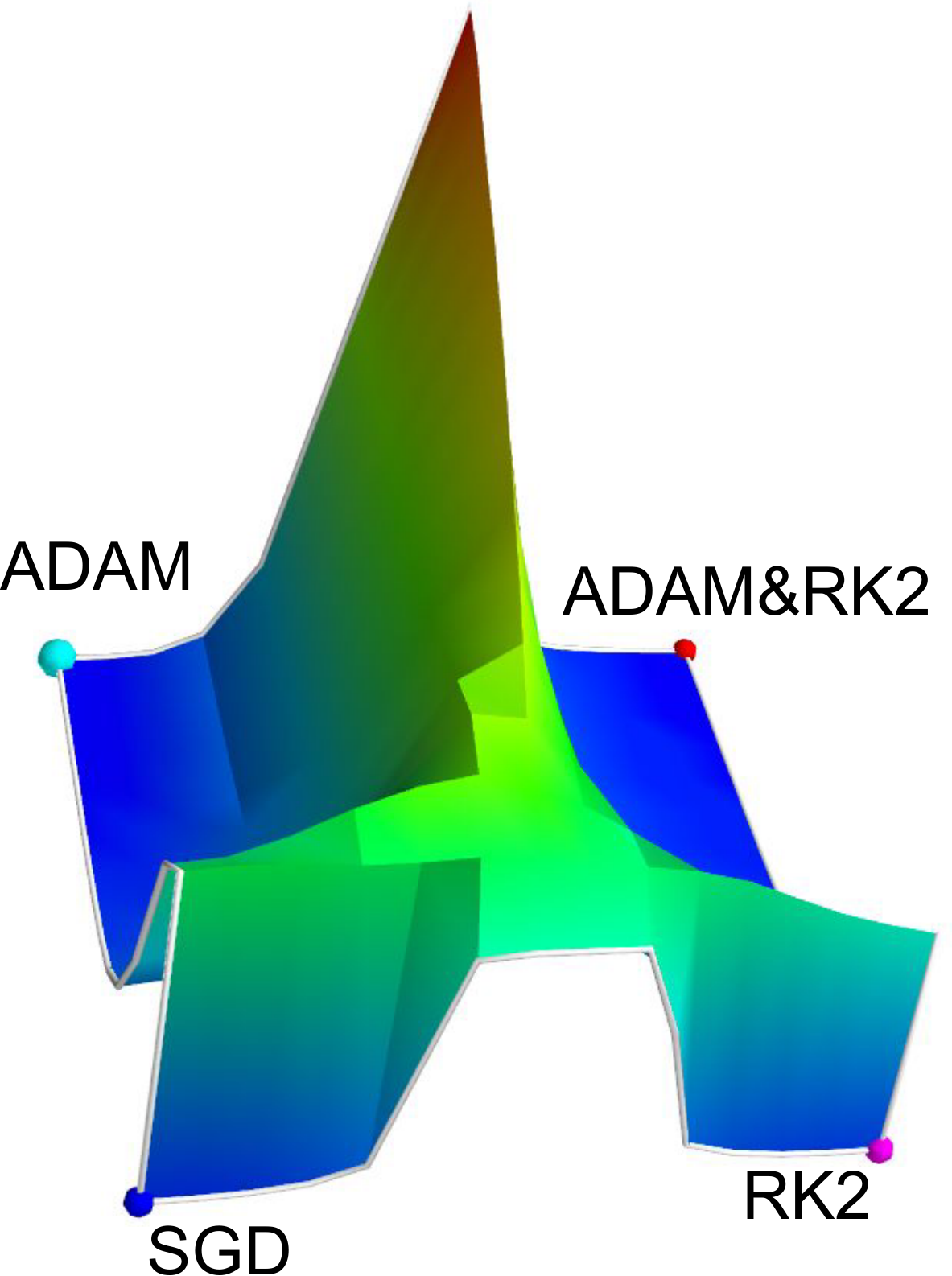}
    \vspace{-.2cm}
    \subcaption{VGG, CIFAR10}
    %\label{fig:vgg_inter_render_f2f}
  \end{minipage}
  \caption{Visualization of the loss surface at weights interpolated between the final points of four different algorithms from the same initialization. \label{fig:quad_render}}
    \vspace{-.4cm}
\end{wrapfigure}
%\end{figure}

\begin{figure*}[t]
\centering
\begin{minipage}{.23\textwidth}
\centering
\includegraphics[width=\linewidth]{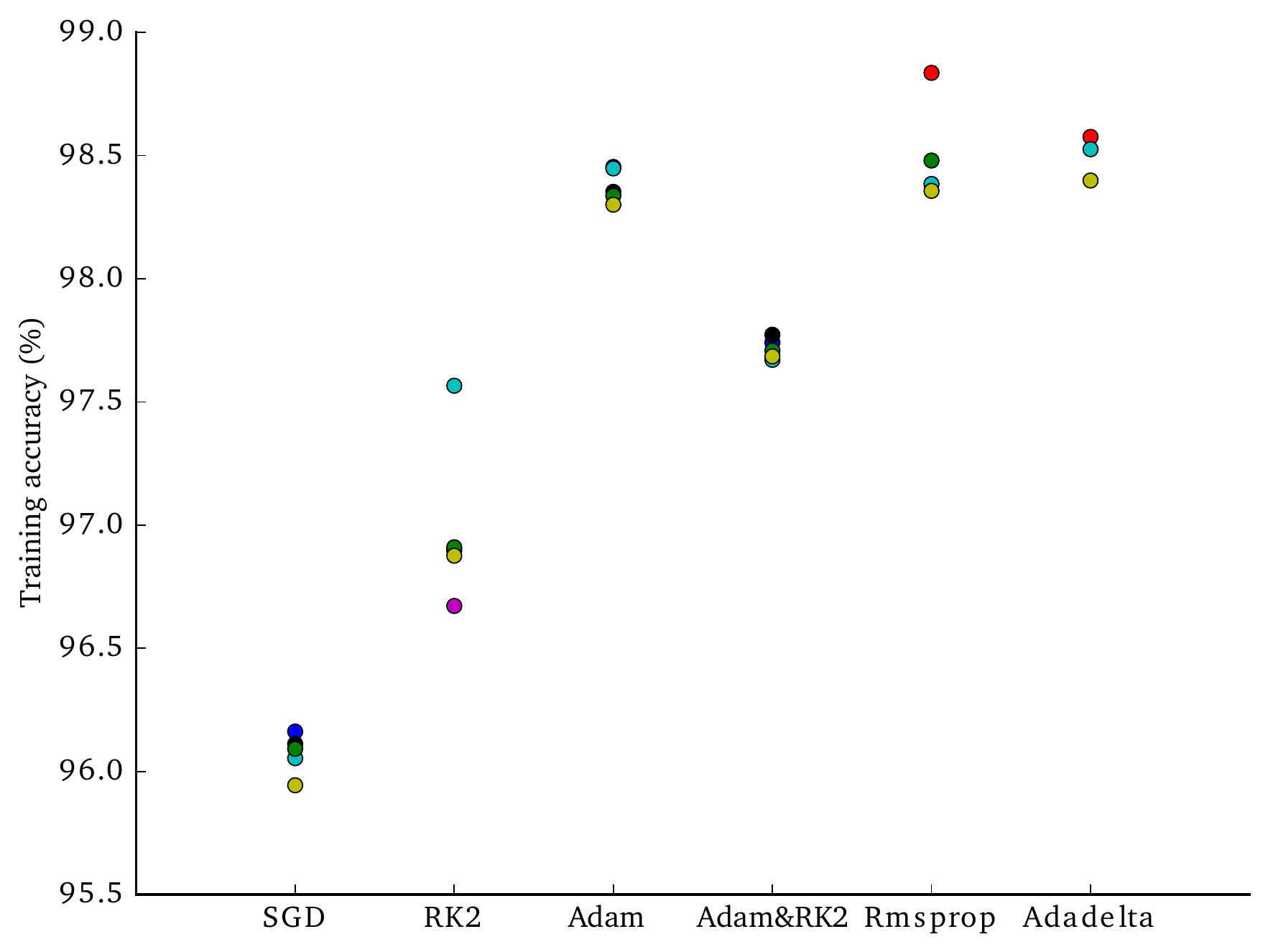}\\
\includegraphics[width=\linewidth]{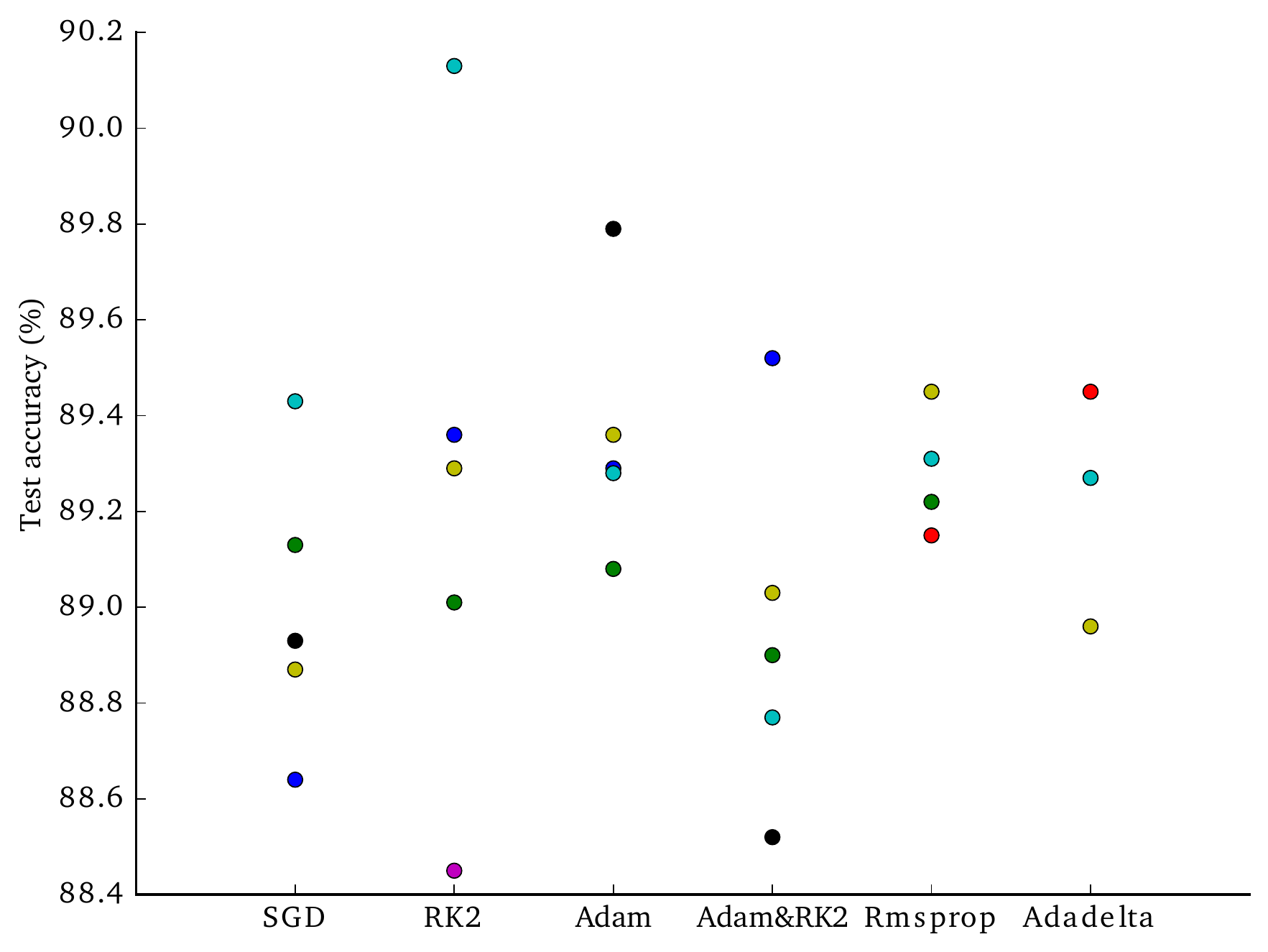}\\
\vspace{-.2cm}
\mysubfigurecaption{(a) VGG, CIFAR10}\\
\vspace{.2cm}
\includegraphics[width=\linewidth]{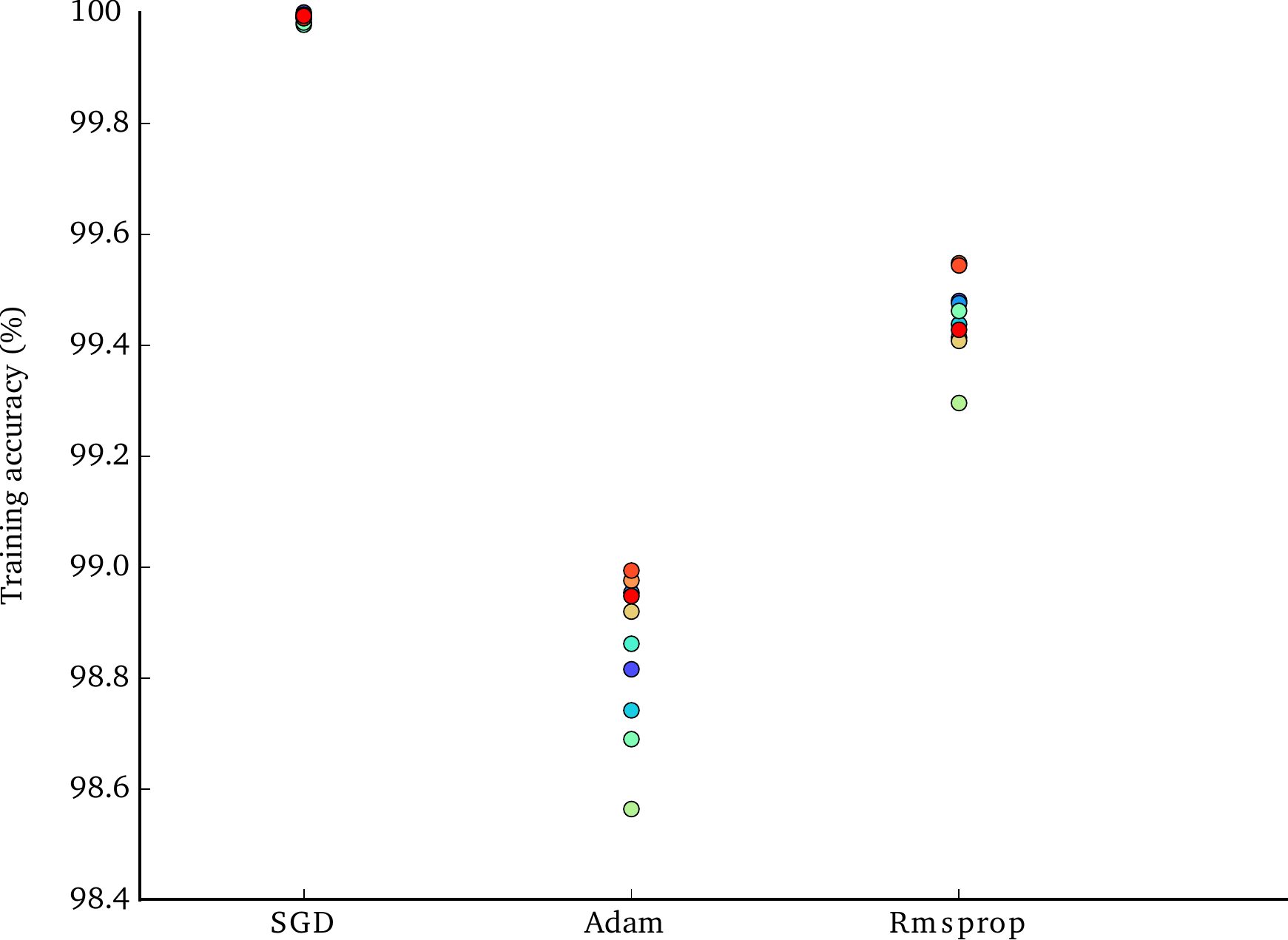}\\
\includegraphics[width=\linewidth]{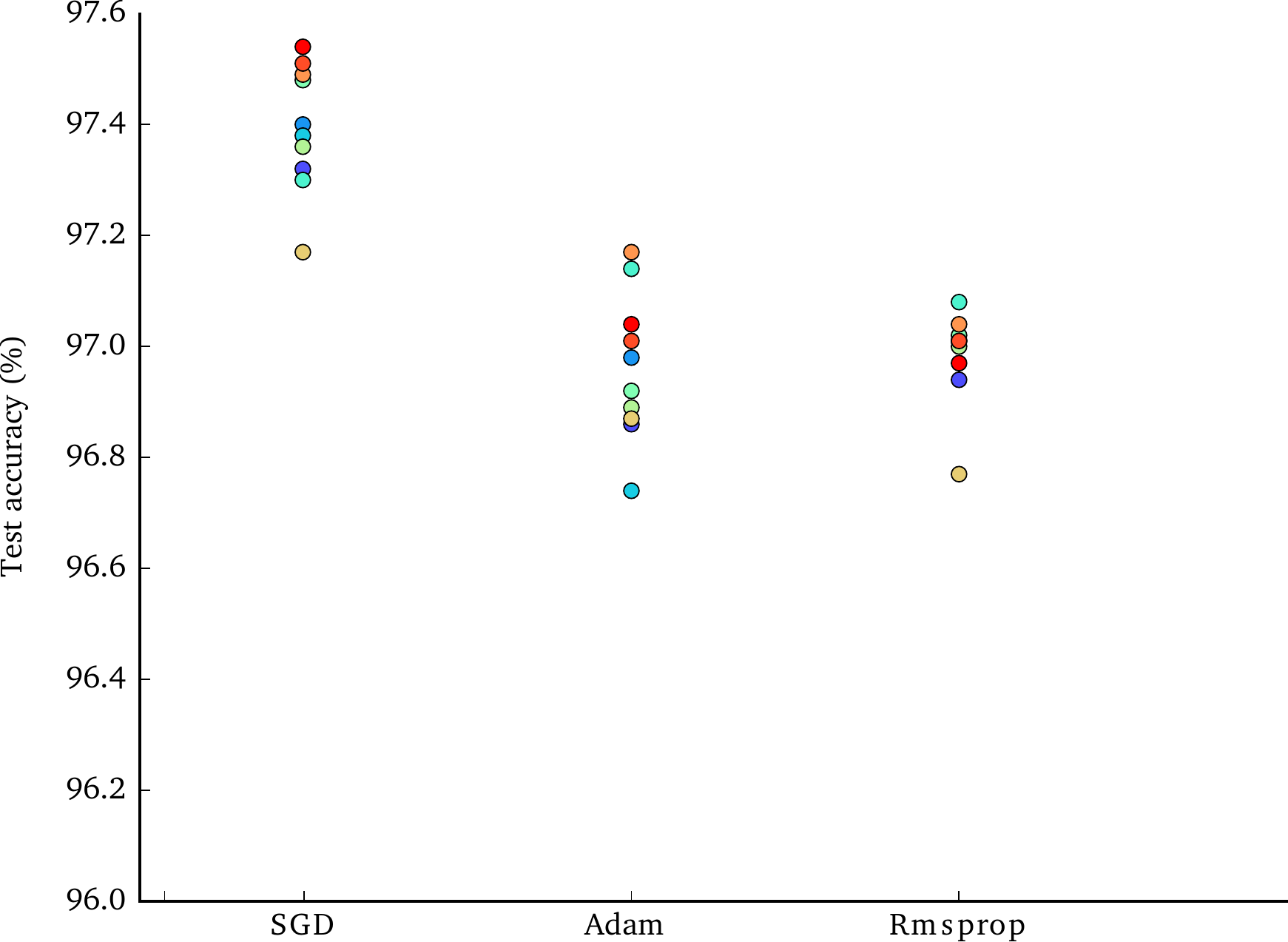}\\
\vspace{-.2cm}
\mysubfigurecaption{(b) FC2, MNIST}
\end{minipage}
\begin{minipage}{.73\textwidth}
\centering
\includegraphics[width=\linewidth]{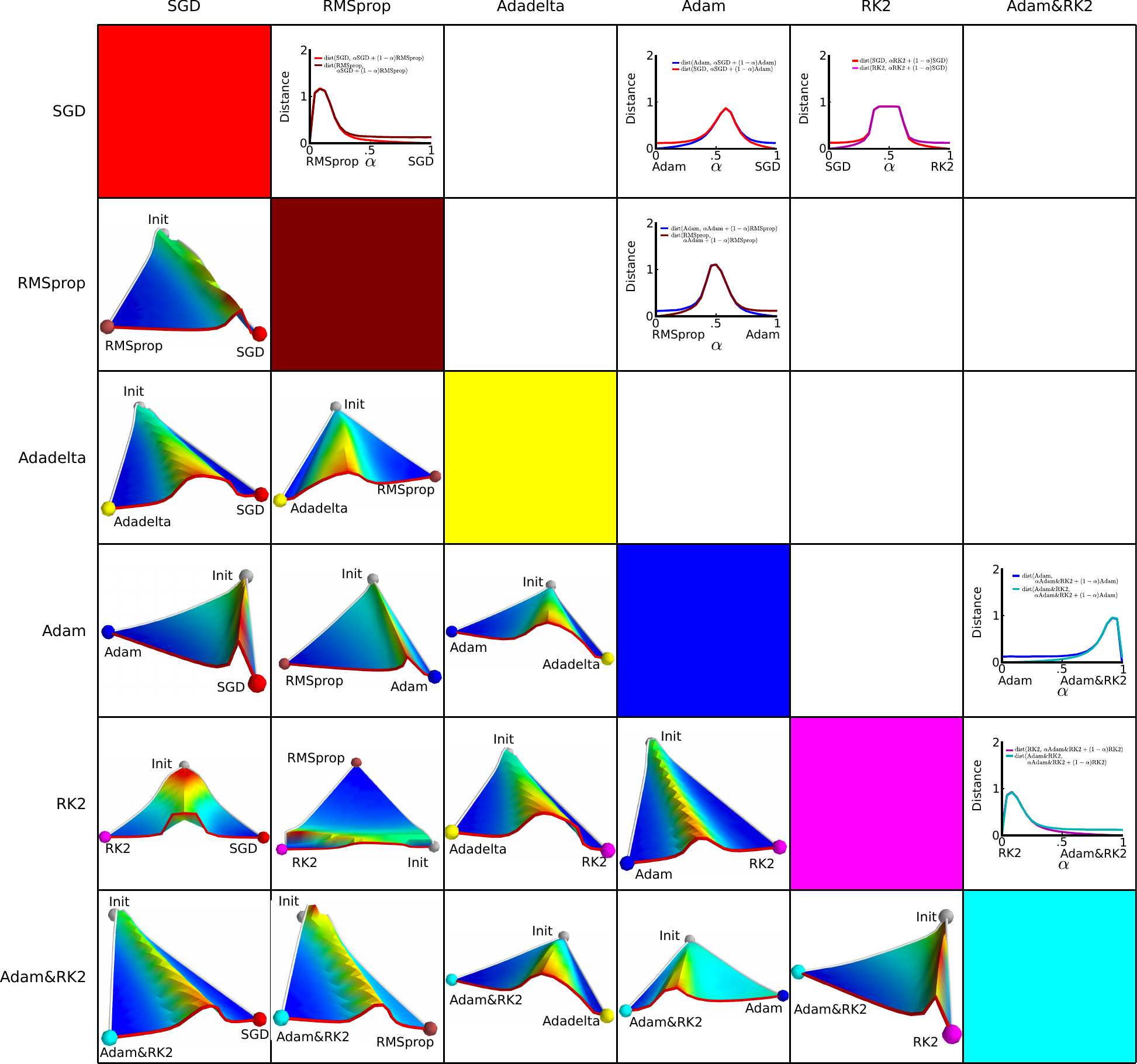}
\mysubfigurecaption{(c) VGG, CIFAR10}
\end{minipage}
\caption{(a-b) Training and test accuracy for each of the optimization methods for (a) the VGG network on CIFAR10 and (b) the FC2 network on MNIST. Colors correspond to different initializations. (c) Visualization of the loss surface near and between the final points found by different optimization methods for the VGG network on CIFAR10. Each box corresponds to a pair of optimization methods. In the lower triangle, we plot the projection of the loss surface at weight vectors between the initial weight and the learned weights found by the two optimization methods. Color as well as height of the surface indicate the loss function value. In the upper triangle, we plot the functional difference between the network corresponding to the learned weights for the first algorithm and networks corresponding to weights linearly interpolated between the first and second algorithm's learned weights.
\label{fig:MinimumComparison}}
\vspace{-.3cm}
\end{figure*}

In this work, we use a similar visualization technique, but choose different low-dimensional subspaces for the projection of the loss function. These subspaces are based on the initial weights as well as the final weights learned using the different optimization algorithms. They are chosen to answer a variety of questions about the loss function and how the different optimization algorithms interact with this loss function. In contrast, Goodfellow et al. only looked at SGDM. In addition, we explore the use of two-dimensional projections of the loss function, allowing us to better visualize the space between final parameter configurations.%\footnote{The final parameter configuration found by training neural networks are called {\em solutions} instead of local minima, since they are not verified to be a local minima even though they appear to be local minima in slice of low-dimensional space.}%local minima. 
We do this via barycentric and bilinar interpolation for triplets and quartets of points, respectively (details in Section~\ref{sec:3D_vis}). See Fig.~\ref{fig:quad_render} and Appendix Fig.~\ref{fig:MinimumComparison}(c) for examples. 

We examine the loss surfaces around the final weight configurations learned by these variants of SGD. For all networks, we train until convergence, when fluctuations in training accuracy based on dropout are much larger than the trending increase over epochs.
The total numbers of epochs of training for each experiment are shown in Table~\ref{table:architectur_table}. 

These final points are local minima in the projected space (Fig.~\ref{fig:basinsize}(a,c)). If a weight vector is a local minimum in the intelligently-chosen projected space, it is suggestive, but {\em not} conclusive, that it is a local minimum in the original high-dimensional space. Hence, we will use the term solutions.

\section{Experimental results}
\label{sec:ExperimentalResults}

\subsection{Different optimization methods find different solutions}
\label{sec:DifferentAlgsDifferentMinima}

We trained the neural networks described in Section~\ref{sec:setup:networksanddatasets} using each optimization method starting from the same initial weights and with the same minibatching. As shown in Fig.~\ref{fig:MinimumComparison}(a-b) for the VGG network, the quality of these final points were quite similar in terms of both training and test error. These results agree with previous work suggesting there are few bad solutions. %local minima.}

To investigate the shape of the loss function, we computed the value of the loss function for weight vectors interpolated between the initial weights, the final weights for one algorithm, and the final weights for a second algorithm for each pairing of algorithms (Fig.~\ref{fig:MinimumComparison}(c)) for the VGG network. For every pair of optimization algorithms, we observe that, within the projected space, the final points are always separated by a high-loss region. With the caveat that we have only visualized a 2-dimensional projection of the loss surface, this suggests that each optimization algorithm found a solution within the basin of a different local minimum (in sliced low-dimensional space), despite starting at the same initialization. We observed similar phenomena for NIN on CIFAR10 and FC2 on MNIST (Fig.~\ref{fig:interpolations}). We investigated the space between other triplets and quadruplets of weight vectors (Fig.~\ref{fig:quad_render}), and even in these projections of the loss function, we still saw that the local minima returned by different algorithms are separated by high loss weight parameters. 

Our observation that different optimization algorithms find very different solutions suggests that they choose different descent directions at saddle points encountered during optimization~\citep{Dauphin2014}. We next investigated whether these saddle points only occur in the early, {\em transient}~\citep{Sutskever2013} phase of optimization in which the loss is decreasing rapidly, or also occur in the later {\em minimization}~\citep{Sutskever2013} phase in which the loss decreases slowly. We investigated the effects of switching from one type of optimization method to another 25\%, 50\%, and 75\% of the way through training. We emphasize that we are not switching methods to improve performance, but rather to investigate the shape of the loss function during the minimization phase of optimization. 

\begin{figure*}[t]
  \centering
  % ADAM -> SGD
  \begin{minipage}{.495\linewidth}
    \begin{center}
      \includegraphics[width=0.505\linewidth]{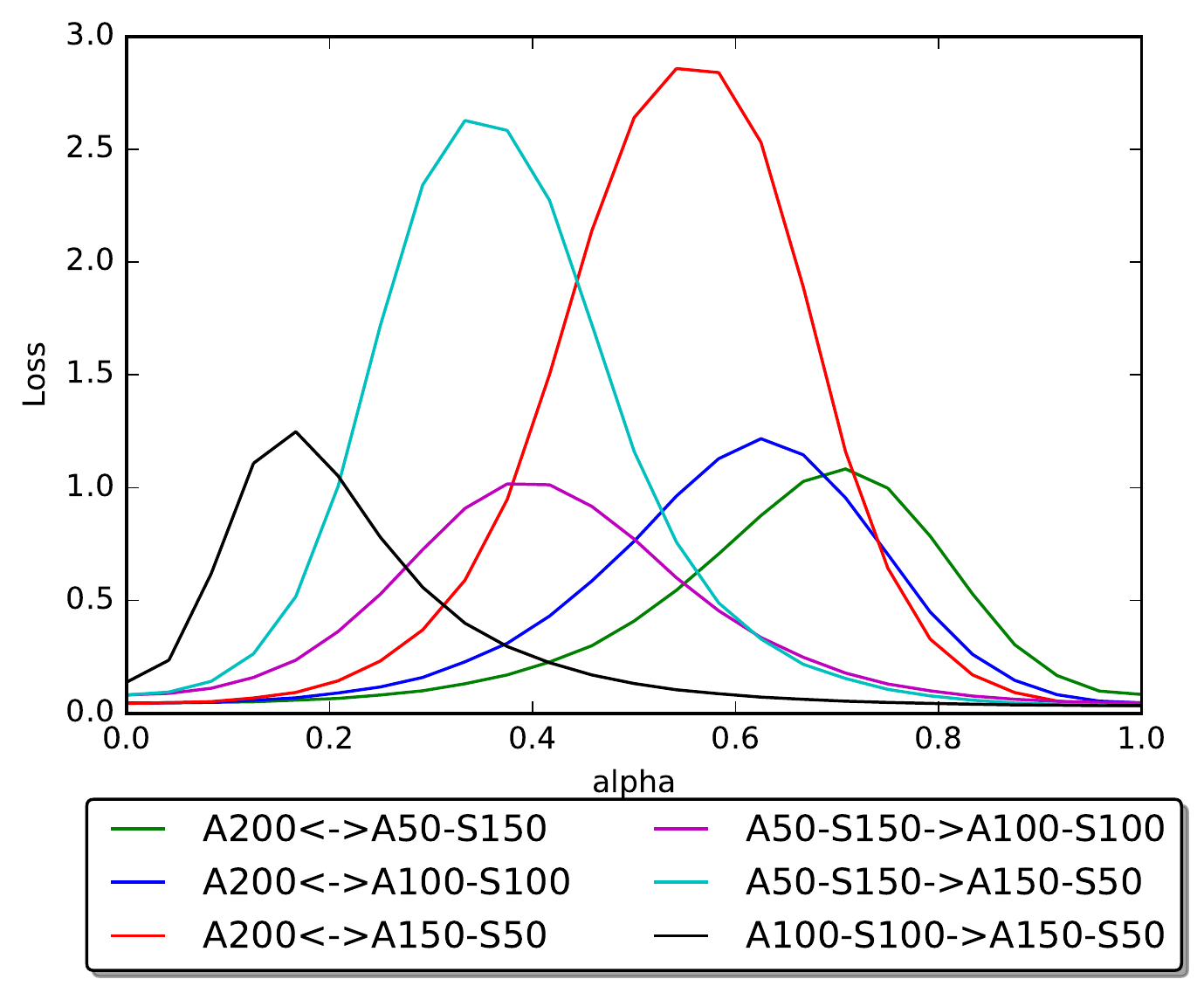}
      \includegraphics[width=0.475\linewidth]{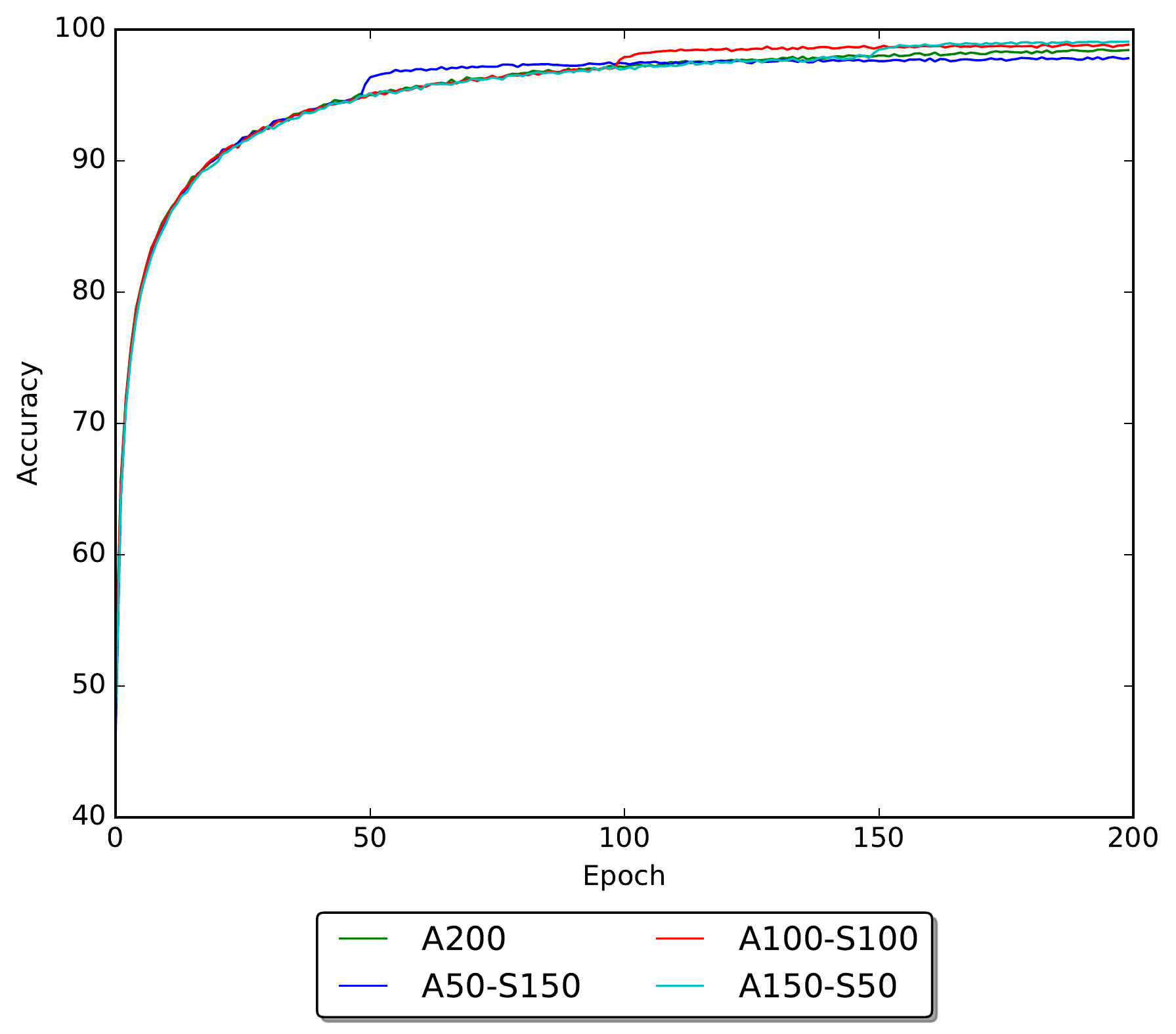}\\
      \subcaption{NIN: Switching from Adam (A, learning rate $\eta = .001$) to SGD (S, $\eta = .01$).}
      % fig:nin_switch_small_adam-sgd
    \end{center}
  \end{minipage}
  \begin{minipage}{.495\linewidth}
    \begin{center}
      \includegraphics[width=0.505\linewidth]{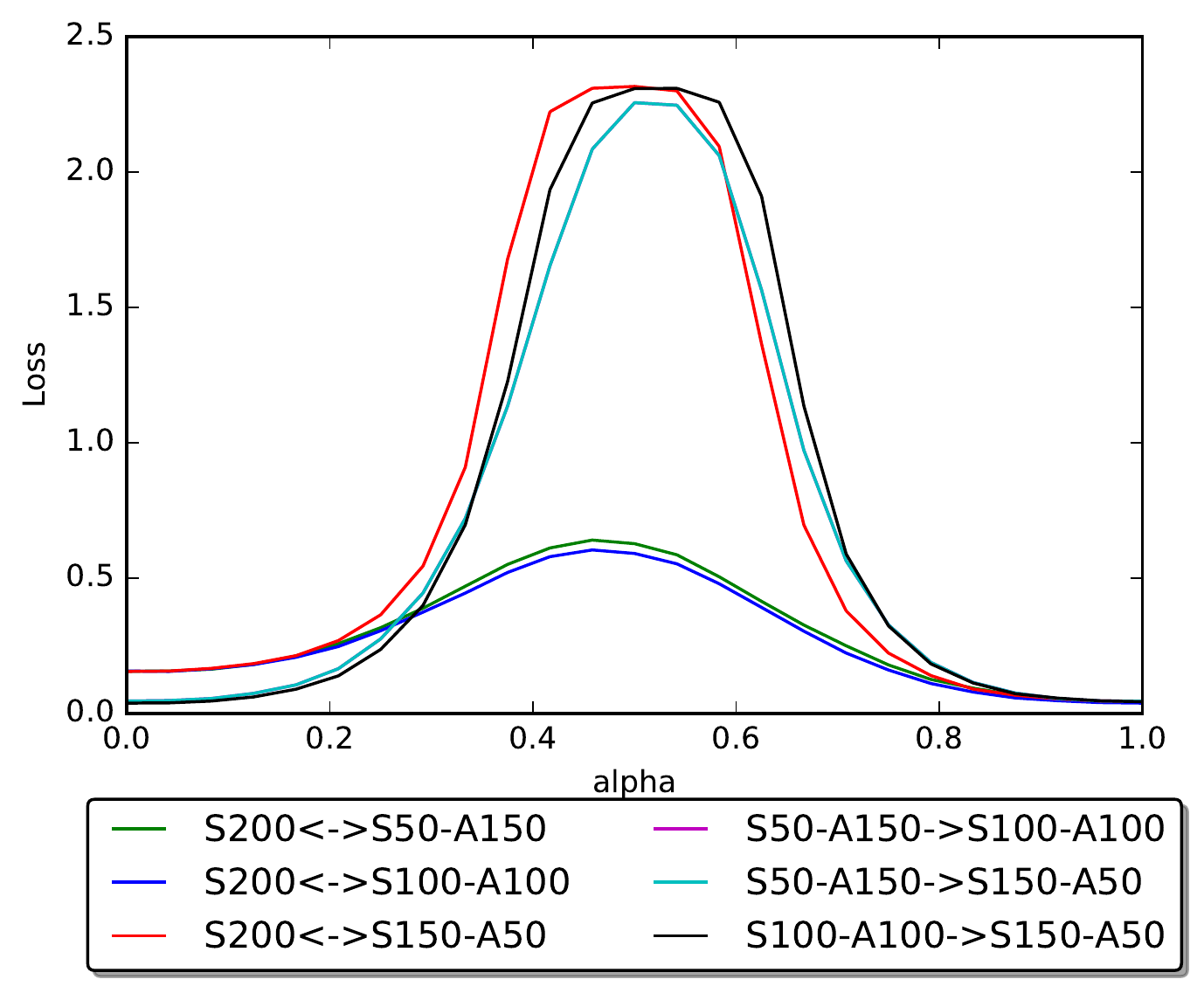}
      \includegraphics[width=0.475\linewidth]{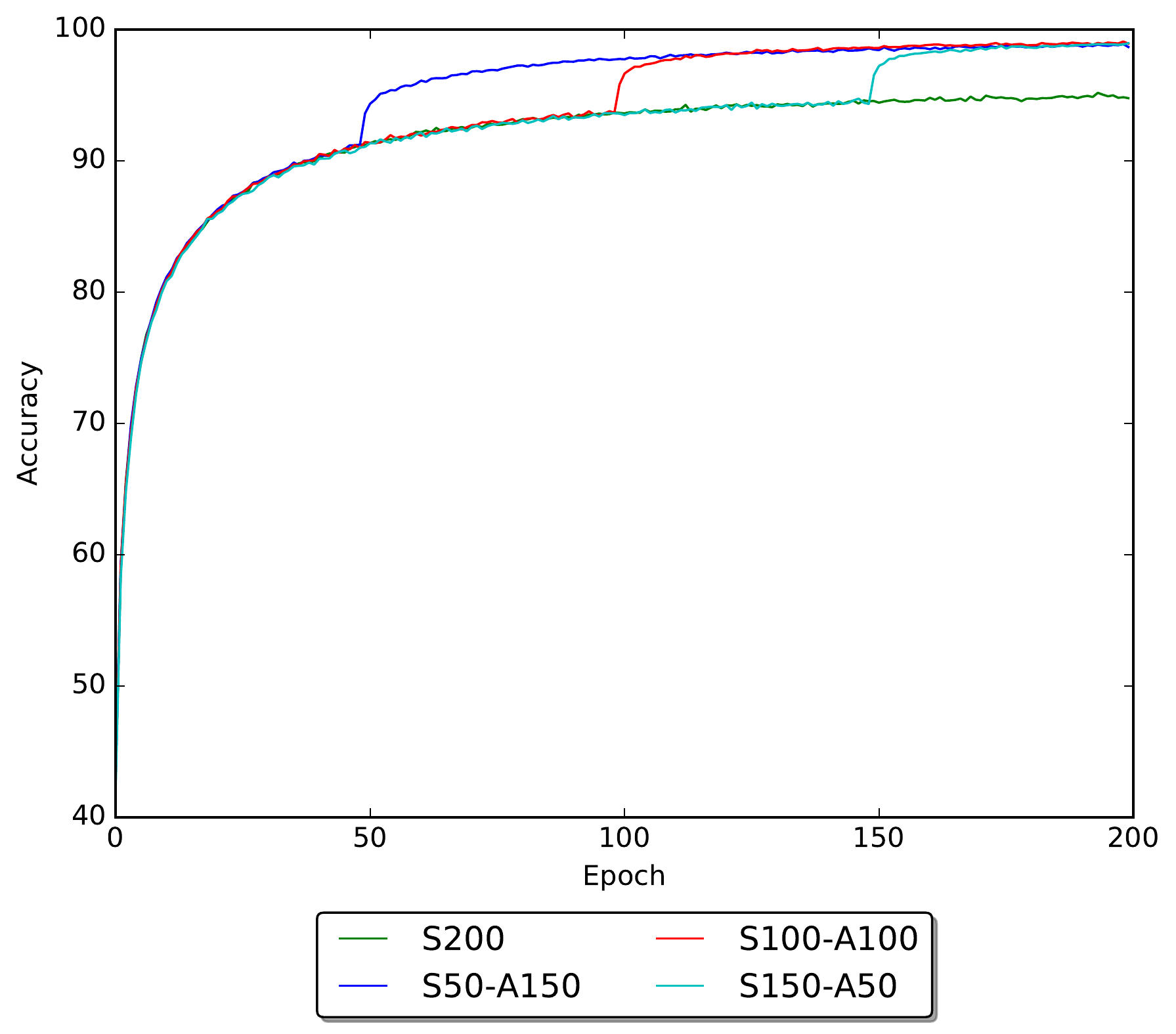}
      \subcaption{NIN: Switching from SGD (S, $\eta = .1$) to Adam (A, $\eta = .0001$).}
      % fig:nin_switch_small
    \end{center}
  \end{minipage}\\
  \begin{minipage}{.495\linewidth}
    \begin{center}
      \includegraphics[width=0.525\linewidth]{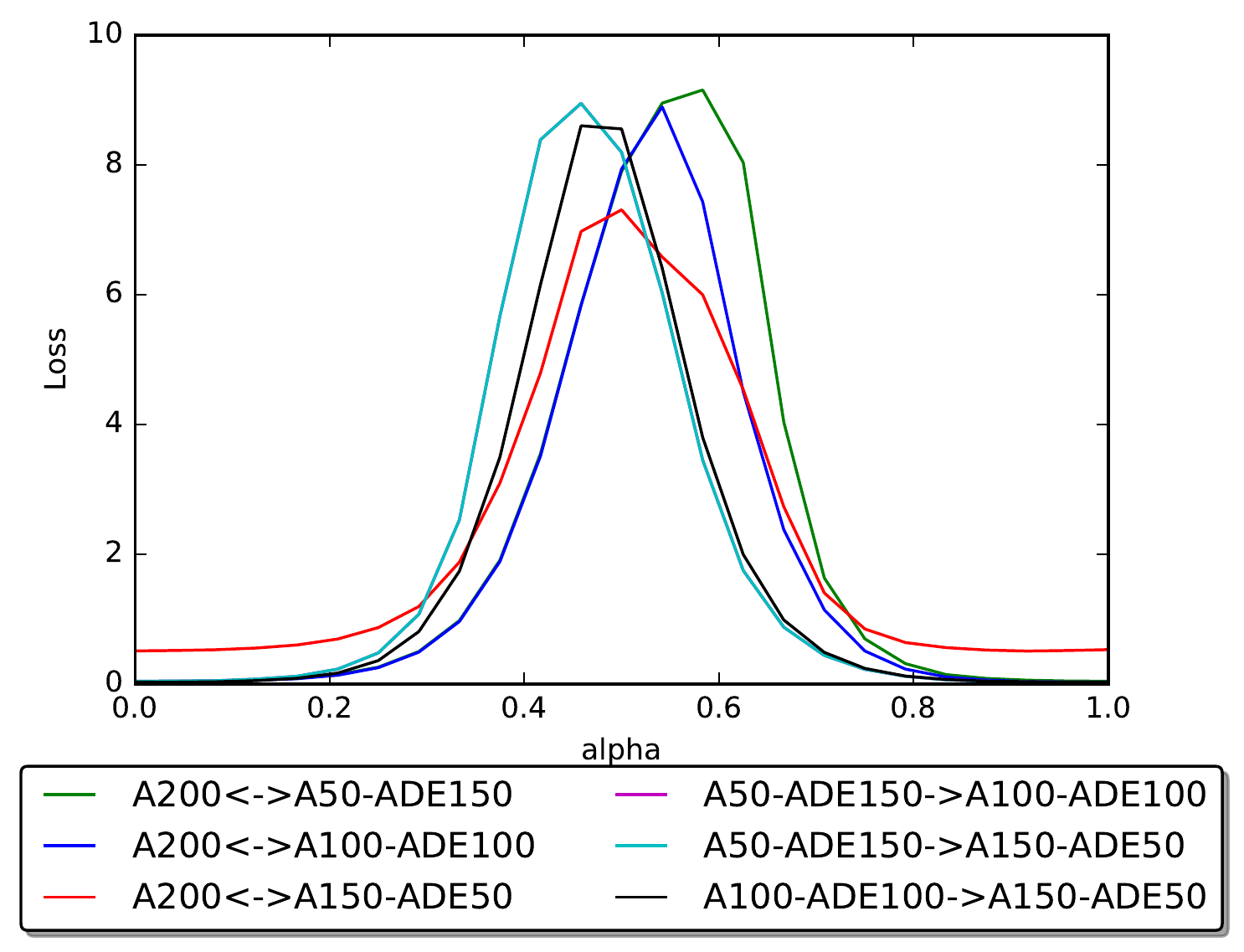}
      \includegraphics[width=0.455\linewidth]{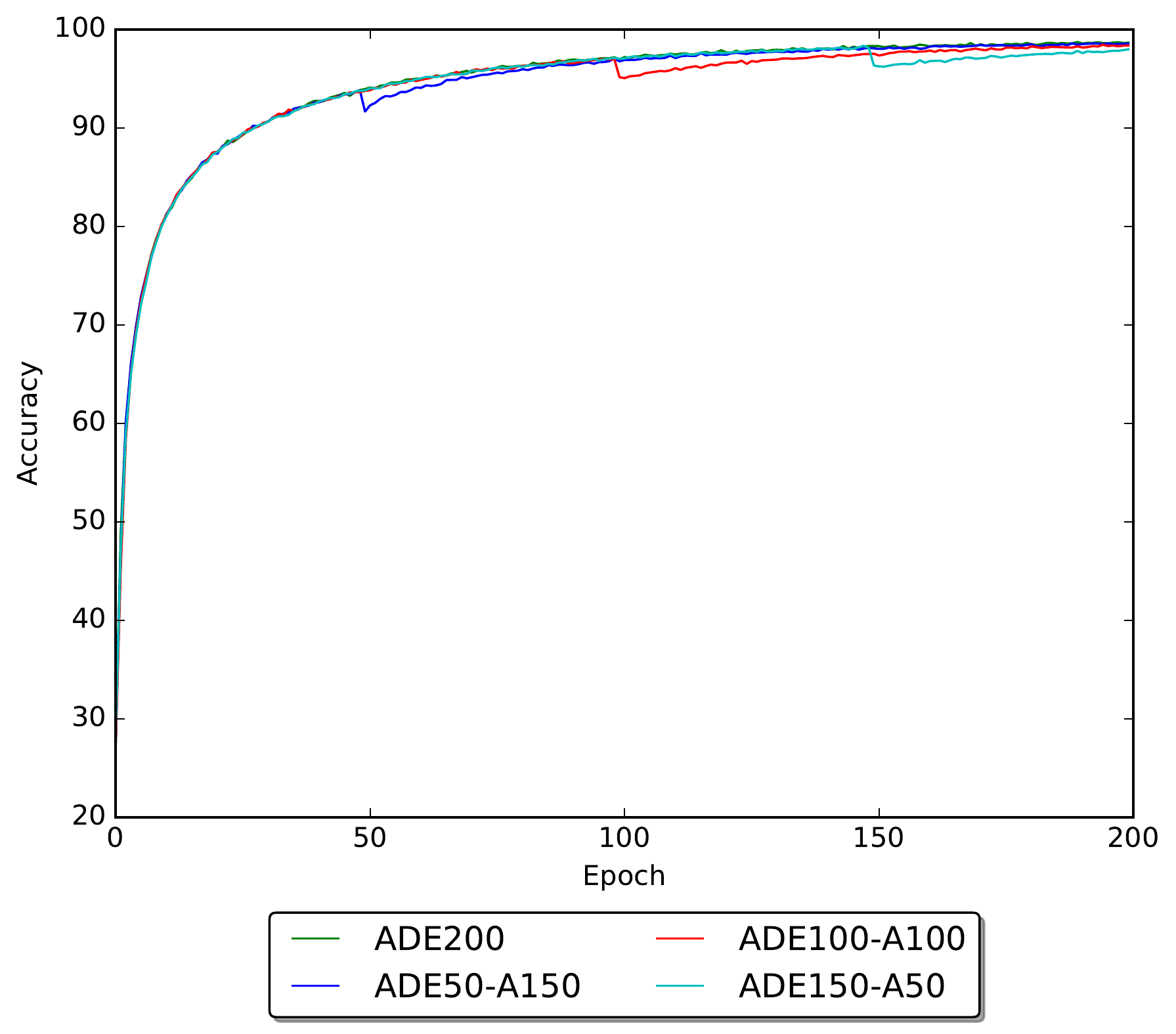}
      \subcaption{VGG: Switching from ADAM (A, $\eta = .001$) to Adadelta (ADE).}
      % fig:nin_switch_adam_adadelta
    \end{center}
    \end{minipage}
    \begin{minipage}{.495\linewidth}
      \begin{center}
        \includegraphics[width=0.525\linewidth]{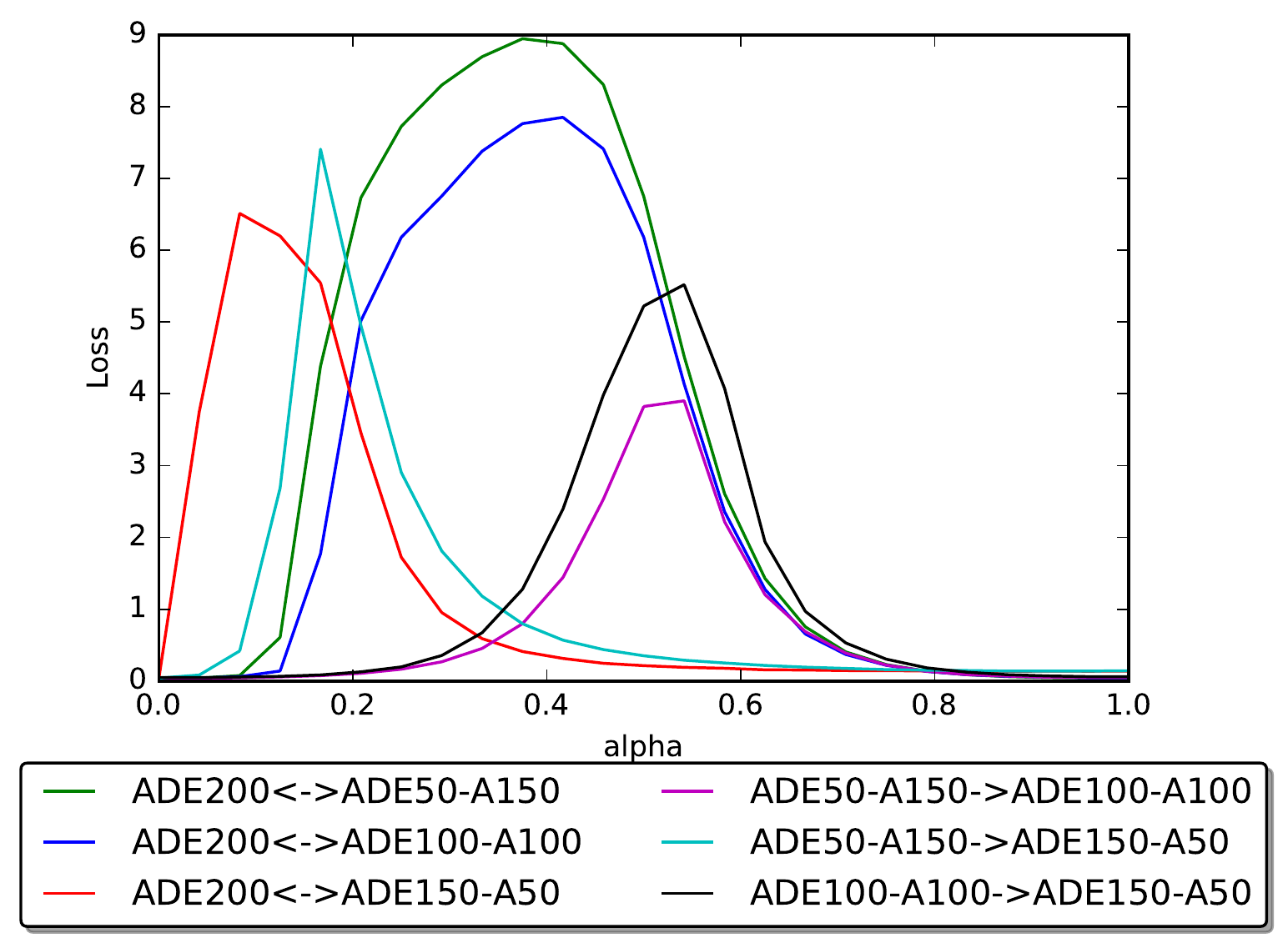}
        \includegraphics[width=0.455\linewidth]{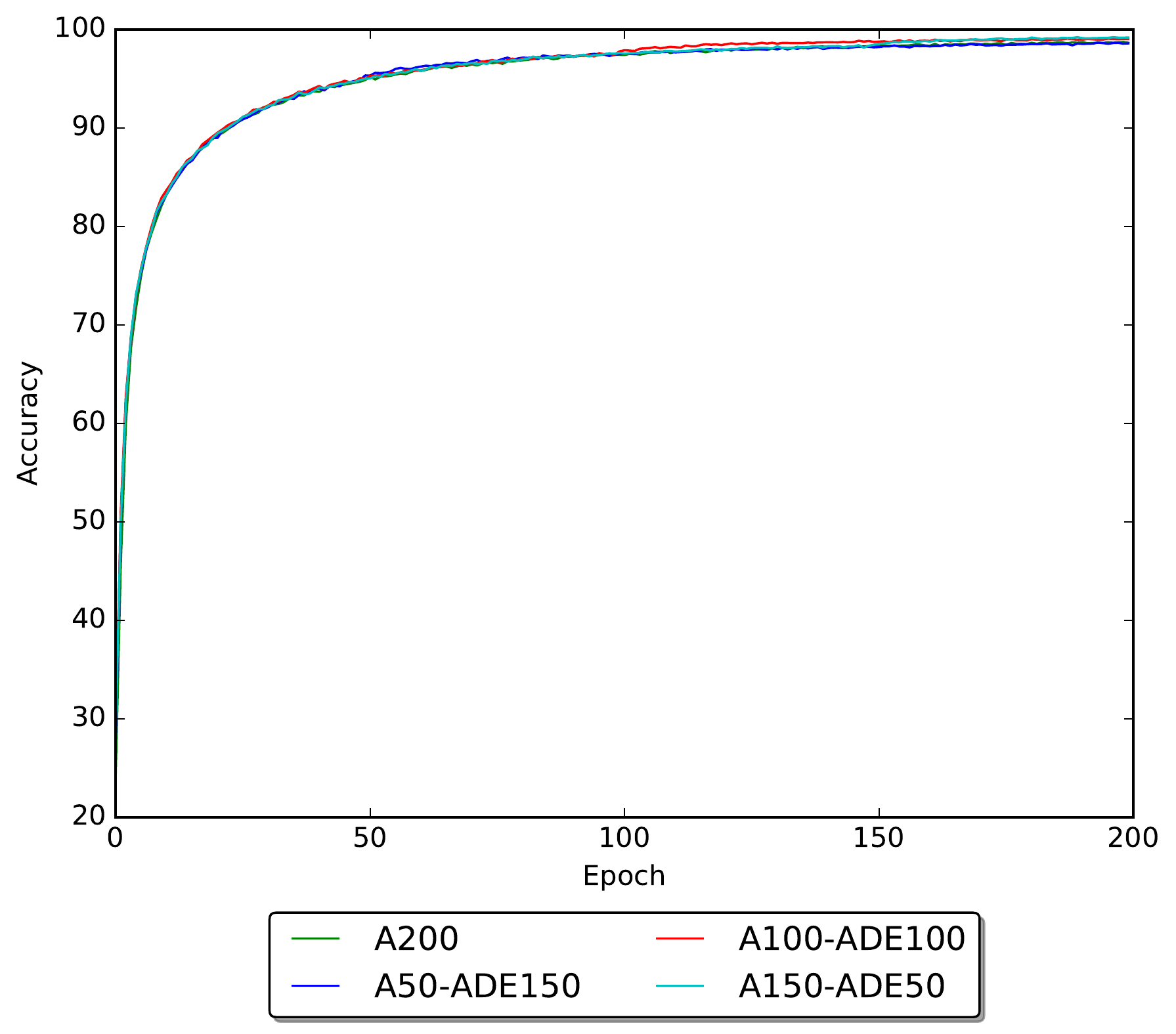}
        \subcaption{VGG: Switching from Adadelta (ADE) to ADAM (A, $\eta = .001$).}
        % fig:nin_switch_adadelta_adam
      \end{center}
    \end{minipage}
    \caption{Effects of switching from one optimization method to another at epochs 50 and 100 and 150, or never. A50-S150 corresponds to training with Adam for 50 epochs, then switching to SGD and training for 150 epochs. A200 corresponds to training with Adam for the entire 200 epochs. {\em Left}: Loss at weight vectors interpolated between two final points. A200\textless-\textgreater A50-S150 is the loss at weight interpolations between the final point corresponding to A200 (alpha = 0) and that corresponding to A50-S150 (alpha = 1). {\em Right}: Accuracy on training data as a function of training epoch. We see qualitatively similar results for FC2 on MNIST (Supp.~Fig.~\ref{fig:opt_switch_mnist}). \label{fig:nin_switch}
}
\vspace{-.5cm}
\end{figure*}

%%\begin{figure}[htb]
%\begin{wrapfigure}{l}{0.475\textwidth}
%  %\vspace{-.3cm}
%  \begin{minipage}{\linewidth}
%    \centering
%    \includegraphics[width=0.49\textwidth]{nin_train_acc.pdf}
%    \includegraphics[width=0.49\textwidth]{nin_valid_acc.pdf}
%    \vspace{-.2cm}
%    \subcaption{VGG}
%    %\label{fig:nin_inter_render_i2f}
%  \end{minipage}\\
%  \begin{minipage}{\linewidth}
%    \centering
%    \includegraphics[width=0.49\textwidth]{mnist_nn_tr_acc_opt_10folds.pdf}
%    \includegraphics[width=0.49\textwidth]{mnist_nn_vl_acc_opt_10folds.pdf}
%    \vspace{-.2cm}
%    \subcaption{MNIST}
%    %\label{fig:vgg_inter_render_i2f}
%  \end{minipage}
%  \caption{(a-b) Training and test accuracy for each of the optimization methods for (a) the VGG network on CIFAR10 and (b) the FC2 network on MNIST. Colors correspond to different initializations.
%    \label{fig:AccuracyComparison}}
%\vspace{-.4cm}
%\end{wrapfigure}
%%\end{figure}

In Figure~\ref{fig:nin_switch}, for a given pair of optimization algorithms, we plot the loss surface between final points found by switching algorithms at different points during training. For all pairs of algorithms and all switching points, the final points correspond to different critical points in the projected space. For example, different critical points in the projected space (left column) are found using Adam versus switching from Adam to SGD after 150 epochs of training, when the training accuracy has nearly plateaued (right column). This suggests that saddle points are encountered late in the optimization at which different descent directions are chosen by the different algorithms. This disagrees with the common lore that the local minimum has effectively been chosen in the transient phase, and instead suggests that which solution is found is still in flux late in optimization~\citep{Sutskever2013}. It appears that this switch from one critical point to another happens almost immediately after the optimization method switches, with the training accuracy jumping to the characteristic accuracy for the given method within a few epochs (Figure~\ref{fig:nin_switch}, right). 
%\begin{figure}[htb]
\begin{wrapfigure}{l}{0.475\textwidth}
  \vspace{-.4cm}
  \begin{minipage}{0.42395\linewidth}
    \centering
    \includegraphics[width=\textwidth]{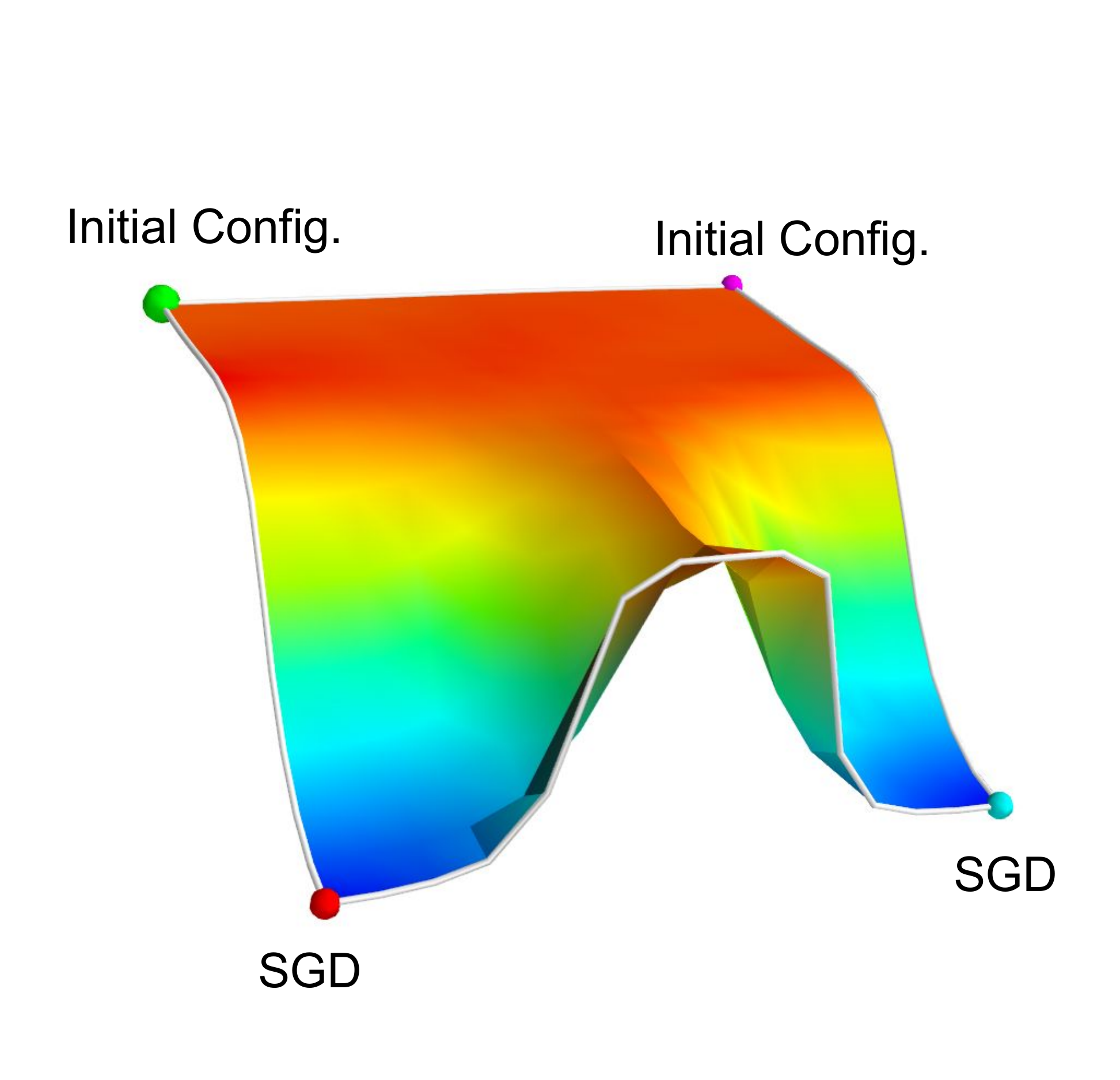}
    \vspace{-.3cm}
    \subcaption{NIN}
    %\label{fig:nin_inter_render_i2f}
  \end{minipage}
  \begin{minipage}{0.42395\linewidth}
    \centering
    \includegraphics[width=\textwidth]{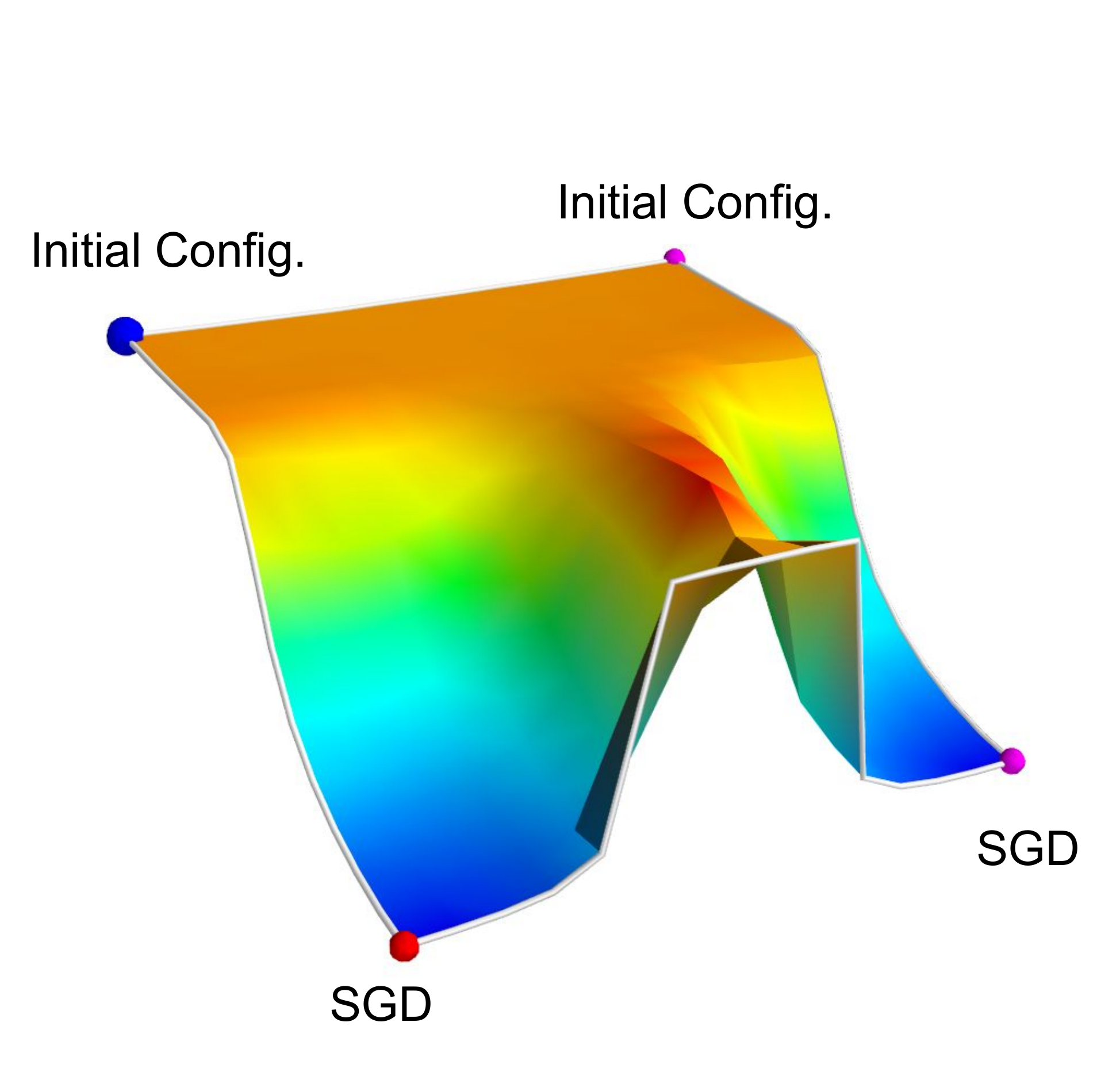}
    \vspace{-.3cm}
    \subcaption{VGG}
    %\label{fig:vgg_inter_render_i2f}
  \end{minipage}
  \caption{Visualization of the loss surface at weights interpolated between two initial configurations and the final weight vectors learned using SGD from these initializations, for the VGG and NIN networks on CIFAR10. \label{fig:quad_render2}}
  \vspace{-.5cm}
\end{wrapfigure}
%\end{figure}

%Interestingly, we also see the distance between the initial and current weight vectors changes drastically after switching from one optimization method to another, and that this distance is characteristic per algorithm (Figure~\ref{fig:nin_switch}, middle column). While distance increases with training epoch for any single optimization method, it actually starts to decrease when switching from ADAM to SGD. 

%What is intriguing is that each switched method has the tendency to travel back to where  

%\begin{compactitem}
%    \item Figure~\ref{fig:nin_switch} show what goes on even after steep drop of loss. 
%        One may possibly think that after the steep drop, the 
%        neural network already stuck in a local valley and is just settling
%        down to the minima. But the (switching-method) figures show that 
%        we can still change different minima going in to another
%        decent direction.
%    \item This could be due to having a massive kernel as neural network is underdeterminded system.
%    \item Another observation is that when the method is switched, they have tendency to converge to the same distance. (Requires running longer number of epoch) 
%\end{compactitem}
%Switching graphs
%More Init -> 2point  point interpolation

Our experiments thus far have suggested that deep network loss function have many similarly good solutions in terms of training and validation error (see Figure~\ref{fig:MinimumComparison}) %local minima. 
However, deep networks are overparameterized. For example, if we switch all corresponding weights for a pair of nodes in our network, we will obtain effectively the same network, with both the original and permuted networks outputting the same prediction for a given input.
%\mtao{its not clear to me whether symmetry up to permutations implies overparameterization in the way you're declaring. Right idea, though}. 
% kb: just one example of an overparameterization
To investigate whether the final points found by different algorithms corresponded to different parameterizations of equivalent networks, or to truly different networks, we compared the outputs of the networks on each example in a validation data set:
\[
    \text{dist}(\theta_1, \theta_2) = \sqrt{ \frac{1}{N_{test}} \sum_{i=1}^{N_{test}} \|F(x_i,\theta_1) - F(x_i,\theta_2)\|^2},
\label{eq:functionaldist}
\]
%\mtao{what about 
%\[
%    \text{dist}(\theta_1, \theta_2) = \sqrt{ \mathbb{E}_{data} \left[\|F(x_i,\theta_1) - F(x_i,\theta_2)\|^2\right]},
%\]
%}
% kb: important to explicitly state that this is empirical over the validation data
where $\theta_1$ and $\theta_2$ are the weights learned by two different optimization algorithms, $x_i$ is the input for a validation example, and $F(x,\theta)$ is the output of the network for weights $\theta$ on input $x$. 

We found that, for all pairs of algorithms, the average distance between the outputs of the networks (Equation~\ref{eq:functionaldist}) was approximately 0.16, corresponding to a label disagreement of about 8\% (upper triangle of Fig.~\ref{fig:MinimumComparison}(c)). Given the generalization error of these networks (approximately 11\%, Fig.~\ref{fig:MinimumComparison}(b)), the maximum disagreement we could see was 22\%. Thus, these networks disagreed on a large fraction of these test examples -- over one third, and the final points found by different algorithms appear to correspond to effectively different networks, not trivial reparameterizations of the same one.

\subsection{Different optimization algorithms find different types of solutions}
\label{sec:DifferentAlgsDifferentTypesOfMinima}

\begin{figure}[htb]
  \vspace{-.2cm}
  \centering
  \begin{minipage}{0.25\linewidth}
    \centering
    \includegraphics[width=\textwidth]{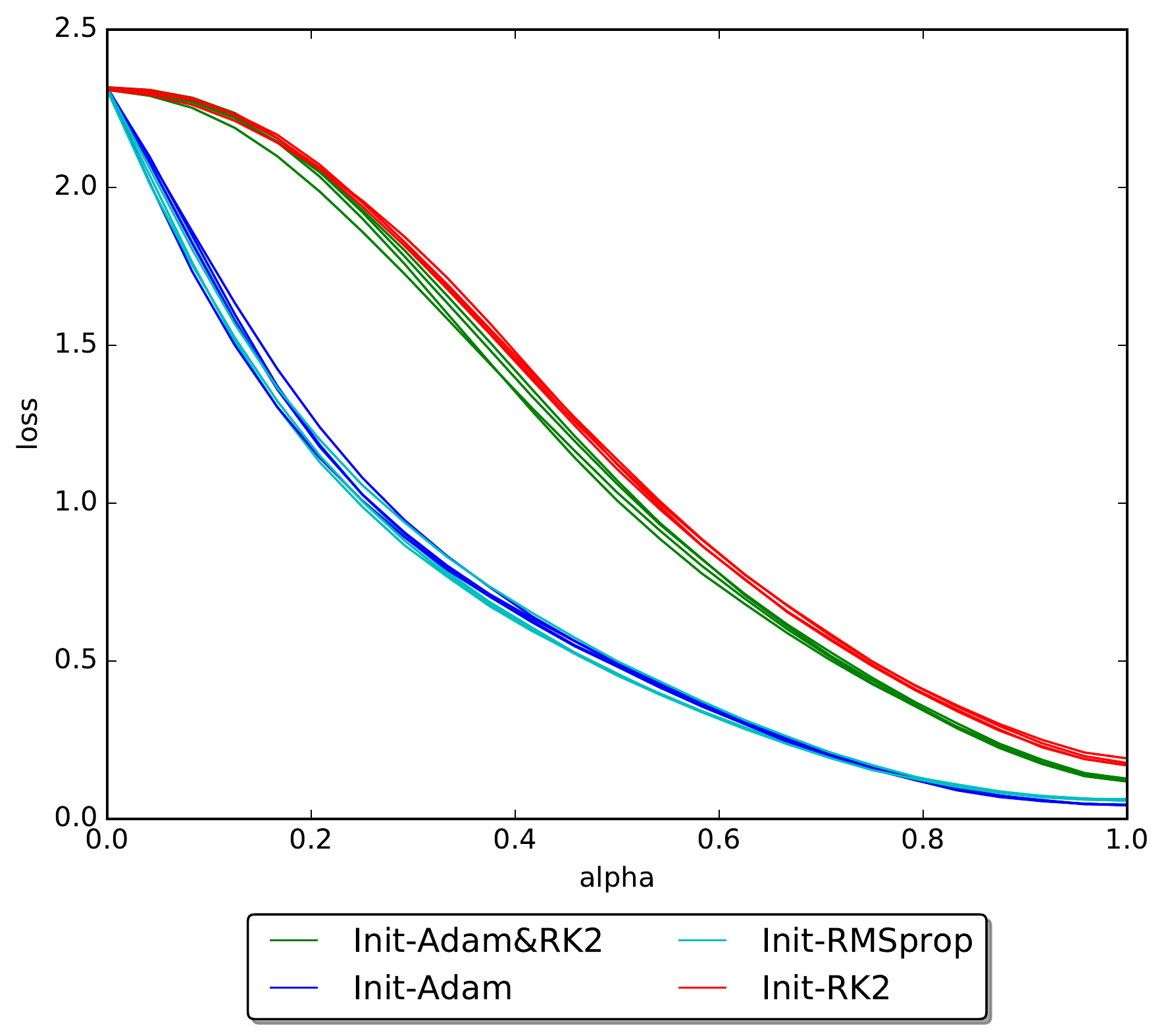}
    \subcaption{CIFAR10, NIN, Initial to Final}
  \end{minipage}
  \begin{minipage}{0.25\linewidth}
    \centering
    \includegraphics[width=\textwidth]{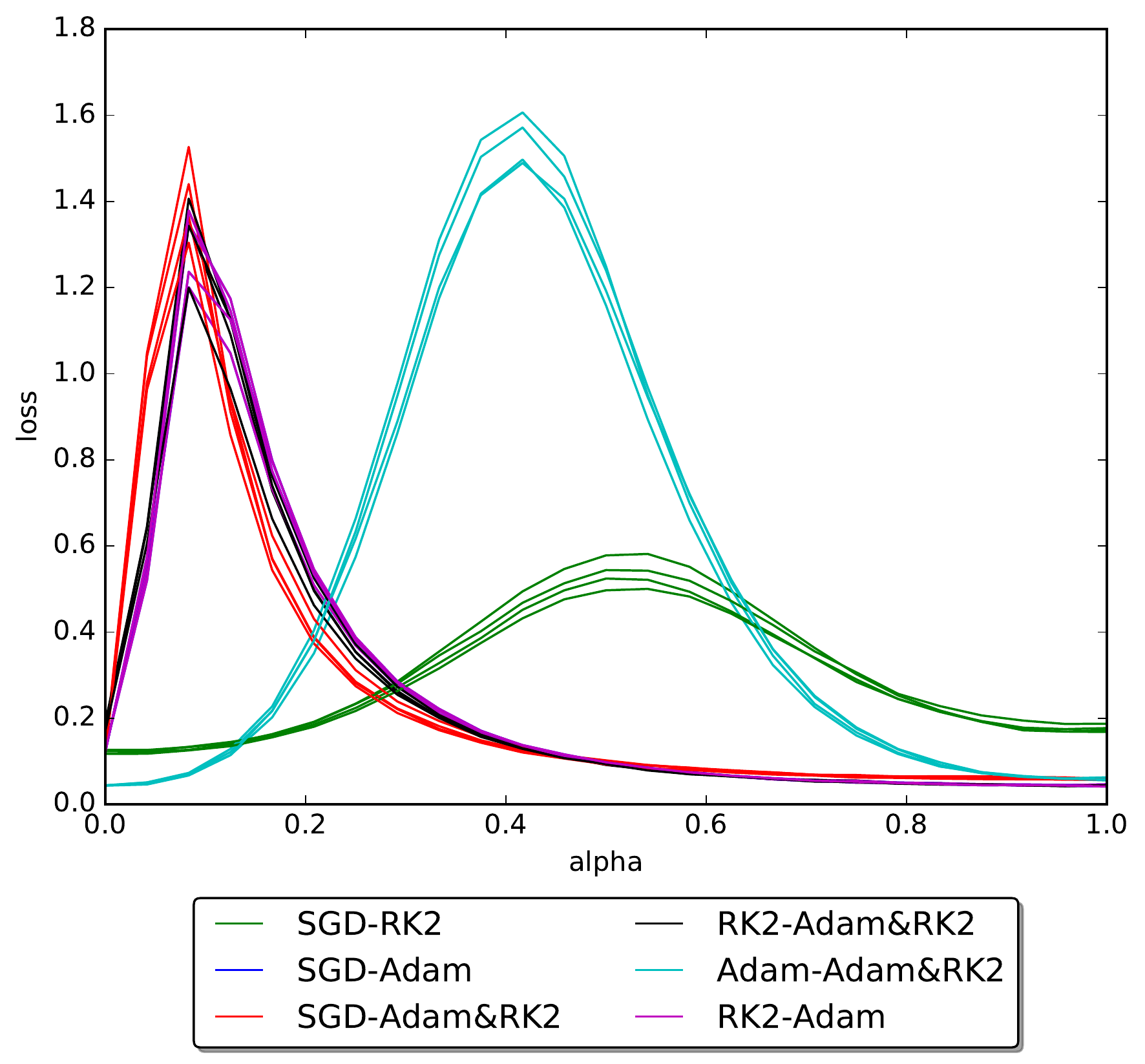}
    \subcaption{CIFAR10, NIN, Final to Final}
  \end{minipage}
  \begin{minipage}{0.24\linewidth}
    \centering
    \includegraphics[width=\textwidth]{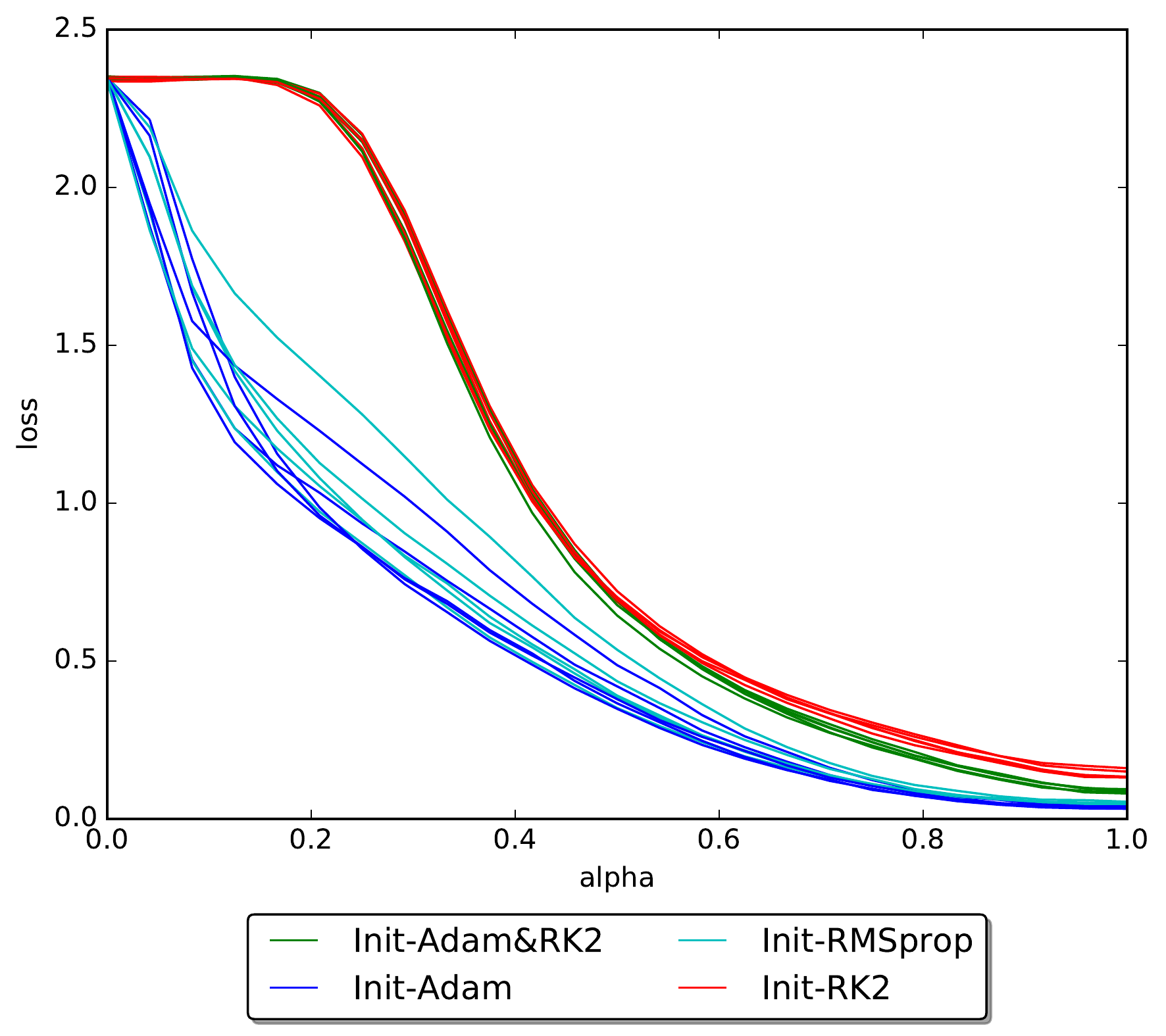}
    \subcaption{CIFAR10, VGG, Initial to Final}
  \end{minipage}
  \begin{minipage}{0.24\linewidth}
    \centering
    \includegraphics[width=\textwidth]{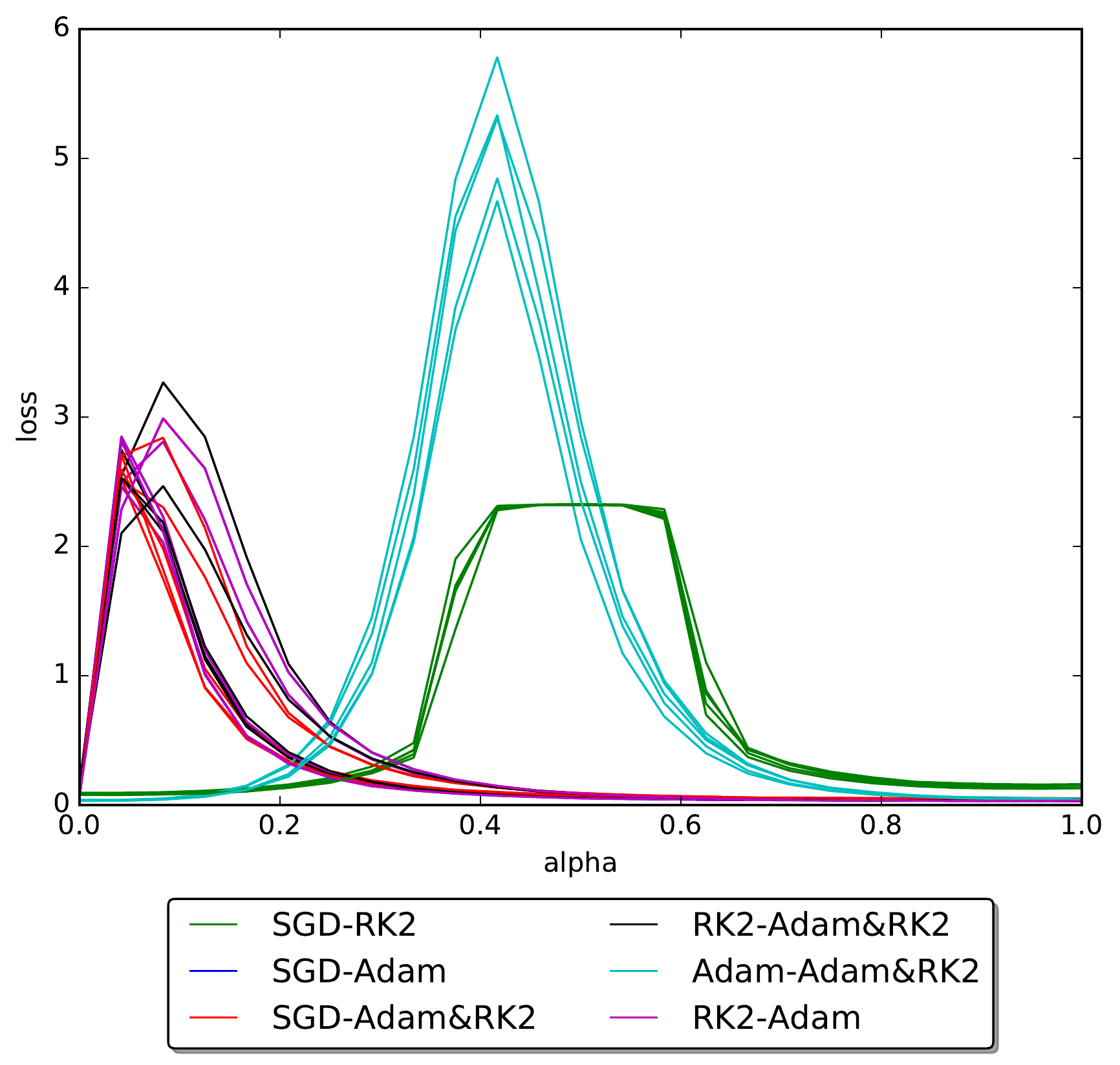}
    \subcaption{CIFAR10, VGG, Final to Final}
  \end{minipage}
  \caption{Loss function visualizations for multiple re-runs of each algorithm. Each re-run corresponds to a different initialization. We see that the loss function near the final point for a given algorithm has a characteristic geometry. \label{fig:interpolations}}
  \vspace{-.1cm}
\end{figure}

\begin{figure}[htb]
  \begin{minipage}{0.322\linewidth}
    \centering
    \includegraphics[width=\linewidth]{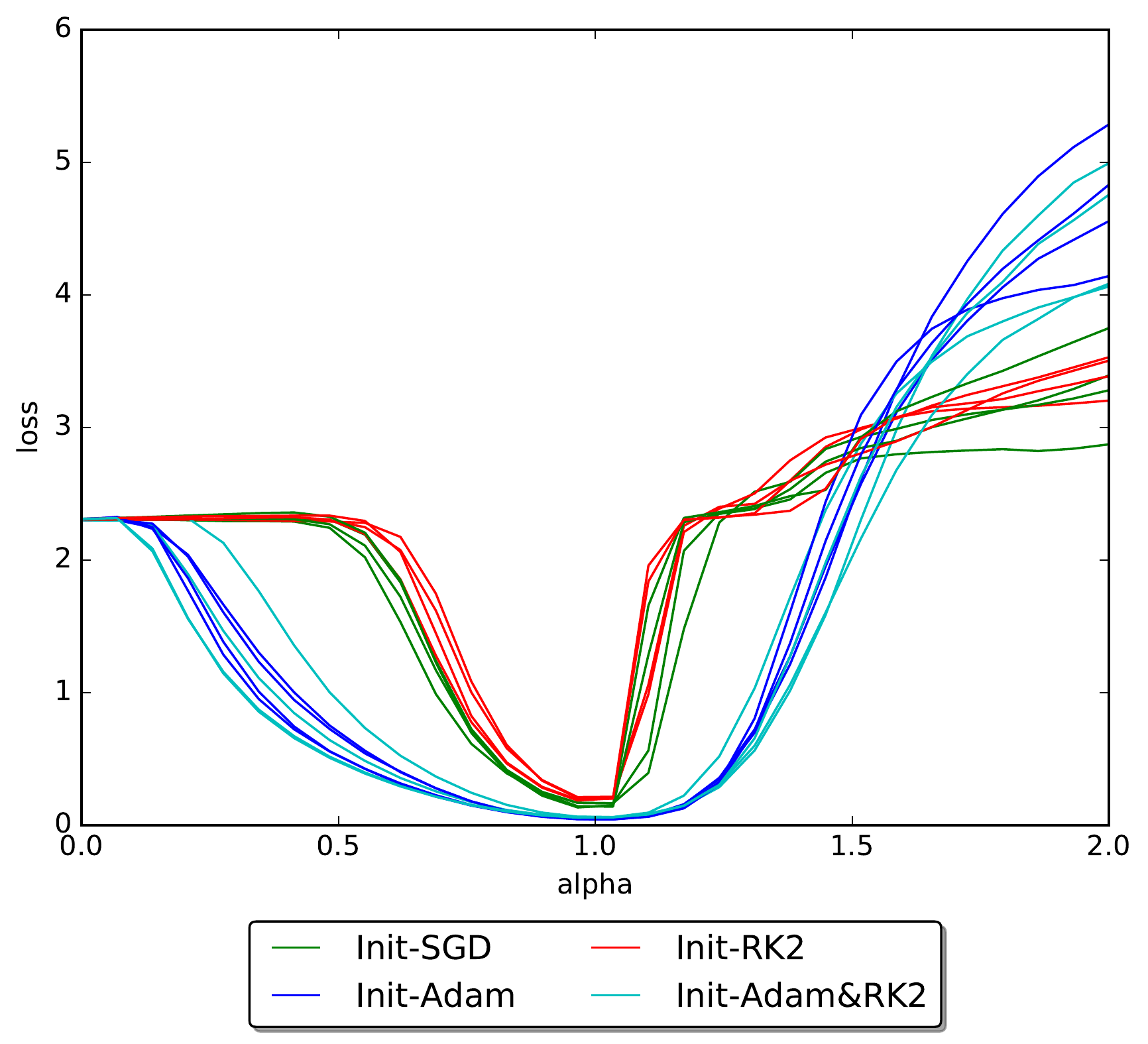}
    \vspace{-.25cm}
    \subcaption{CIFAR10, NIN}
    %\label{fig:interpolation_alpha2}
  \end{minipage}
  \begin{minipage}{0.322\linewidth}
    \centering
    \includegraphics[width=\linewidth]{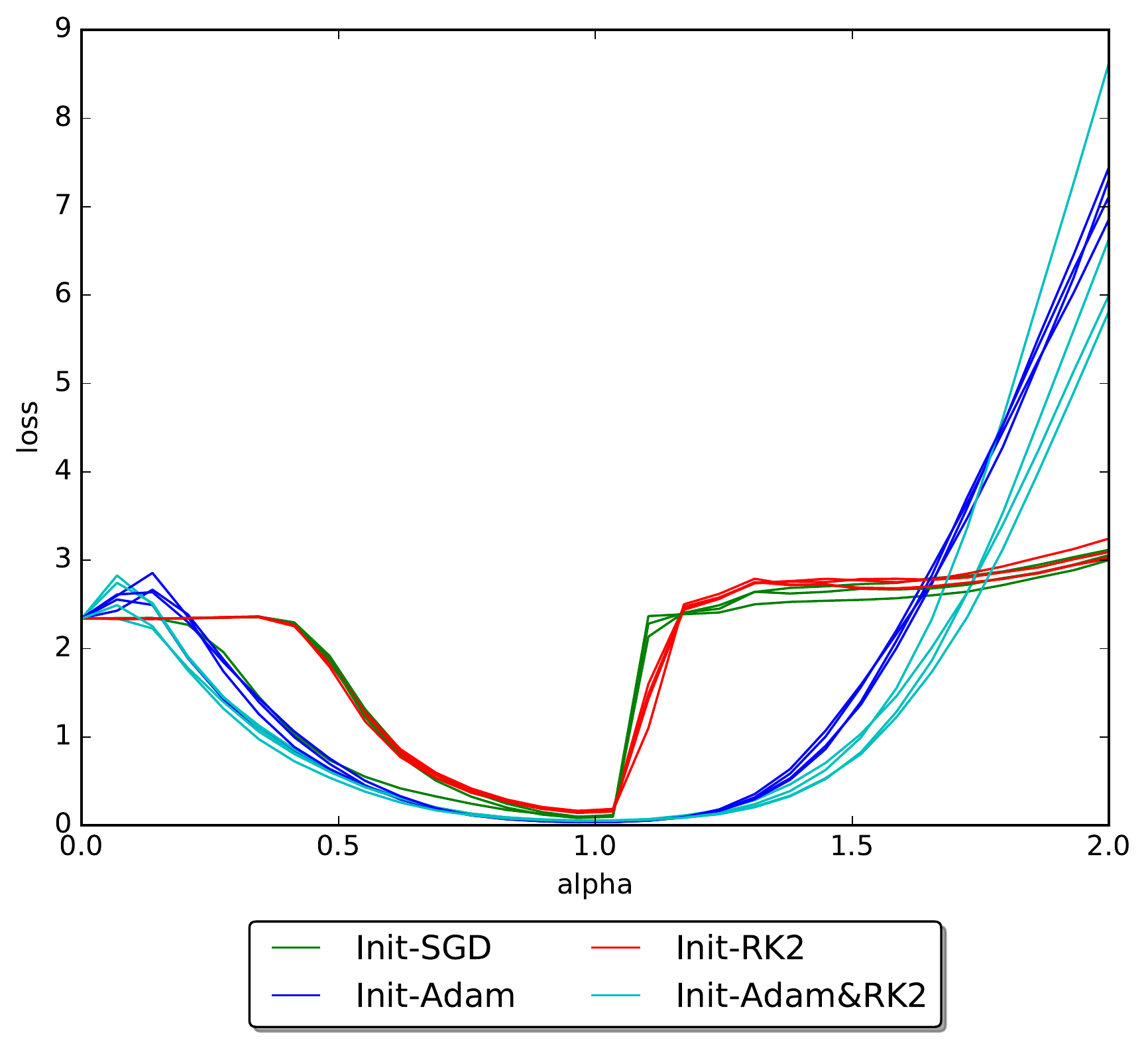}
    \vspace{-.25cm}
    \subcaption{CIFAR10, VGG}
    %\label{fig:interpolation_alpha2}
  \end{minipage}
  \begin{minipage}{0.322\linewidth}
    \centering
    \includegraphics[width=\linewidth]{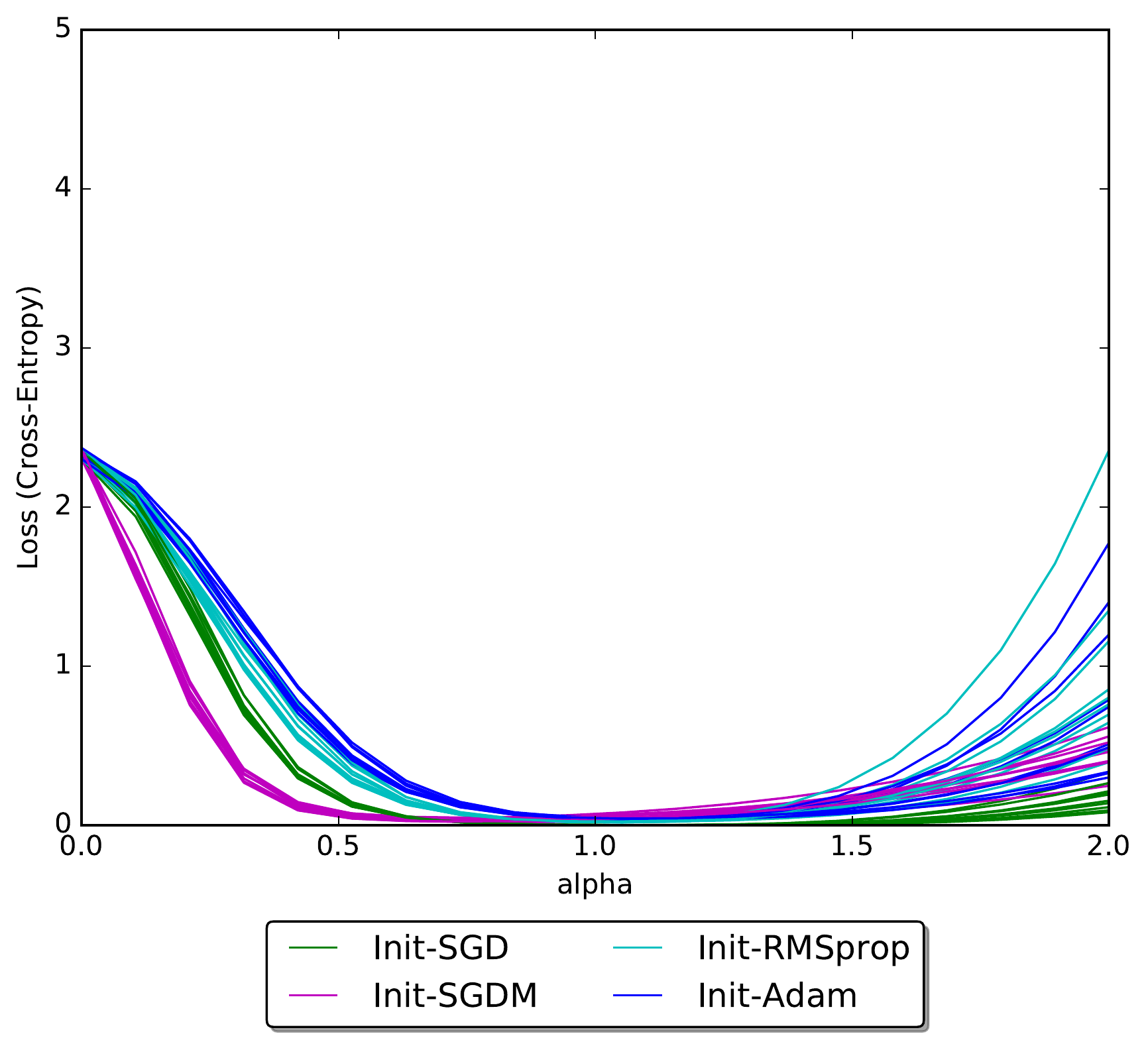}
    \vspace{-.25cm}
    \subcaption{MNIST, FC2}
    %\label{fig:interpolation_alpha2}
  \end{minipage}\\
  \begin{minipage}{0.322\linewidth}
    \centering
    \includegraphics[width=\linewidth]{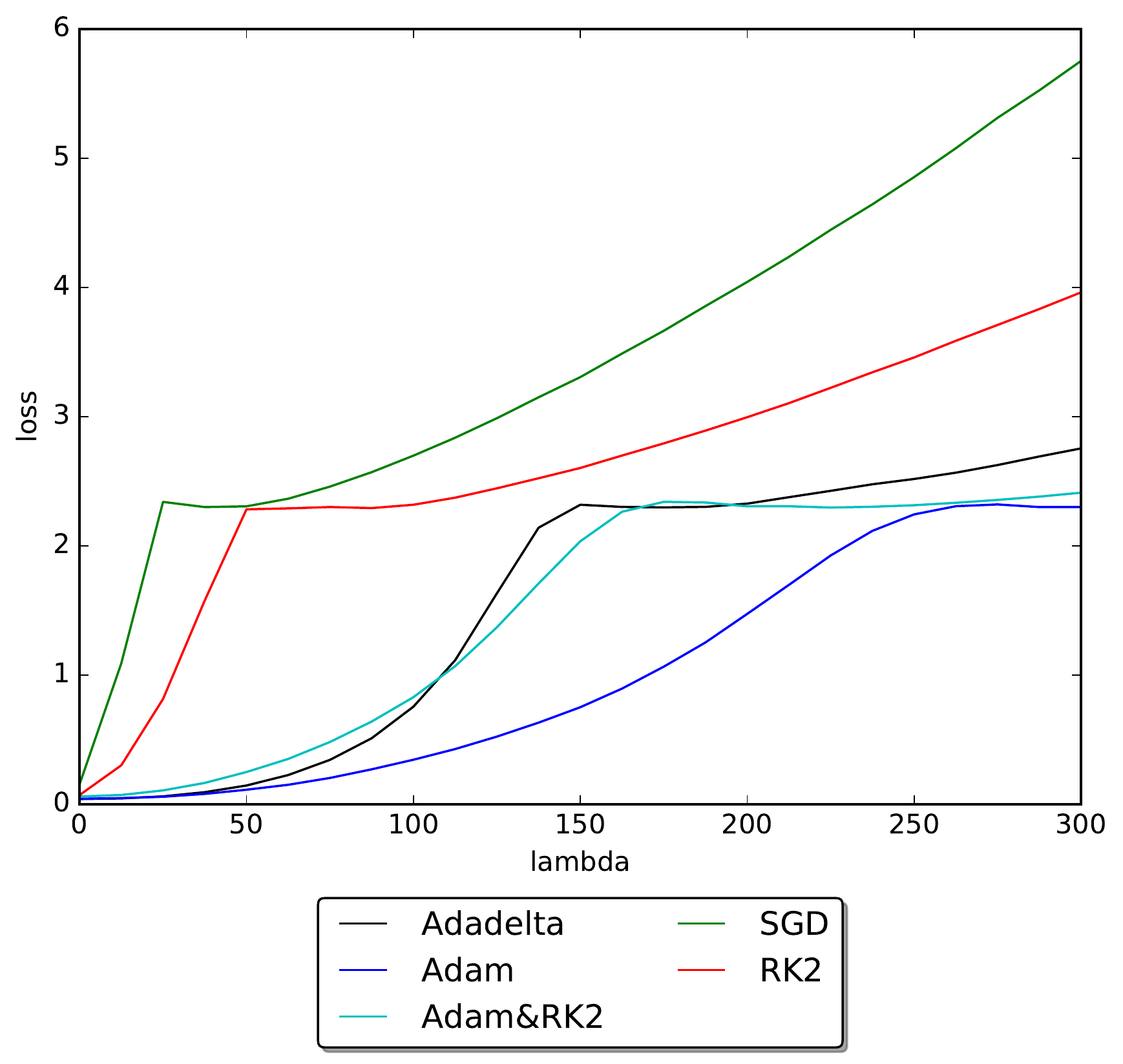}
    \vspace{-.2cm}
    \subcaption{CIFAR10, NIN}
    %\label{fig:absolute_loss_surface}
  \end{minipage}
  \begin{minipage}{0.322\linewidth}
    \centering
    \includegraphics[width=\linewidth]{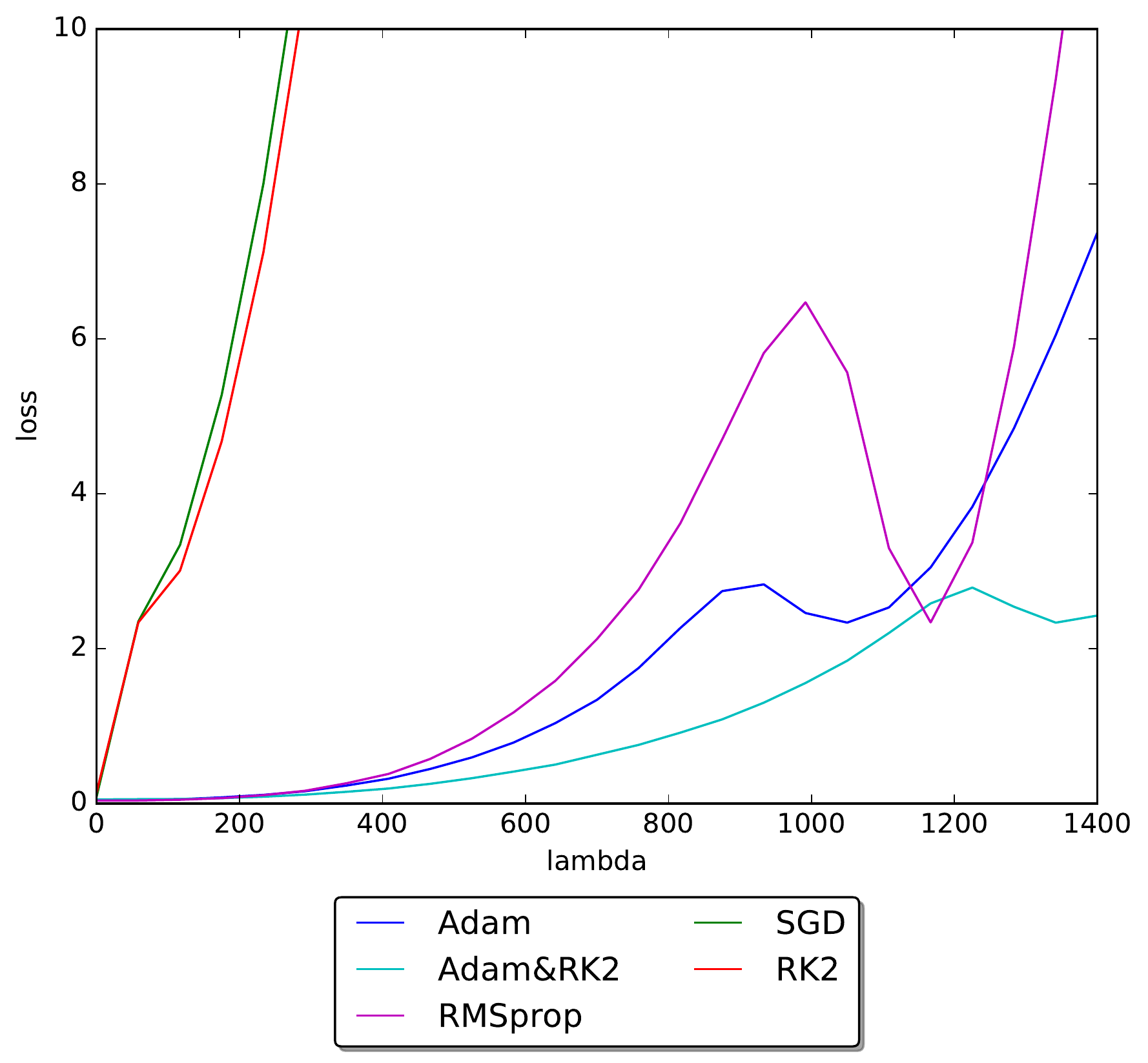}
    \vspace{-.25cm}    
    \subcaption{CIFAR10, VGG}
    %\label{fig:absolute_loss_surface}
  \end{minipage}
  \begin{minipage}{0.322\linewidth}
    \centering
    \includegraphics[width=\linewidth]{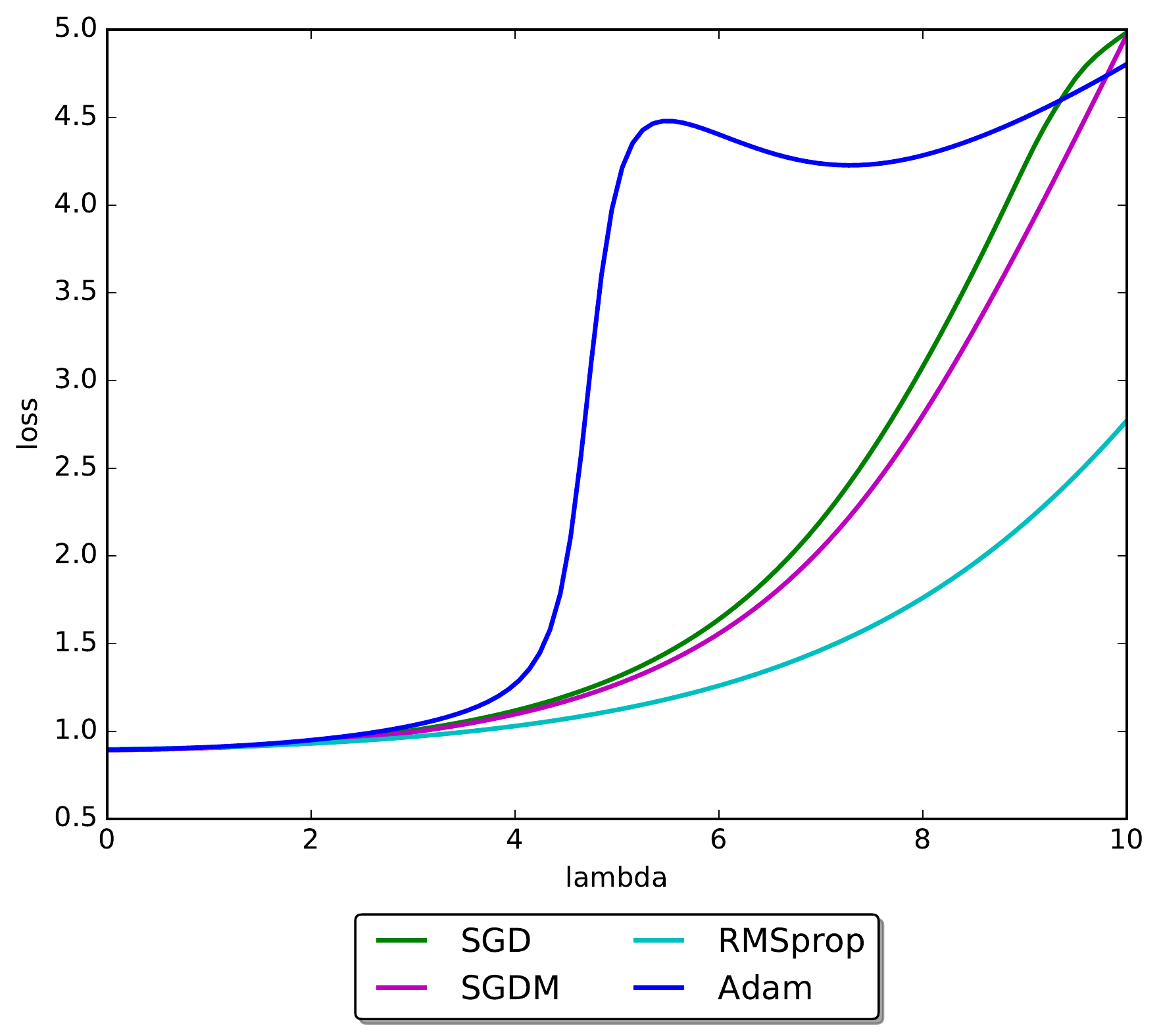}
    \vspace{-.25cm}    
    \subcaption{MNIST, FC2}
    %\label{fig:absolute_loss_surface}
  \end{minipage}
  \caption{Comparison of the size of the projected space local minima basins for different algorithms for NIN (a-b) and VGG (c-d) on CIFAR10 and for FC2 on MNIST (e-f). In (a,c,e), the units of the x-axis (alpha) are relative to the scale of the final weight vector of the algorithm, with alpha$=0$ corresponding to the initial weight vector, alpha$=1$ corresponding to the final weight vector and alpha$>1$ corresponding to points interpolated beyond the final weight vector. In (b,d,f), the units of the x-axis (lambda) are the same for each algorithm -- the units of the high-dimensional weight-vector space (Eq.~\ref{eq:absolutebasin}). lambda$=0$ indicates the final weight vector found by the algorithm. In (b,d), the kink near loss$=2.4$ corresponds to the initial weight vector. \label{fig:basinsize}}
%\end{wrapfigure}
\end{figure}

\begin{figure}[htb]
%\begin{wrapfigure}{L}{0.5\textwidth}
  \centering
  % ADAM -> SGD
  \begin{minipage}{.24\textwidth}
    \begin{center}
      \includegraphics[width=\textwidth]{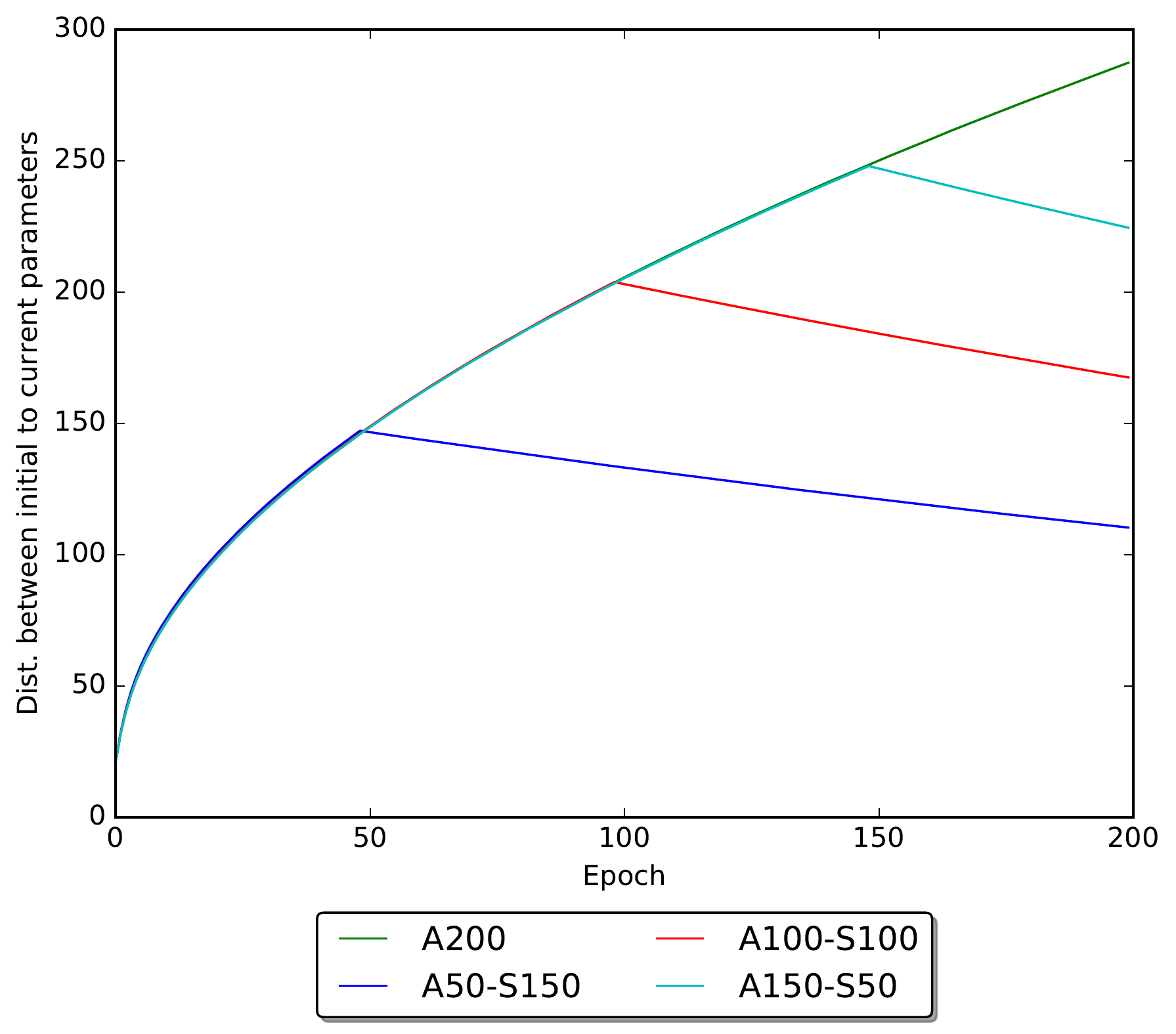}
      \vspace{-.2cm}
      \subcaption{}
    \end{center}
  \end{minipage}
  \begin{minipage}{.24\textwidth}
    \begin{center}
      \includegraphics[width=\textwidth]{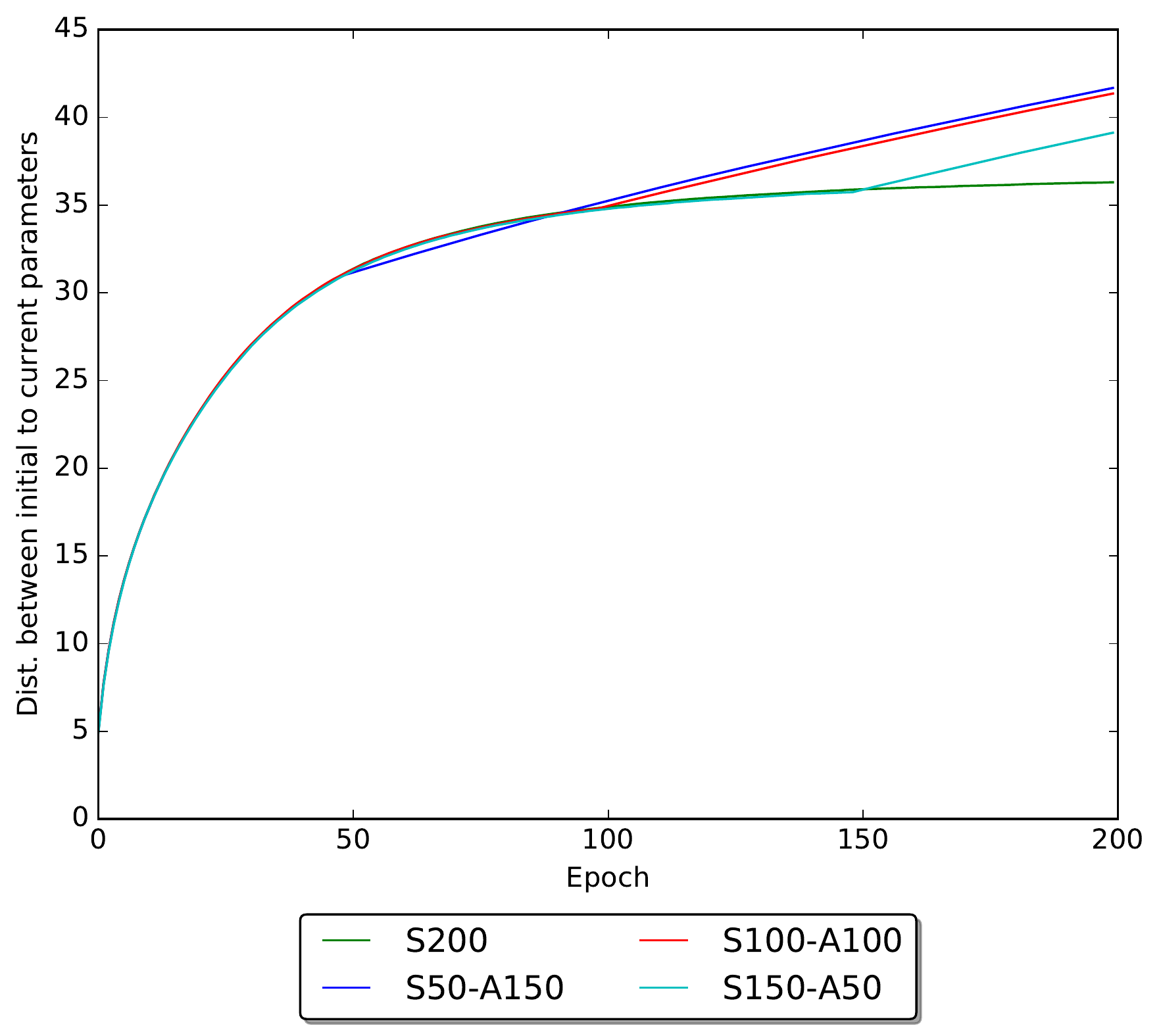}
      \vspace{-.2cm}
      \subcaption{}
    \end{center}
  \end{minipage}
  \begin{minipage}{.24\textwidth}
    \begin{center}
      \includegraphics[width=\textwidth]{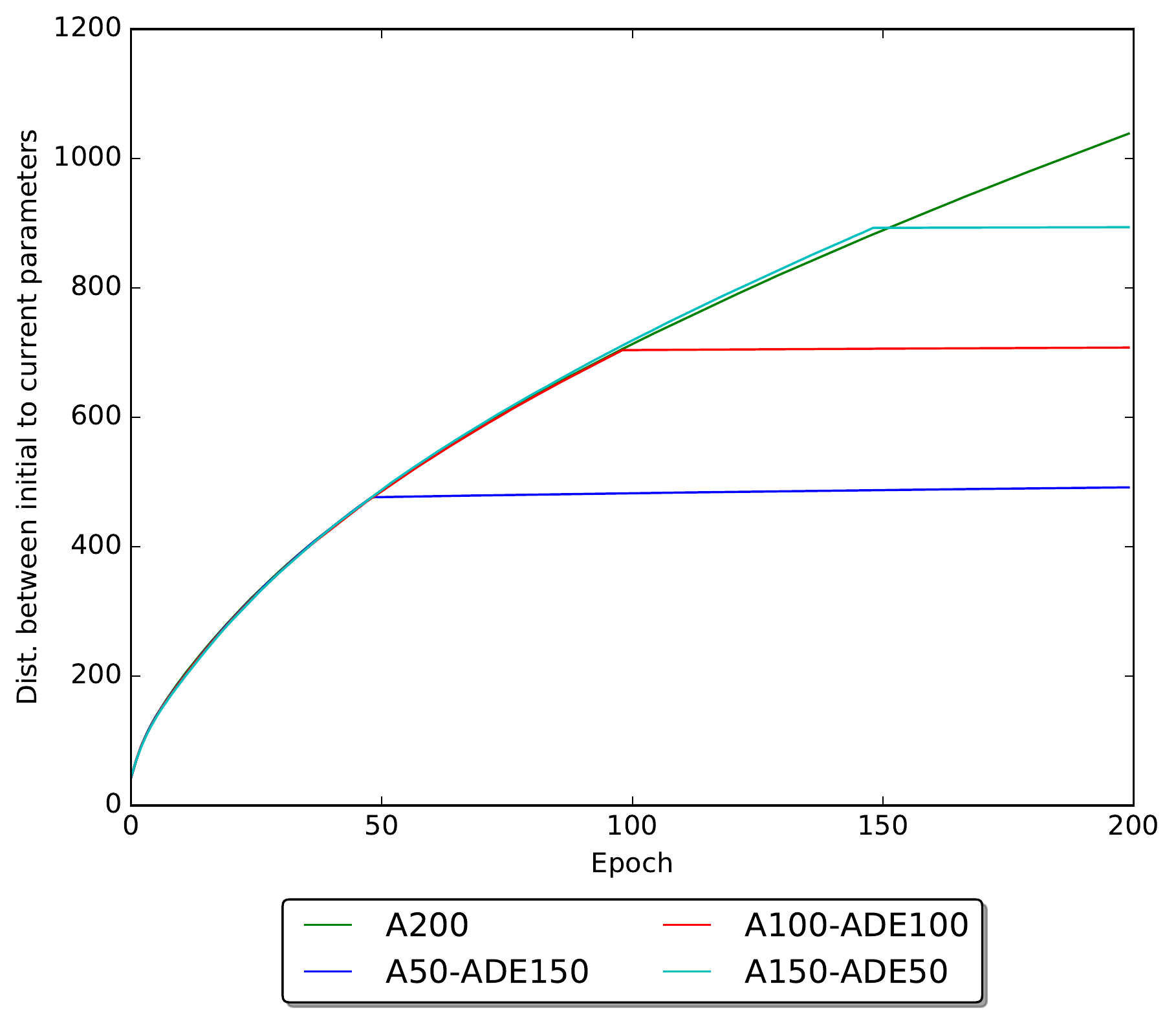}
      \vspace{-.2cm}
      \subcaption{}
    \end{center}
  \end{minipage}
  \begin{minipage}{.24\textwidth}
    \begin{center}
      \includegraphics[width=\textwidth]{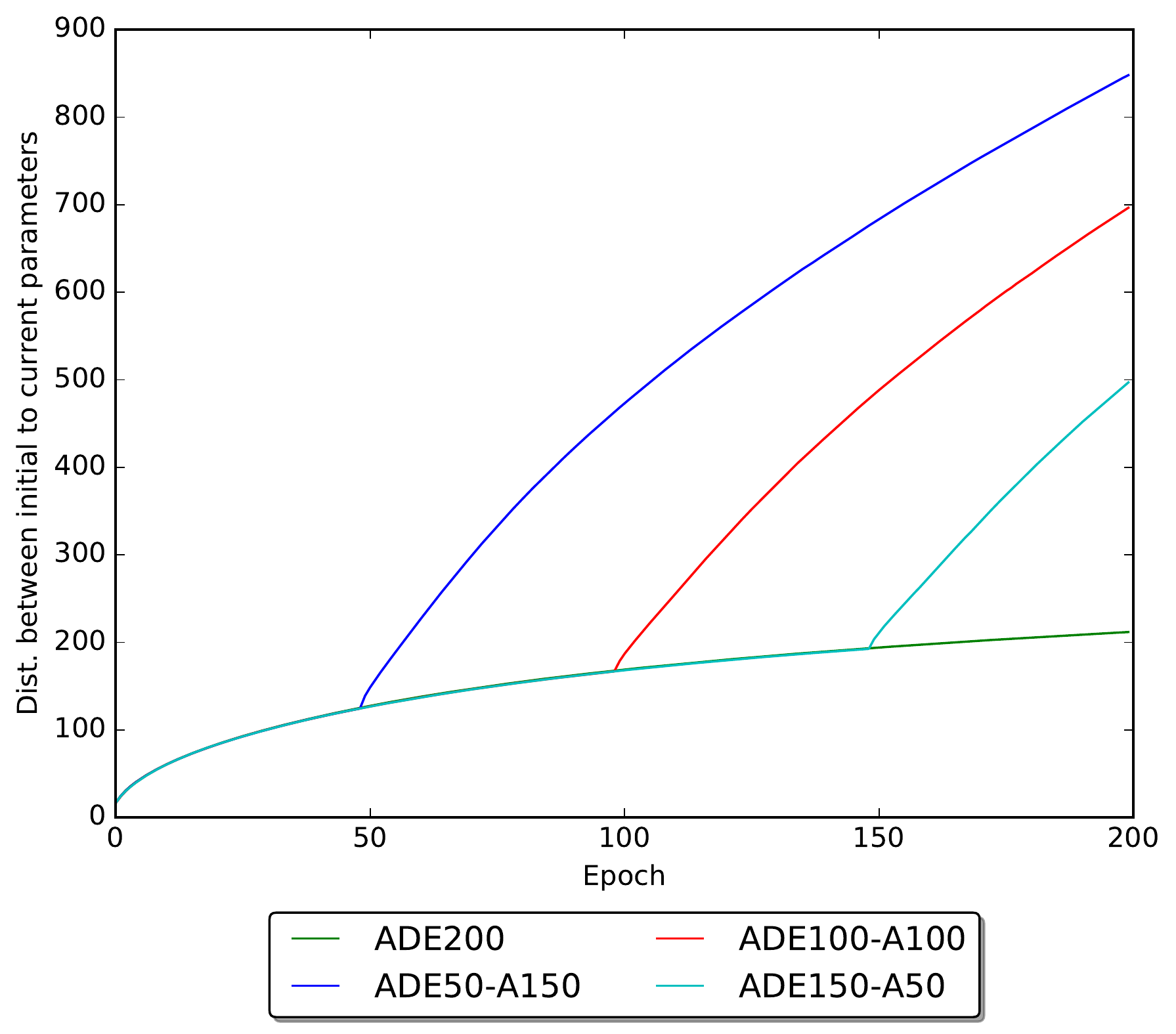}
      \vspace{-.2cm}
      \subcaption{}
    \end{center}
  \end{minipage}
  \caption{
    Distance from initial to current weights during training in which we switch from one optimization method to another at epochs 50 and 100 and 150, or never. Different plots correspond to different networks and pairs of algorithms, following Fig.~\ref{fig:nin_switch}. Different optimization algorithms have a characteristic distance. After switching, this distance quickly approaches the new characteristic distance. (a) NIN: Switching from Adam (A, learning rate $\eta = .001$) to SGD (S, $\eta = .01$). (b) NIN: Switching from SGD (S, $\eta = .1$) to Adam (A, $\eta = .0001$). (c) VGG: Switching from Adam (A, $\eta = .001$) to Adadelta (ADE). (d) VGG: Switching from Adadelta (ADE) to Adam (A, $\eta = .001$). 
\label{fig:disttraveled_switch}}
\vspace{-.2cm}
%\end{wrapfigure}
\end{figure}

Next, we investigated whether the solutions found by the different optimization algorithms had distinguishing properties. To do this, we trained the networks with each optimization algorithm from different initializations. We then compared differences between runs of the same algorithm from different initializations to differences between different algorithms. 

As shown in Figure~\ref{fig:MinimumComparison}(a), in terms of training accuracy, we see some stereotypy for the final points found by different algorithms, with SGD having the lowest training accuracy and ADAM, RMSprop, and Adadelta having the highest training accuracy. However, the generalization accuracy of these different final points on validation data was not different between algorithms (Figure~\ref{fig:MinimumComparison}(b)). We also did not see a relationship between the weight initialization and the validation accuracy. This supports the conclusions of previous work that most local minima are equally good. As found in Goodfellow et al.~\citep{Goodfellow2015}, these final points correspond to different local minima in the projected space~\ref{fig:quad_render2}.

We visualized the loss surface around each of the final points for the multiple runs from different initializations. To do this, we plotted the value of the loss function between the initial and final weights for each algorithm for several runs of the algorithm with different initializations (Figure~\ref{fig:interpolations}(a,c)). In addition, we plotted the value of the loss function between the final weights for selected pairs of algorithms for each run (Figure~\ref{fig:interpolations}(b,d)). We see that the surfaces look strikingly similar for different runs of the same algorithm, but characteristically different for different algorithms. Thus, we found evidence that the final weights found by different algorithms are of similar quality in terms of validation error, they are qualitatively different. This suggests that there is some stereotypy in the descent direction chosen at saddle points by each algorithm. 

In particular, we see in Figure~\ref{fig:basinsize}(a,c) that the size of the basins in the projected spaces around the local minima found by Adam and Adam\&RK2 are larger than those found by SGD and RK2 for NIN and VGG on CIFAR10, i.e.~that the training loss is small for a wider range of $\alpha$ values. 
%\mtao{Given the odd radial symmetry of batch normalization and the skewed metric of polar coordinates (i.e $rd\Theta$), this is somewhat expected?}
% kb: I'm not sure I understand the question. I think batch normalization introduces a scale ambiguity because you essentially have two linear layers next to each other, which I think is a confound of this analysis
The x-axis of Fig.~\ref{fig:basinsize}(a,c,e), $\alpha$, is a multiplier of the weight vector. If the norm of the weight vector found by one algorithm is larger than that found by another, then a change of $\Delta \alpha$ for this curve would correspond to a larger absolute change in the weight vector, $\Delta \alpha (\theta_1 - \theta_0)$ , where $\theta_0$ is the initial weight vector $\theta_1$ is the result found by a given optimization algorithm. In Figure~\ref{fig:basinsize}(b,d,f) we again plot the basin, but normalize for the norm of the weight vector difference, and show the loss as a function of the absolute distance in parameter space:
\begin{align}
  \theta(\lambda) = \theta_1 + \lambda \frac{\theta_0 - \theta_1}{\|\theta_0 - \theta_1\|}
  \label{eq:absolutebasin}
\end{align}
Even after normalization, we see the same result for NIN and VGG on CIFAR10: the sizes of the basins in the projected spaces around the local minima are bigger for Adam and Adam\&RK2 (Fig.~\ref{fig:basinsize}(b,d)). This trend does not repeat for FC2 on MNIST, however, with Adam corresponding to the smallest basin (Fig.~\ref{fig:basinsize}(e,f)).

\begin{figure}[htb]
%\begin{wrapfigure}{L}{0.475\textwidth}
\centering
\begin{minipage}{.325\linewidth}
  \centering
  \includegraphics[width=\linewidth]{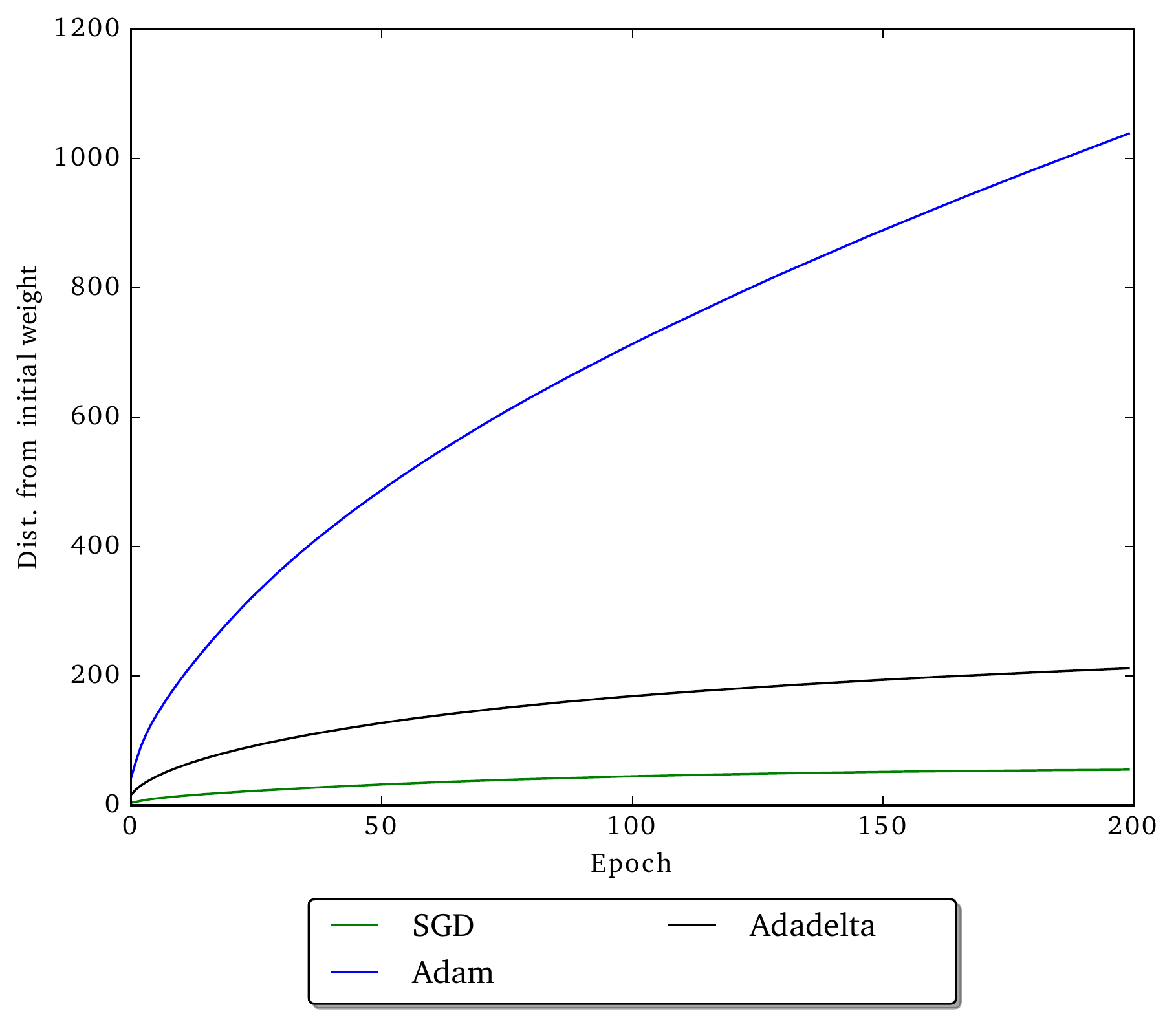}
    \vspace{-.4cm}
  \subcaption{VGG}
\end{minipage}
\begin{minipage}{.325\linewidth}
  \centering
  \includegraphics[width=\linewidth]{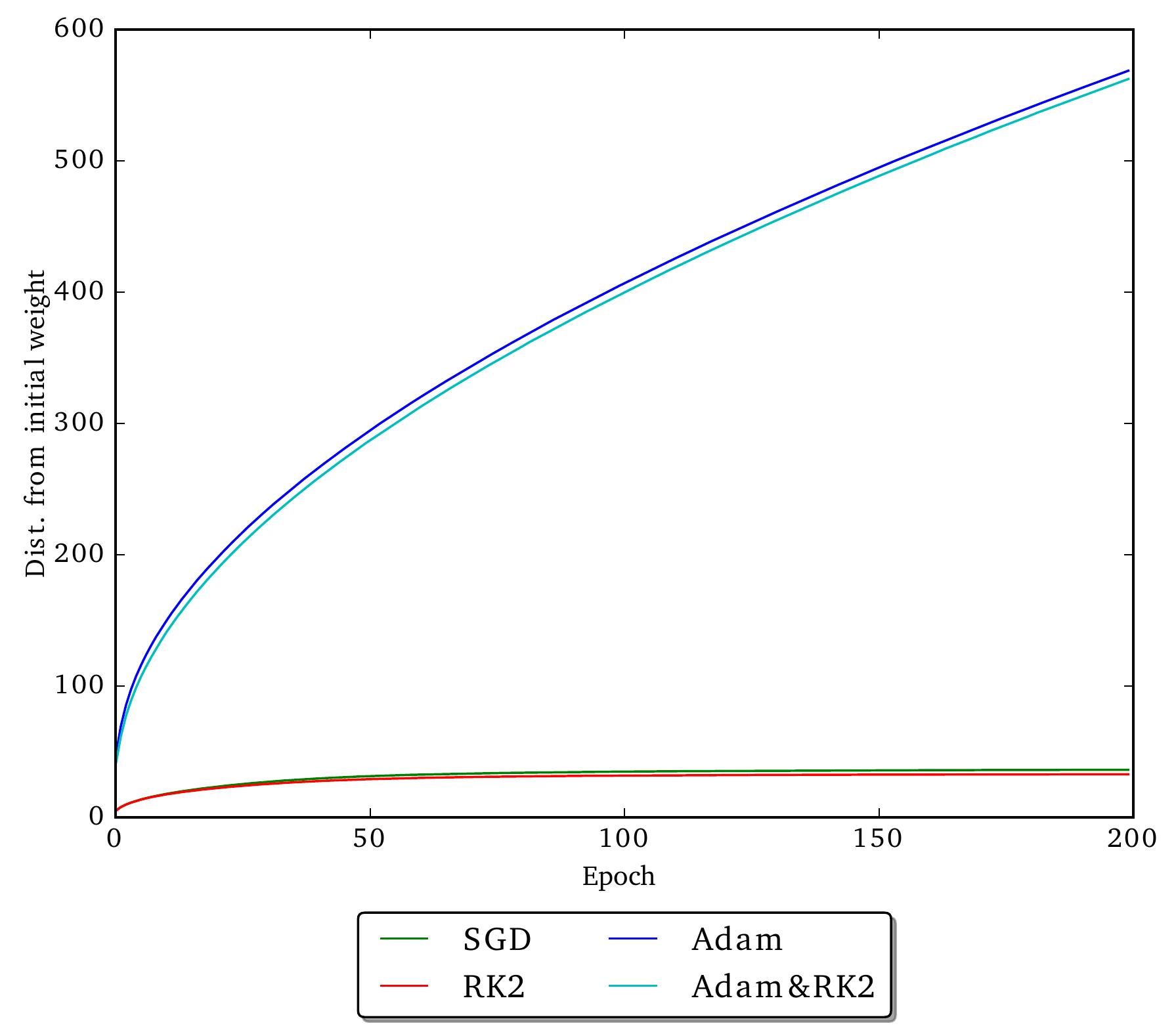}
    \vspace{-.4cm}
  \subcaption{NIN}
\end{minipage}
\begin{minipage}{.325\linewidth}
  \centering
  \includegraphics[width=\linewidth]{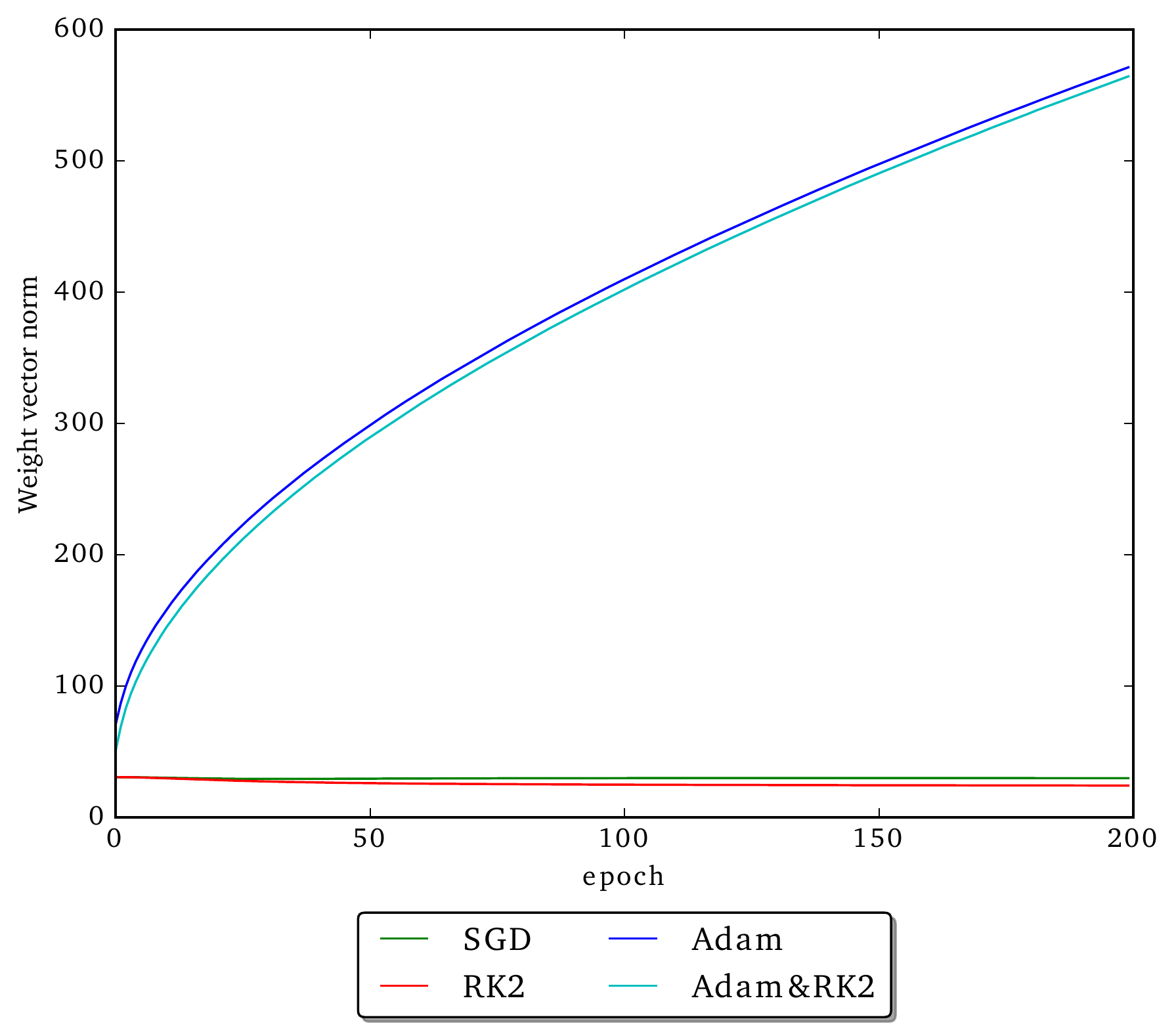}
    \vspace{-.4cm}
  \subcaption{NIN}
\end{minipage}
\caption{(a-b) Distance from initial to current weights during training for different optimization algorithms. Left: VGG network, right: NIN on CIFAR10. (c) Norm of weight vectors during training for different optimization algorithms for NIN on CIFAR10. (c) and (b) are nearly identical. \label{fig:disttraveled}}
\vspace{0.2cm}
%\end{wrapfigure}
%\begin{figure}[htb]
%\begin{wrapfigure}{R}{0.5\textwidth}
    \includegraphics[width=\linewidth]{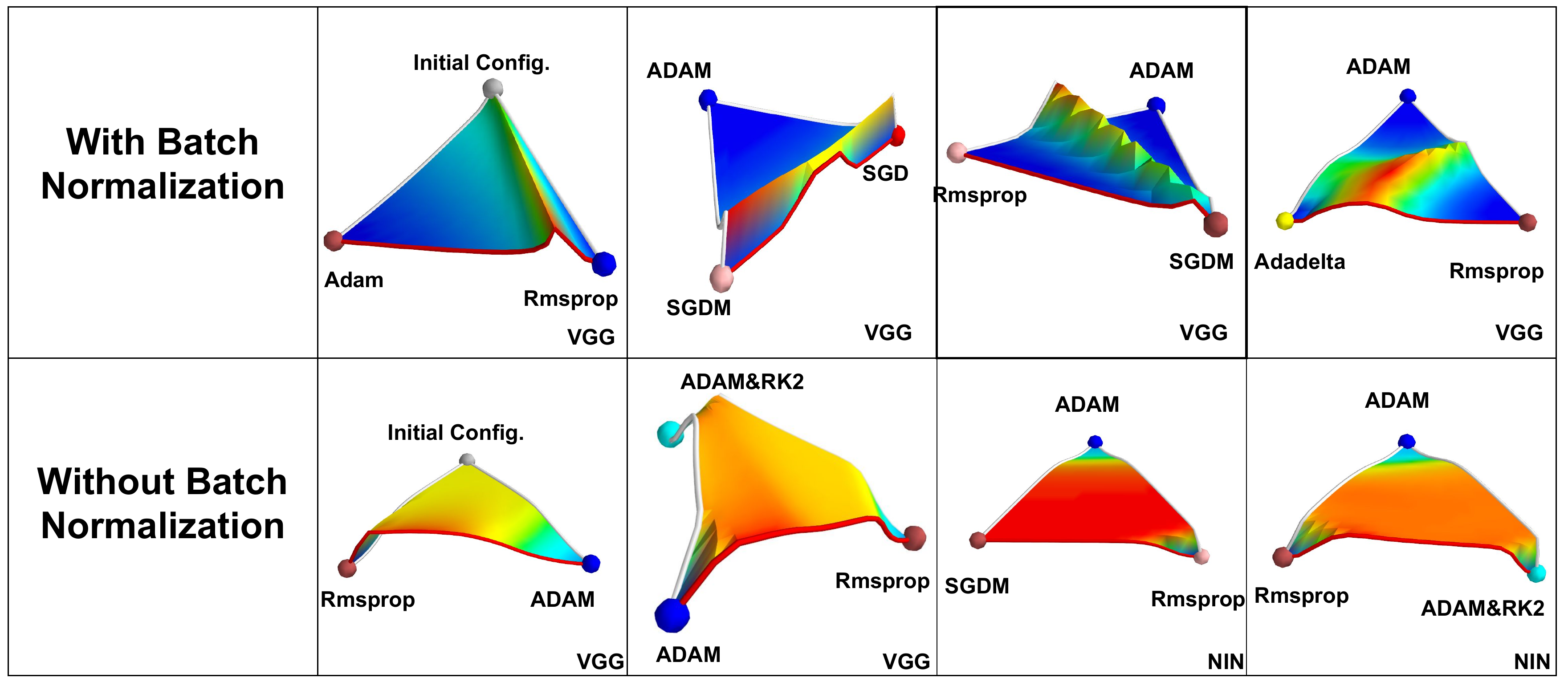}
    \caption{Visualization of the loss surface with and without batch-normalization. }
    \label{fig:batchnorm_zoo}
%\end{wrapfigure}
\vspace{-.2cm}
\end{figure}

To try to understand why some algorithms find final points within bigger basins in the projected space, we looked for characteristic properties of the training process and the final weights found by different algorithms. We found that, for NIN on CIFAR10, Adam-based algorithms result in final weights that are much farther from the initial weight vectors (Fig.~\ref{fig:disttraveled}). When we switch from one optimization algorithm to another, we see strong effects on the distance from the initial to current weight vectors (Fig.~\ref{fig:disttraveled_switch}). Interestingly, when we switch from Adam to SGD (Fig.~\ref{fig:disttraveled_switch}(a)), the weight vectors travel {\em back toward} the initialization. This implies that, for this pair of algorithms, it is not just that the step size is smaller for SGD, but that SGD has a characteristic range of distances from the initial weights. Because the norm of the initial weight vector is so small, this may instead correspond to a characteristic weight vector norm. Fig.~\ref{fig:disttraveled}(c) shows that the distance traveled and weight vector norm curves are nearly identical. Saddle points in deep {\em linear} networks, which have a scale ambiguity, have been shown to correspond to weight vector norm scale~\citep{Saxe2014}. Batch normalization in deep nonlinear networks introduces such a scale ambiguity between the incoming weights and the normalization~\citep{Ba2016}. 

\subsection{Effects of batch-normalization}
\label{sec:BatchNorm}

To understand how batch normalization affects the types of solutions found, we performed a set of experiments comparing loss surfaces near solutions found with and without batch normalization for each of the optimization methods. We visualized the surface near these solutions by interpolating between the initial weights and the final weights as well as between pairs of final weights found with different algorithms (Fig.~\ref{fig:batchnorm_zoo}).

\begin{figure}[htb]
%\begin{wrapfigure}{R}{0.5\textwidth}
  \centering
  \begin{minipage}{0.245\textwidth}
    \centering
    \includegraphics[width=\textwidth]{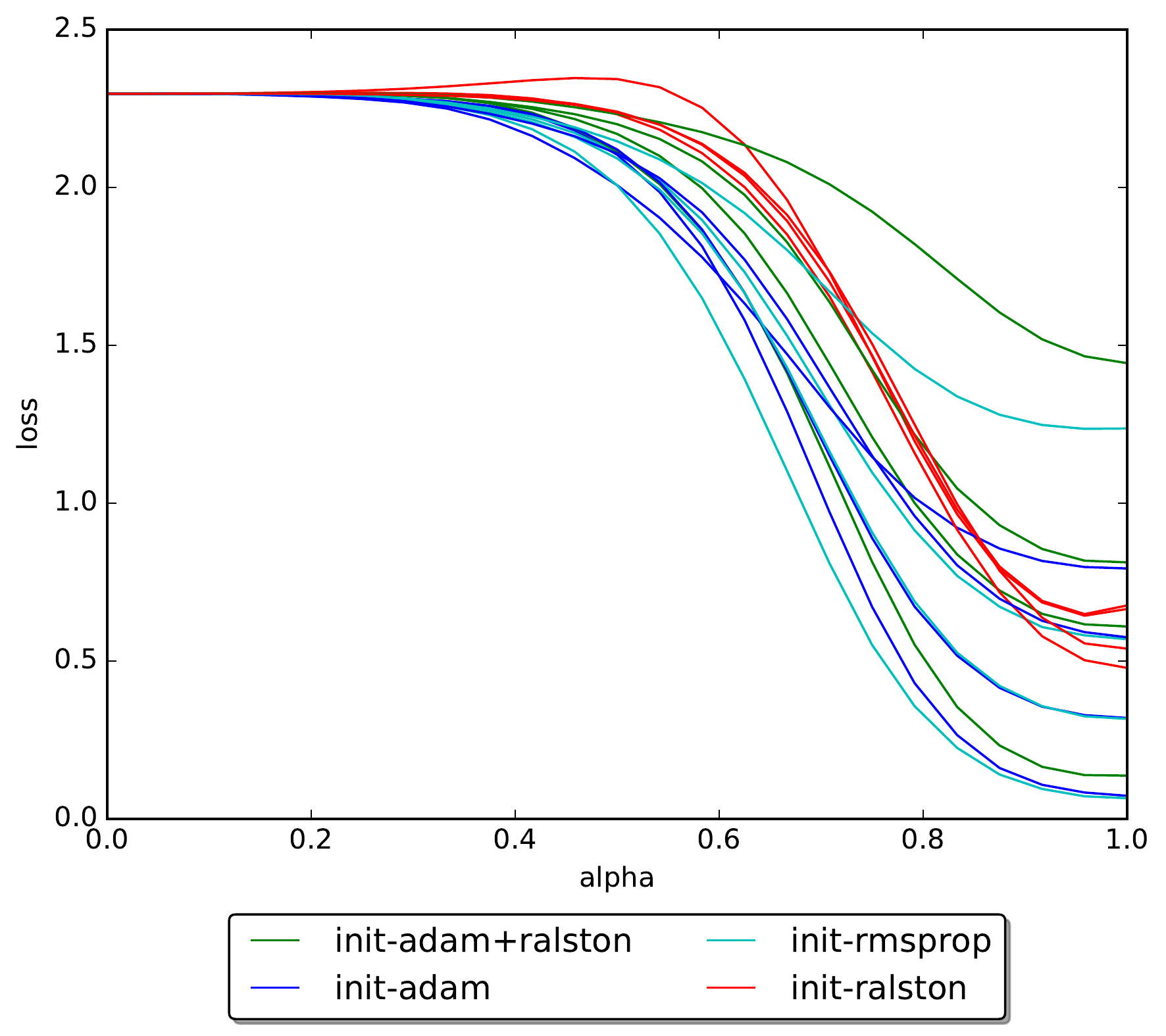}
    \vspace{-.1cm}
    \subcaption{NIN - Initial to Final.}
  \end{minipage}
  \begin{minipage}{0.245\textwidth}
    \centering
    \includegraphics[width=\textwidth]{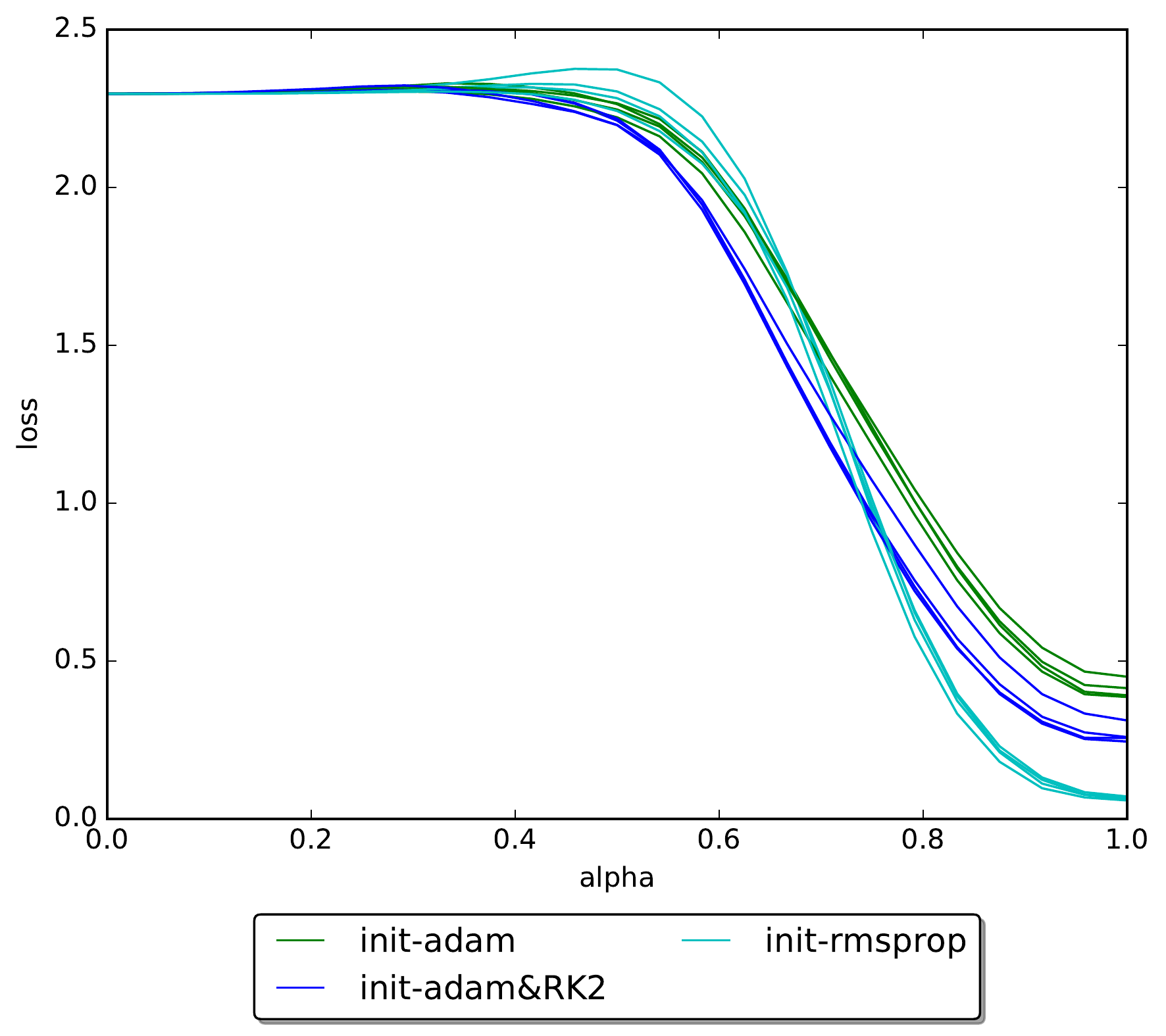}
    \vspace{-.1cm}
    \subcaption{VGG - Initial to Final.}
  \end{minipage}
%  \begin{minipage}{0.329\textwidth}
%    \centering
%    \includegraphics[width=\linewidth]{lstm_nb_repeat_inter_if.pdf}
%    \mysubfigurecaption{LSTM - Initial to Final.}
%  \end{minipage}\\
  \begin{minipage}{0.245\textwidth}
    \centering
    \includegraphics[width=\textwidth]{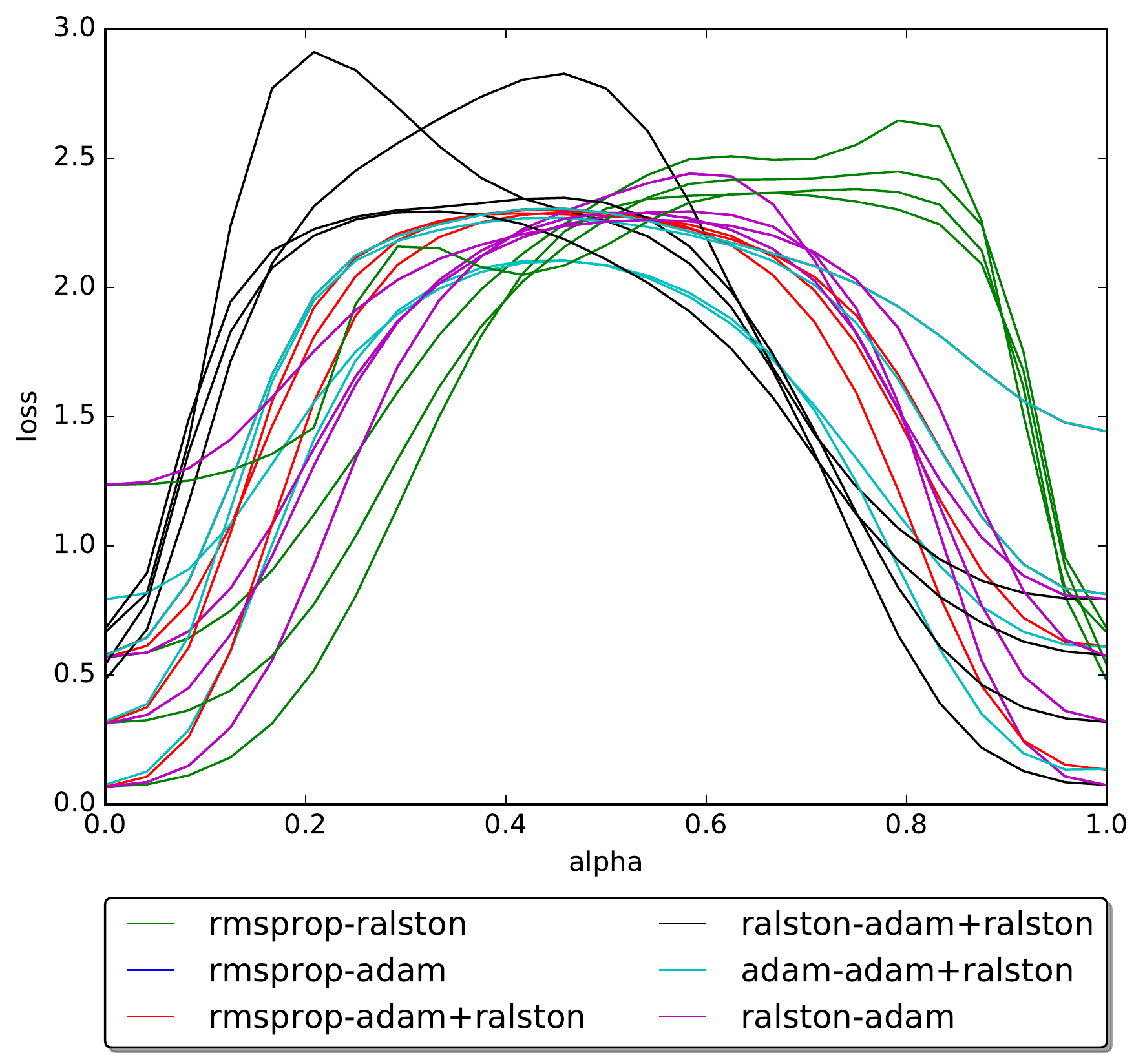}
    \vspace{-.1cm}
    \subcaption{NIN - Final to Final.}
    \end{minipage}
  \begin{minipage}{0.245\textwidth}
    \centering
    \includegraphics[width=\textwidth]{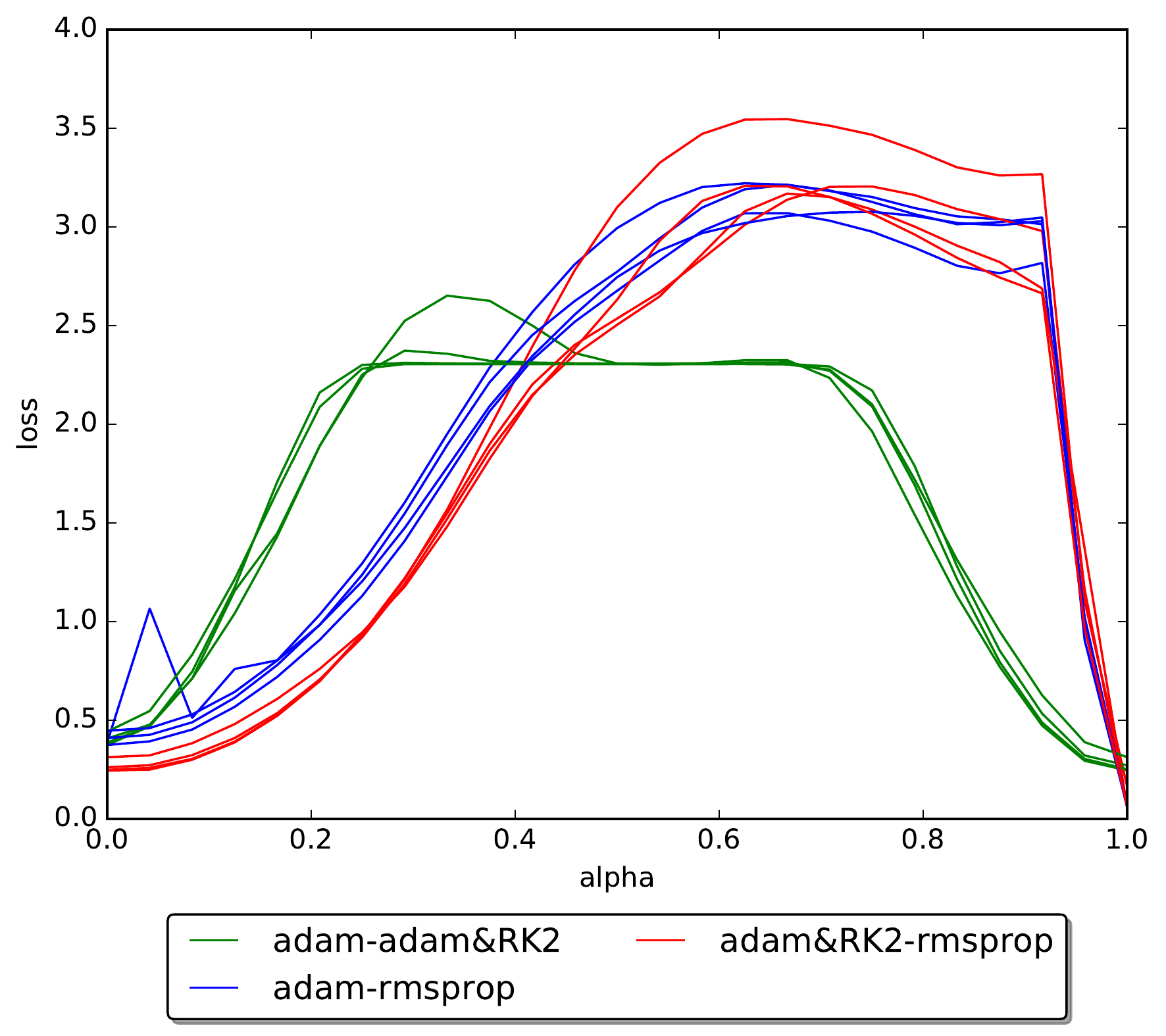}
    \vspace{-.1cm}
    \subcaption{VGG - Final to Final.}
  \end{minipage}
%  \begin{minipage}{0.329\textwidth}
%    \centering
%    \includegraphics[width=\linewidth]{lstm_nb_repeat_inter_ff.pdf}
%    \mysubfigurecaption{LSTM - Final to Final.}
%  \end{minipage}
  \caption{Loss function visualizations for multiple re-runs of each algorithm without batch normalization. Each re-run corresponds to a different initialization. Without batch normalization, re-runs are much less consistent. \label{fig:nobatch_interpolations}}
  \vspace{-.2cm}
%\end{wrapfigure}
\end{figure}

We observed clear qualitative differences between optimization with (Figure~\ref{fig:interpolations}) and without (Figure~\ref{fig:nobatch_interpolations}) batch normalization. We see that, without batch normalization, the quality of the solutions found by a given algorithm is much more dependent on the initialization. In addition, the surfaces between different solutions are more complex in appearance: with batch normalization we see sharp unimodal jumps in performance but without batch normalization we obtain wide bumpy shapes that aren't necessarily unimodal.

%We notice that the structure of the linearly interpolated 1D space are
%similar even though the parameters are initialized differently. 
%In contrast, the loss surface of NIN without batch-normalization looks
%somewhat different. 
%From this results, we speculate that this is due to havin batch 
%normalization, in which it normalizes all the loss surface of neural
%networks. 
%TODO: Similar results has been observed for VGG network as 
%well in Figure~\ref{fig:vgg_inter_supp} and Figure~\ref{?}. 

%%3 different methods 
%\begin{figure}[htp]
%    \centering
%    \begin{minipage}{0.33\textwidth}
%        \centering
%        \includegraphics[width=\linewidth]{VGG_3D_adam_sgdm_sgd.pdf}
%        \caption{SGD-ADAM-SGDM} 
%        \label{fig:vgg_loss_adam_adadelta}
%    \end{minipage}
%    \begin{minipage}{0.33\textwidth}
%        \centering
%        \includegraphics[width=\linewidth]{VGG_3D_adadelta_adam_rmsrpop.pdf}
%        \caption{ADAM-RMSprop-AdaDelta} 
%        \label{fig:vgg_loss_adam_adadelta_rmsprop}
%    \end{minipage}
%    \begin{minipage}{0.32\textwidth}
%        \centering
%        \includegraphics[width=0.99\linewidth]{VGG_3D_rmsprop_adam_sgdm.pdf}
%        \caption{SGDM-ADAM-RMSprop} 
%        \label{fig:vgg_loss_rmsprop_adam_adadelta}
%    \end{minipage}
%    \caption{VGG - The Loss Surfaces from 
%    two found final points by different optimization methods.}
%    \label{fig:2D_rendering_sgdm_rmsprop_adadelta}
%\end{figure}

The neural networks are typically initialized with very small parameter 
values \citep{Glorot2010, He2015}. Instead, we trained NIN with exotic 
intializations such as initial parameters drawn from $\mathcal{N}(-10.0, 0.01)$ 
or $\mathcal{N}(-1.0, 1.0)$ and observe the loss surface behaviours (We used same initialization distribution as \citep{Swirszcz2016}.)
The details of results are discussed in Appendix~\ref{app:initializations}.

\section{Conclusions}

%In this paper, we have performed a series of empirical analyses to understand the geometry of the loss functions corresponding to deep neural networks, and how different optimization methods minimize this loss. 
%Our experiments show that different optimization methods find different minima, even when switching from one method to another very late during training. In addition, we found that the minima found by different optimization methods have different shapes, but that these minima are similar in terms of the most important measure, generalization accuracy. Finally, we compared optimization methods with and without batch normalization, to gain intuition as to how batch normalization may be improving the speed of convergence. 
%
%These points do not seem to be methods falling off of saddle points as the minima found seem to always have similar loss.
%In fact, despite this large number of minima found throughout our experiments, the minima all seem to be qualitatively similar with respect to the loss function.
%We confirmed that these minima appear to represent different functions as well, so we were not converging toward equivalent networks.
%%We also performed some experiments on the effects of batch normalization and found that the loss functions ????

In this work, we have presented and analyzed a series of experiments with neural networks.
The intent of these experiments was to elaborate our understanding on the nature of neural network loss functions, specifically with respect to the geometry of these loss functions and the sensitivities of different optimization methods for minimizing them.

We found that different optimization techniques find different final weights, even when they were seeded by the same initialization point.
Furthermore, we saw that when batch normalization causes the validation error for these final networks are similar. In the projected spaces we chose for visualizing the loss function, any pair of final weights was always separated by a high loss bump. This is suggestive that the different algorithms find different solutions, and thus diverge at saddle points encountered during optimization. We also found that switching from one optimization algorithm to another late in training, when the training loss is decreasing slowly, still resulted in different local minima within the projected space. This is suggestive that there are saddle points even at this late stage in training, at which different optimization algorithms diverge. Our observations suggest that the loss surfaces of deep networks have a large number of saddle points that are encountered during optimization. 

While our observations are consistent with those of Goodfellow et al.~\citep{Goodfellow2015}, they offer a more complex picture. Goodfellow et al. showed that the projected loss between initial and final weights monotonically decreases. They concluded that the loss function shape was simple, and that optimization could proceed without overcoming local minima or saddle points. Instead, we see that, during optimization, many saddle points are encountered even late in optimization, at which points different optimization algorithms choose different descent directions. Our observations are thus more consistent with previous theoretical analyses on simplified networks, which have suggested that deep network loss surfaces have many saddle points and local minima are similar in terms of training and validation error (Sec.~\ref{sec:intro}). Most surprisingly, we see a characteristic shape to the loss function projected around the final weights found by different algorithms, which suggests that different algorithms make consistently different choices at saddle points, which, to our knowledge, is a novel hypothesis. 

While we found evidence that these different final weights are not simply different parameterizations of the same network, we do not yet know if the qualitative differences in final weights found by different algorithms matter in a practical sense, i.e.~in some quality of the predictions of the network, and we hope this is a fruitful direction of future research. Because of the many saddle points encountered during optimization, and the qualitative differences in the shape of the loss function around the final points found by different algorithms, we conclude that measuring the efficiency of different optimization algorithms on the same strictly convex loss function is insufficient for estimating the efficiency of these algorithms on deep network loss surfaces. 

\clearpage
\bibliography{main}
\bibliographystyle{plain}

\clearpage
\appendix

\onecolumn
\section{Supplemantary Materials}
\subsection{3D Visualization} \label{sec:3D_vis}
\citep{Goodfellow2015} introduced the idea of visualizing
1D subspace of the loss surface between the parameters. 
Here, we propose to visualize loss surface in 3D space
through interpolating over three and four vertices. 

\paragraph{Linear Interpolation} 
Given two parameters $\bm \theta_1$ and $\bm \theta_2$,
\begin{equation}
    \bm \theta_i = \alpha \bm \theta_1 + (1-\alpha) \bm \theta_2, \qquad 
    \forall \alpha \in [0,1].
    \label{eqn:linear_intepolation}
\end{equation}

\paragraph{Bilinear Interpolation}
Given four parameters $\bm \theta_0,\ \bm \theta_1,\ \bm \theta_2,$ 
and $\bm \theta_3$,
\begin{align}
    \bm \phi_i    &= \alpha \bm \theta_1 + (1-\alpha) \bm \theta_2\\
    \bm \varphi_i &= \alpha \bm \theta_3 + (1-\alpha) \bm \theta_4\\
    \bm \theta_j  &= \beta  \bm \phi_i   + (1-\beta)  \bm \varphi_i
\end{align}
for all $\alpha \in [0,1]$ and $\beta \in [0,1]$. 

\paragraph{Barycentric Interpolation}
Given four parameters $\bm \theta_0,\ \bm \theta_1$, and $\bm \theta_2$,
let $\bm d_1 = \bm \theta_1 - \bm \theta_0$ 
and $\bm d_2 = \bm \theta_2 - \bm \theta_0$. 
Then, the formulation of the interpolation is 
\begin{align}
    \bm \phi_i    &= \alpha \bm d_1 + \bm \theta_0\\
    \bm \varphi_i &= \alpha \bm d_2 + \bm \theta_0\\
    \bm \theta_j  &= \beta  \bm \phi_i   + (1-\beta)  \bm \varphi_i
\end{align}
for all $\alpha \in [0,1]$ and $\beta \in [0,1]$. 

\subsection{Optimization Methods}\label{sec:background}

\subsubsection{Stochastic Gradient Descent}
In many deep learning applications both the number of parameters and quantity of input data points can be quite large.
This makes the full evaluation of $U(\bm\theta)$ be prohibitively expensive.
A standard technique for aleviating computational loadis to apply an
stochastic approximation to the gradient~\citep{Robbins1951}.
More precisely, one approximates $U$ by a subset
of $n$ data points, denoted by $\{\sigma_j\}_{j=1}^N$ at each
timestep:
\begin{equation}
    U^n(\bm\theta)  = \frac{1}{n} \sum_{j=1}^n \ell(\bm\theta, \bx_{\sigma_j})
    \simeq \frac{1}{N} \sum_{i=1}^N \ell(\bm\theta, \bx_i) = U(\bm\theta)
\end{equation}

Of course this approximation also carries over to the gradient, which is of vital importance to optimization techniques:
\begin{equation}
    \nabla U^n(\bm\theta)  = \frac{1}{n} \sum_{j=1}^n \nabla\ell(\bm\theta, \bx_{\sigma_j})
    \simeq \nabla U(\bm\theta)
    \label{eqn:sto_grad}
\end{equation} 
This method is what is commonly called \textit{Stochastic Gradient Descent} or \textit{SGD}.
So long as the data is distributed nicely the approximation error of $U^n$
should be sufficiently small such that not only will SGD still behave like
normal GD , but it's wall clock time for to converge should be significantly lower as well.

Usually one uses the stochastic  gradient rather than the true gradient, but the inherent noisiness must be kept in mind.
In what follows we will always mean the stochastic gradient.

%In this context we let $f = -\eta\nabla_{\bm\theta} U^n$ where $\eta$ is a scaling parameter.
%With some rearrangement of terms we see
%\begin{equation}
%    \bm{\theta}_{t+1} = \bm{\theta}_{t} -  \Delta_t \eta\nabla_{\bm\theta} U^n
%    \label{eqn:euler_explicit}
%\end{equation}
%where $\Delta_t \eta$ is the \textit{learning rate} and $\nabla_{\bm\theta} U^n$ is the \textit{descent direction}.
\subsubsection{Momentum}
In order to aleviate both noise in the input data as well as noise from stochasticity used in computing quantities one often maintains history of previous evaluations.
In order to only require one extra variable one usually stores variables of the form
\begin{equation}
    \mathbb{E}[F]_{t} = \alpha F_{t} + \beta \mathbb{E}[F]_{t-1}.
\end{equation}
where $F_t$ is some value changing over time and $\mathbb{E}[F]_t$ is the averaged quantity.

An easy scheme to apply this method to is to compute a rolling weighted average of gradients such as 
\[
    \mathbb{E}[\bm g]_t = (1-\alpha) \bm g_t + \alpha \mathbb{E}[\bm g]_{t-1}
\]
but there will be other uses in the future.

%There's of course an
%implicit symmetry in our system as if we scale $\eta$ up but scale
%$\mathcal{X}_t$ down the resulting $\bm\theta_t$ will be identical. This means that one can be somewhat sloppy with the treatment $\eta$.
%In particular one can treat $\eta\mathcal{X}_t$ as a single quantity and tune that quantity instead of just $\mathcal{X}_t$ and several methods do indeed take advantage of this fact.

\subsubsection{Pertinent Methods}
With the aforementioned tools there are a variety of methods that can be constructed.
We choose to view these algorithms as implementations of Explicit Euler on a variety 
of different vector fields to remove the ambiguity between $\eta$ and $\bm g_t$.
We therefore can define a method by the vector field $\mathcal{X}_t$ that explicit 
Euler is applied to with a single $\eta$ that is never changed.

\paragraph{SGD with Momentum (SGDM)}
By simply applying momentum to $\bm{g}_t$ one obtains this stabilized stochastic version of gradient descent:
\begin{equation}
    \mathcal{X}_t = -\mathbb E[\bm g]_t.
        \label{eqn:sgdm}
\end{equation}
This is the most fundamental method that is used in practice and the basis for everything that follows.

\paragraph{Adagrad}
    Adagrad rescales $\mathcal{X}_t$ by summing up the sqaures of all previous gradients in a coefficient-wise fashion:
    \begin{equation} 
        \mathcal{X}_t = -\frac{\bm{g}_t}{\sqrt{\sum^t_{i=1}\bm{g}_i^2 + \epsilon}}.
        \label{eqn:adagrad}
    \end{equation} 
    Here $\epsilon$ is simply set to some small positive value to prevent division-by-zero.
    In the future we will neglect this term in denominators because it is always necessary.

    The concept is to accentuate variations in $\bm g_t$, but because the
    denominator is monotonically nondecreasing over time this method is doomed
    to retard its own progress over time.
    The denominator can also be seen as a form of momentum where $\alpha$ and $\beta$ are both set to $1$.

\paragraph{Rmsprop}
A simple generalization of ADAGrad is to simply allow for $\alpha$ and $\beta$ to be changed from $1$.
In particular one usually chooses a $\beta$ less than $1$, and presumably $\alpha = 1-\beta$.
Thus one arrives at a method where the effects of the distance history are diminished:
    \begin{align} 
        \mathcal{X}_t =& -\frac{\bm{g}_t}{\sqrt{\mathbb{E}[\bm{g}^2]_t}}.
        \label{eqn:rmsprop}
    \end{align} 

\paragraph{Adadelta}
Adadelta adds another term to RMSprop in order to guarantee that the magnitude of $\mathcal{X}$ is balanced with $\bm g_t$~\citep{Zeiler2011}.
More precisely it maintains
    \begin{align} 
        \frac{\mathcal{X}_t}
        {\sqrt{\mathbb{E}[\mathcal{X}^2_t]}} &= -
        \frac{\bm{g}_t}
                         {\sqrt{\mathbb{E}[\bm{g}^2_t]}}
        %\mathcal{X}_t =& -\frac{\sqrt{\mathbb{E}[\mathcal{X}_t^2]}}
        %                 {\sqrt{\mathbb{E}[\bm{g}^2_t]}} \bm{g}_t.
    \end{align} 
    which results in the following vector field:
    \begin{align}
        \mathcal{X}_t =& -\frac{\sqrt{\mathbb{E}[\mathcal{X}_t^2]}}
                         {\sqrt{\mathbb{E}[\bm{g}^2_t]}} \bm{g}_t.
    \end{align}
    and $\eta$ is set to 1.
    %According to \cite{Zeiler2011}, the ``unit correction'' refers to 
    %the cancellation between the unit of the numerator and the denominator.
    %This leaves us the unit of $\bm{g}_t$. This allow us to get rid of
    %learning rate, i.e $\eta$ is always set to 1.
 
\paragraph{ADAM}
By applying momentum to both $\bm g_t$ and $\bm g_t^2$ one arrives at what is called ADAM.
    This is often considered a combination of SGDM + RMSprop,
    \begin{equation} 
        \mathcal{X}_t = c_{t} \frac{\mathbb{E}[\bm{g}]_t}{\sqrt{\mathbb{E}[\bm{g}^2]_t}}.
        \label{eqn:adam}
    \end{equation} 
    $c_{t} = \frac{\sqrt{1-\beta_{2}^t}}{1-\beta_{1}^t} $ 
    is the \textit{initialization bias correction term} with $\beta_{1}, \beta_{2}\in [0,1)$
    being the $\beta$ parameters used in momentum for $\bm g$ and $\bm g^2$ respectively.
    Initialization bias is caused by the history of the momentum variable being initially set to zero.

%We futher dicuss the characteristics of adaptive learning rate methods in
%the Supplementary material~\ref{sec:opt_char}. 

\subsection{Runge Kutta}
\label{sec:rk}
Runge-Kutta methods \citep{Butcher1963} are a broad class of numerical integrators categorized by their truncation error.
Because the ordinary differential equations Runge-Kutta methods solve generalize gradient descent, our augmentation is quite straightforward.
Although our method applies to all explicit Runge-Kutta methods we will only describe second order methods for simplicity.

\begin{wraptable}{r}{5.5cm}
\centering
\caption{The coefficients of various second order 
Runge-Kutta methods \citep{Hairer1987}}
    \label{tab:rk_methods}
\begin{tabular}{| l | c | c | c |}
    \hline
    Method Name & $a_1$ & $a_2$ & $q_1$\\[1ex]
    \hline\hline
    \rule[1.1ex]{0pt}{1ex}
    Midpoint & $0$ & $1$ & $\frac{1}{2}$\\[.1ex]
    \hline
    \rule[1.1ex]{0pt}{1ex}
    Heun& $\frac{1}{2}$ & $\frac{1}{2}$ & $1$\\[.1ex]
    \hline
    \rule[1.1ex]{0pt}{1ex}
    Ralston& $\frac{1}{3}$ & $\frac{2}{3}$ & $\frac{3}{4}$\\[.1ex]
    \hline
    %\rule[1.1ex]{0pt}{1ex}
    %ITB& $\frac{2}{3}$ & $\frac{1}{3}$ & $\frac{3}{2}$\\[.1ex]
    %\hline
\end{tabular}
\vspace{-0.7cm}
\end{wraptable}

The general form of second-order explicit Runge-Kutta on a time-independent vector field is
\begin{align}
    \bm{\theta}_{t+1} &= \bm{\theta}_t + (a_1\bm{k}_1 + a_2\bm{k}_2) h \label{eqn:grt}\\
    \bm{k}_1 &= \mathcal{X}(\bm{\theta}_t) \label{eq:k1}\\
    \bm{k}_2 &= \mathcal{X}(\bm{\theta}_t + q_1h \bm{k}_1 )\label{eq:k2}
\end{align}
%\begin{align}
%    \bm{\theta}_{t+1} &= \bm{\theta}_t + (a_1\bm{k}_1 + a_2\bm{k}_2) h \label{eqn:grt}\\
%    \bm{k}_1 &= \nabla U(\bm{\theta}_t) \\
%    \bm{k}_2 &= \nabla U(\bm{\theta}_t + q_1 \bm{k}_1 h; t + p_1h, )
%\end{align}
where $a_1, a_2$, and $q_1$ are parameters that define a given Runge-Kutta method.
Table~\ref{tab:rk_methods} refers to the parameters used for the different Runge-Kutta variants we use in our experiments.

\subsubsection{Augmenting Optimization with Runge Kutta}
For a given timestep, explicit integrators can be seen as a morphism over vector fields  $\mathcal{X} \rightarrow \mathcal{\bar X}^{h}$.
For a gradient $\bm g_t = \nabla_{\bm \theta} U$ we can solve a modified RK2 gradient $\bar{\bm g}_t$ in the following fashion:
\begin{align}
    \bm \theta_{t+1} = \bm \theta_{t} + \bar{\bm g}_t h &= Advect^{rk2}_{\bm g}(\bm\theta,h)
\end{align}
rearranged with respect to $\bar {\bm g}_t$
\begin{align}
    \bar{\bm g}_t &= \frac{Advect^{rk2}_{\bm g}(\bm\theta,h) - \theta_{t}}{h} \\
    &= \frac{\bm \theta_t +(a_1\bm{k}_1 + a_2\bm{k}_2)h - \theta_{t}}{h} \\
    &= (a_1\bm{k}_1 + a_2\bm{k}_2).
\end{align}
If we simply substitute the gradient $\bm g_t$ with $\bar{ {\bm g}}_t$ one obtains an RK2-augmented optimization technique.

\subsection{Network architecture and data set details}
\label{sec:supp:networkanddatadetails}

We used the VGG and NIN implementations from {\em https://github.com/szagoruyko/cifar.torch.git}.
As well, we used two layer full connected neural network (FC) with 50 hidden units and batch normalization on each layer.
Table~\ref{table:architectur_table} presents experiment status for each figure.

The batch size was set to 128 and the number of epochs was set to 100 for FC and 200 for NIN and VGG.
The FC parameters were ininitialized using Xavier initalization, $\mathcal{U}(\frac{-\sqrt{6}}{N_{in}+N_{out}}, \frac{\sqrt{6}}{N_{in}+N_{out}})$ where $N_{in}$ and $N_{out}$ are input and output dimension of each layer.
NIN and VGG were intialized from $\mathcal{N}(0,0.05)$.
NIN had $50\%$ dropouts and VGG had $40\%$ dropouts on convolutional layer and $50\%$ dropouts for fully connected layers during the training.
The learning rate was chosen from the discrete range between 
$[0.2, 0.1, 0.05,0.01]$ for SGD and 
$[0.002, 0.001, 0.0005, 0.0001]$ for adaptive learning methods\footnote{Learning rate was fixed for different optimization methods. When we switched
one method to another, we switched the learning rate accordingly. We also
experimented the learning rates that were applied after switching. 
For example, we tried with 0.001, 0.0001, 0.00001 for ADAM. We'll make this more explicit in our Experimental results section.}.
(Table~\ref{table:lr_table} presents the learning rate of the optimizers
that was used for different experiments.)
We doubled the learning rates when we ran our augmented versions with 
Runge-Kutta because they required two stochastic gradient computations per epoch.
We used batch-normalization and dropout to regularize our networks.
For all parts of the optimization algorithms that require randomness, we use the same random choices: we used the same initializations, dropout masks, and sequences of mini-batch samples across algorithms. We'll make this explicit in our Experimental setup section.
%``SGD x2'' is the stochastic gradient descent with twice of the learning 
%rate of ``SGD''.
All experiments were run on a 6-core Intel(R) Xeon(R) CPU @ 2.40GHz with a TITAN X.

\begin{table}
\begin{center}
    \caption{The Neural Network architecture status for different experiments.} 
    \label{table:architectur_table}
\begin{tabular}{ |c|cccc|}
 \hline
    Figure Ref. & Network Type & Batch-Norm. & data & \# of Epoch\\
 \hline \hline
    Fig. \ref{fig:function_diff} \& \ref{fig:opt_switch_mnist} & 2-layer NN & \text{\sffamily X} & MNIST & 100\\
    Fig. \ref{fig:quad_render2}, \ref{fig:quad_render}, \ref{fig:interpolations} \& \ref{fig:acc_curves} 
                                          & NIN \& VGG & \checkmark &  CIFAR10 & 200\\
    Fig. \ref{fig:MinimumComparison}, \ref{fig:nin_switch}, \ref{fig:basinsize} \& \ref{fig:nin_switch_full}
                                         & NIN & \checkmark & CIFAR10  & 200\\
    Table \ref{fig:batchnorm_zoo}  & NIN \& VGG & Both & CIFAR10  & 200 w. BN \& 600 w.o. BN\\
    Fig. \ref{fig:nobatch_interpolations} & NIN, VGG, \& LSTM & \text{\sffamily X} & CIFAR10 & 600\\
    Fig. \ref{fig:exotic_init}  & NIN & Both & CIFAR10 & 200 w. BN \& 600 w.o. BN\\
    Fig. \ref{fig:vgg_switch_sgd_adam}, \ref{fig:vgg_switch_sgd_adadelta} \& \ref{fig:vgg_switch_adam_adadelta} 
                                & VGG & \checkmark & CIFAR10 & 200 \\
    \hline
\end{tabular}
\end{center}
\end{table}

\begin{table}
\begin{center}
    \caption{Learning rate of the optimizers that is used for various experiments.} 
    \label{table:lr_table}
\begin{tabular}{ |c|c|c|}
 \hline
    Learning rate ($\eta$) & Optimizer & Figures \\
 \hline \hline
    -      & Adadelta       & Fig.~\ref{fig:MinimumComparison},~\ref{fig:basinsize},
                                  ~\ref{fig:vgg_switch_sgd_adadelta}, ~\ref{fig:vgg_switch_adam_adadelta}\\ 
    0.1    & SGD, SGDM      & Fig.~\ref{fig:quad_render2},~\ref{fig:MinimumComparison},
                                  ~\ref{fig:interpolations},~\ref{fig:basinsize},
                                  ~\ref{fig:nin_switch}(b),~\ref{fig:nin_switch}(c),
                                  ~\ref{fig:nobatch_interpolations},~\ref{fig:acc_curves},
                                  ~\ref{fig:function_diff},
                                  ~\ref{fig:opt_switch_mnist},~\ref{fig:basinsize},
                                  ~\ref{fig:nin_switch_full},~\ref{fig:vgg_switch_sgd_adam}(a),
                                  ~\ref{fig:vgg_switch_adadelta_large}\\
    0.05   & SGD, SGDM      & Fig.~\ref{fig:nin_switch_full},~\ref{fig:vgg_switch_sgd_adam}(b),
                                  ~\ref{fig:vgg_switch_adadelta_medium}\\
    0.01   & SGD, SGDM      & Fig.~\ref{fig:nin_switch}(a),~\ref{fig:nin_switch}(d),
                                  ~\ref{fig:exotic_init},~\ref{fig:nin_switch_full},
                                  ~\ref{fig:vgg_switch_sgd_adam}(c), ~\ref{fig:vgg_switch_adadelta_small},\\
    0.001  & ADAM, RMSProp  & Fig.~\ref{fig:nin_switch}(a),~\ref{fig:nin_switch}(c),~\ref{fig:nobatch_interpolations},
                                  ~\ref{fig:opt_switch_mnist},~\ref{fig:nin_switch_full},
                                  ~\ref{fig:vgg_switch_sgd_adam}(a),~\ref{fig:vgg_switch_adam_adadelta}\\
    0.0005 & ADAM, RMSProp  & Fig.~\ref{fig:nin_switch_full},~\ref{fig:vgg_switch_sgd_adam}(b)\\
    0.0001 & ADAM, RMSProp  & Fig.~\ref{fig:quad_render2},~\ref{fig:MinimumComparison},
                                  ~\ref{fig:interpolations},
                                  ~\ref{fig:basinsize}, ~\ref{fig:nin_switch}(b),
                                  ~\ref{fig:nin_switch}(d),~\ref{fig:acc_curves},
                                  ~\ref{fig:function_diff},
                                  ~\ref{fig:basinsize},~\ref{fig:nin_switch_full},
                                  ~\ref{fig:vgg_switch_sgd_adam}(c)\\   

    %Fig. \ref{fig:interpolations_mnist}, \ref{fig:function_diff} \& \ref{fig:opt_switch_mnist} \\
    %Fig. \ref{fig:quad_render2}, \ref{fig:quad_render}, \ref{fig:interpolations} \& \ref{fig:acc_curves} \\
    %Fig. \ref{fig:TrainAndValidAccuracy}, \ref{fig:nin_switch}, \ref{fig:interpolation_alpha2} \& \ref{fig:nin_switch_full}
    %                                     & NIN & \checkmark & CIFAR10  \\
    %Table \ref{fig:batchnorm_zoo}  & NIN \& VGG & Both & CIFAR10 \\
    %Fig. \ref{fig:nobatch_interpolations} & NIN, VGG, \& LSTM & \text{\sffamily X} & & CIFAR10 \\
    %Fig. \ref{fig:exotic_init}  & NIN & Both & CIFAR10 \\
    %Fig. \ref{fig:vgg_switch_sgd_adam}, \ref{fig:vgg_switch_sgd_adadelta} \& \ref{fig:vgg_switch_adam_adadelta} 
    %                            & VGG & \checkmark & CIFAR10 \\
    \hline
\end{tabular}
\end{center}
\end{table}
\pagebreak

\subsection{Experiments with Runge-Kutta integrator}
\begin{figure}[htp]
    \begin{minipage}{0.50\textwidth}
        \centering
        \includegraphics[width=\linewidth]{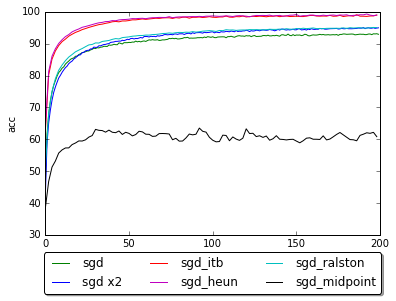}
        \vspace{-0.5cm}
        \mysubfigurecaption{NIN - SGD \& RK2}
        %\label{fig:nin_inter_render}
    \end{minipage}
    \begin{minipage}{0.49\textwidth}
        \centering
        \includegraphics[width=\linewidth]{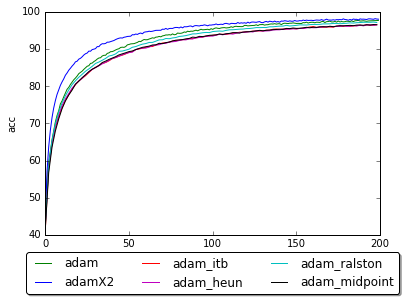}
        \vspace{-0.5cm}
        \mysubfigurecaption{VGG - SGD \& RK2} 
        %\label{fig:nin_inter_render}
    \end{minipage}\\
    \begin{minipage}{0.50\textwidth}
        \centering
        \includegraphics[width=\linewidth]{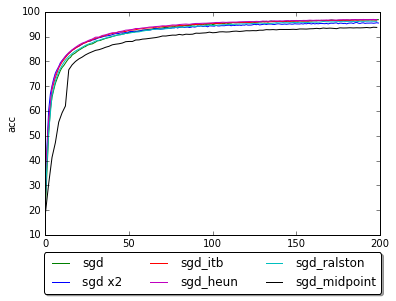}
        \vspace{-0.5cm}
        \mysubfigurecaption{NIN - ADAM \& ADAM+RK2} 
        %\label{fig:nin_inter_render}
    \end{minipage}
    \begin{minipage}{0.49\textwidth}
        \centering
        \includegraphics[width=\linewidth]{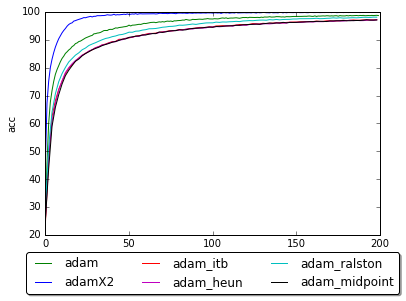}
        \vspace{-0.5cm}
        \mysubfigurecaption{VGG - ADAM \& ADAM+RK2} 
        %\label{fig:nin_inter_render}
    \end{minipage}
    \caption{Training accuracy curve}
    \label{fig:acc_curves}
\end{figure}

The results in Figure~\ref{fig:acc_curves} illustrates 
that, with the exception of the Midpoint method, 
stochastic Runge-Kutta methods outperform SGD.
``SGD x2'' is the stochastic gradient descent with 
twice of the learning rate of ``SGD''. From the figure,
we observe that the Runge-Kutta methods perform even 
better with half the number of gradient computed by SGD.
The reason is because SGD has the accumulated truncated 
error of $O(h)$ while second-order Runge-Kutta methods 
have the accumulated truncated error of $O(h^2)$.

Unfortunately, ADAM outperforms ADAM+RK2 methods. We 
speculate that this is because the way how ADAM's 
renormalization of input gradients in conjunction with 
momentum eliminates the value added by using our 
RK-based descent directions.

\pagebreak

\subsection{The results on Two layer Fully Connected Neural Network on MNIST dataset}
\begin{figure}[htb]
  \vspace{-.2cm}
  \centering
  \begin{minipage}{0.49\linewidth} 
    \centering
    \includegraphics[width=\textwidth]{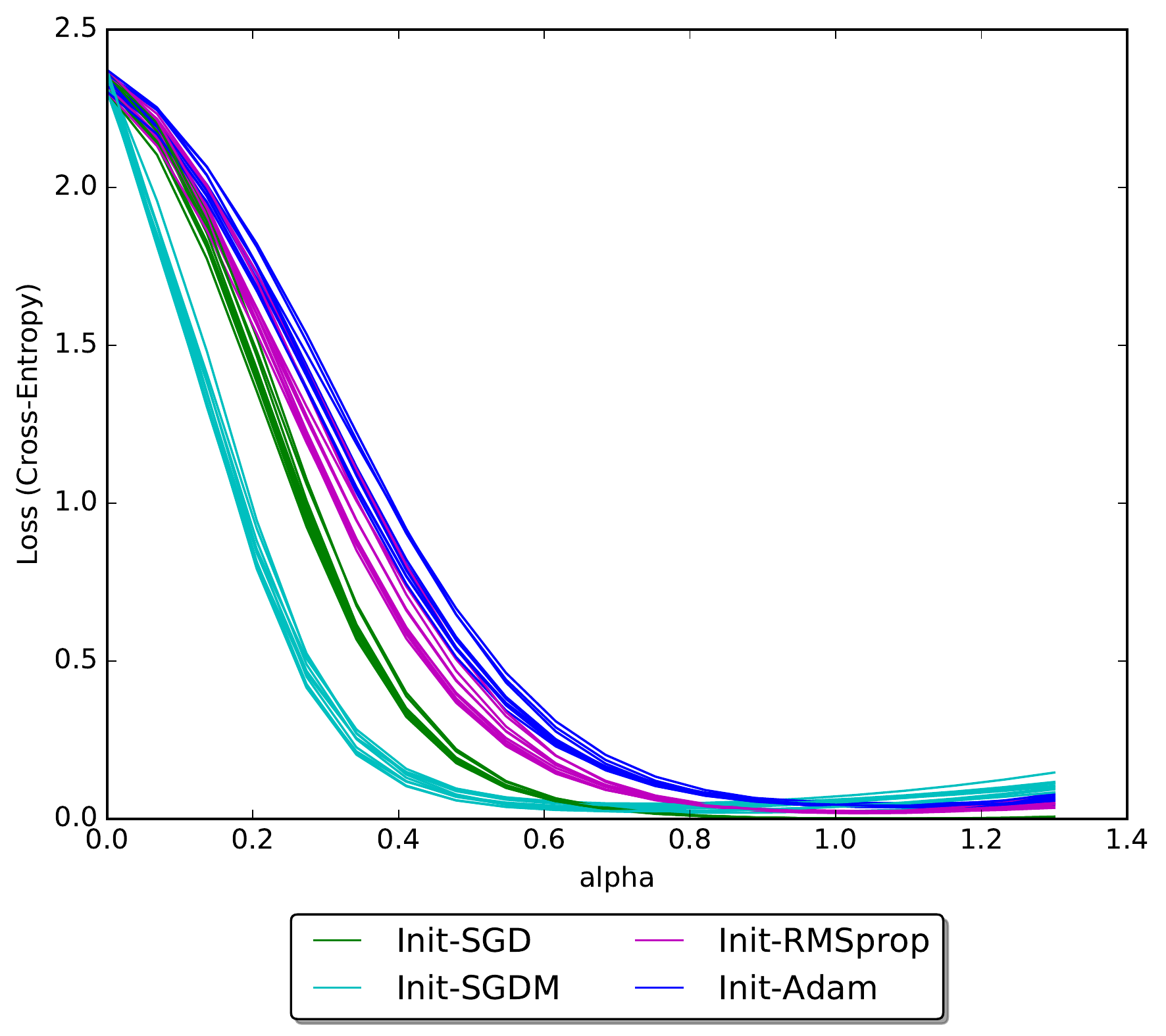}
    \subcaption{MNIST, FC2, Initial to Final}
  \end{minipage}
  \begin{minipage}{0.49\linewidth}
    \centering
    \includegraphics[width=\textwidth]{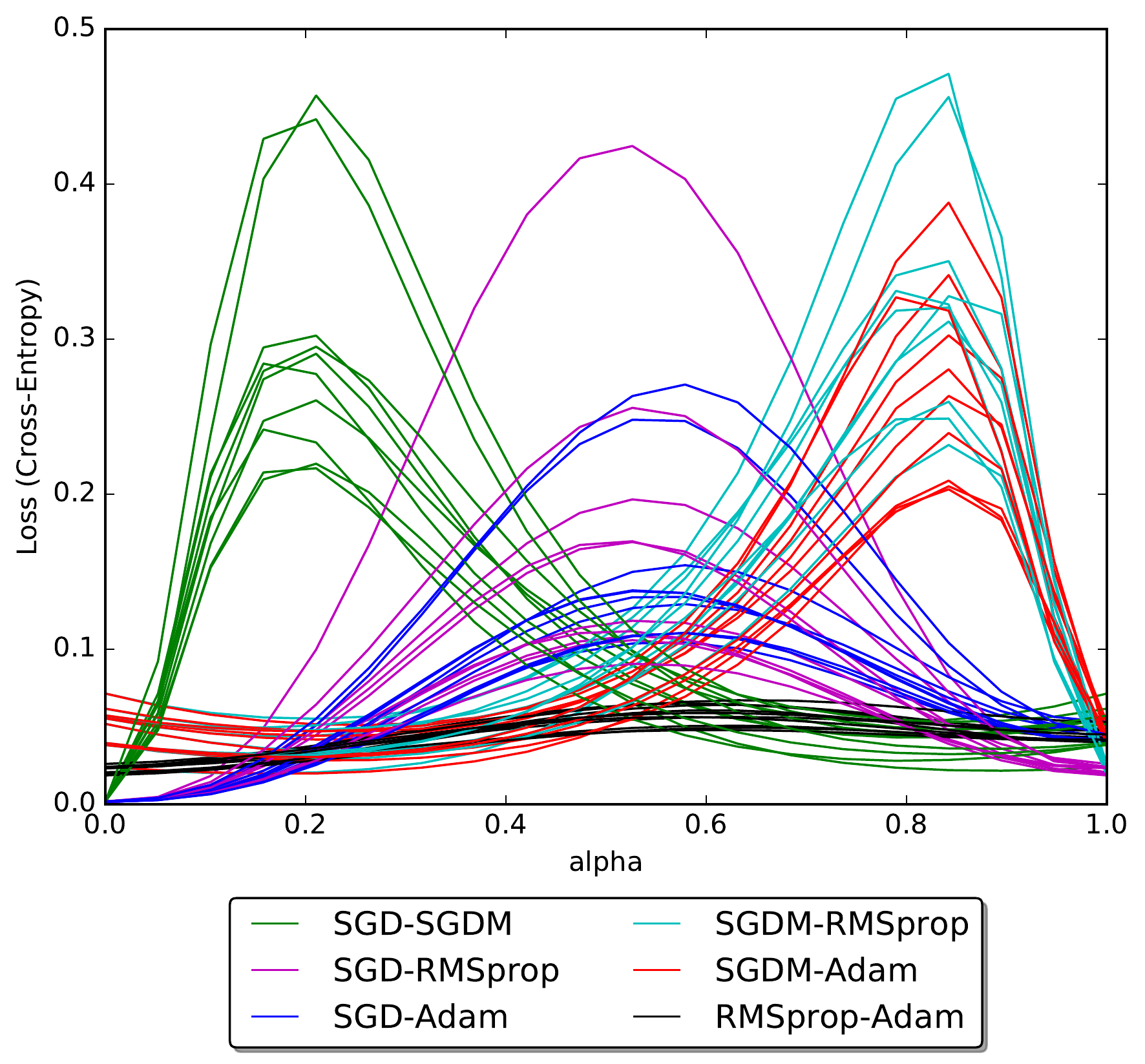}
    \subcaption{MNIST, FC2, Final to Final}
  \end{minipage}
  \caption{Loss function visualizations for multiple re-runs of each algorithm. Each re-run corresponds to a different initialization. We see that the loss function near the final point for a given algorithm has a characteristic geometry. \label{fig:mnist_interpolations}}
  \vspace{-.1cm}
\end{figure}

%\begin{figure}[htp]
%    \centering
%    \begin{minipage}{0.245\textwidth}
%        \includegraphics[width=\linewidth]{mnist_nn_tr_acc_opt_10folds.pdf}
%        \vspace{-0.5cm}
%        \caption{NN - Train Accuracy.}
%    \end{minipage}
%    \begin{minipage}{0.245\textwidth}
%        \includegraphics[width=\linewidth]{mnist_nn_vl_acc_opt_10folds.pdf}
%        \vspace{-0.5cm}
%        \caption{NN - Valid. Accuracy.}
%    \end{minipage}
%    \vspace{-0.25cm}
%    \caption{Loss function value near local minima found 
%             by multiple restarts of each algorithm.}
%    \label{fig:interpolations_mnist}
%\end{figure}

\begin{figure}[htp]
    \centering
    \begin{minipage}{0.245\textwidth}
        \includegraphics[width=\linewidth]{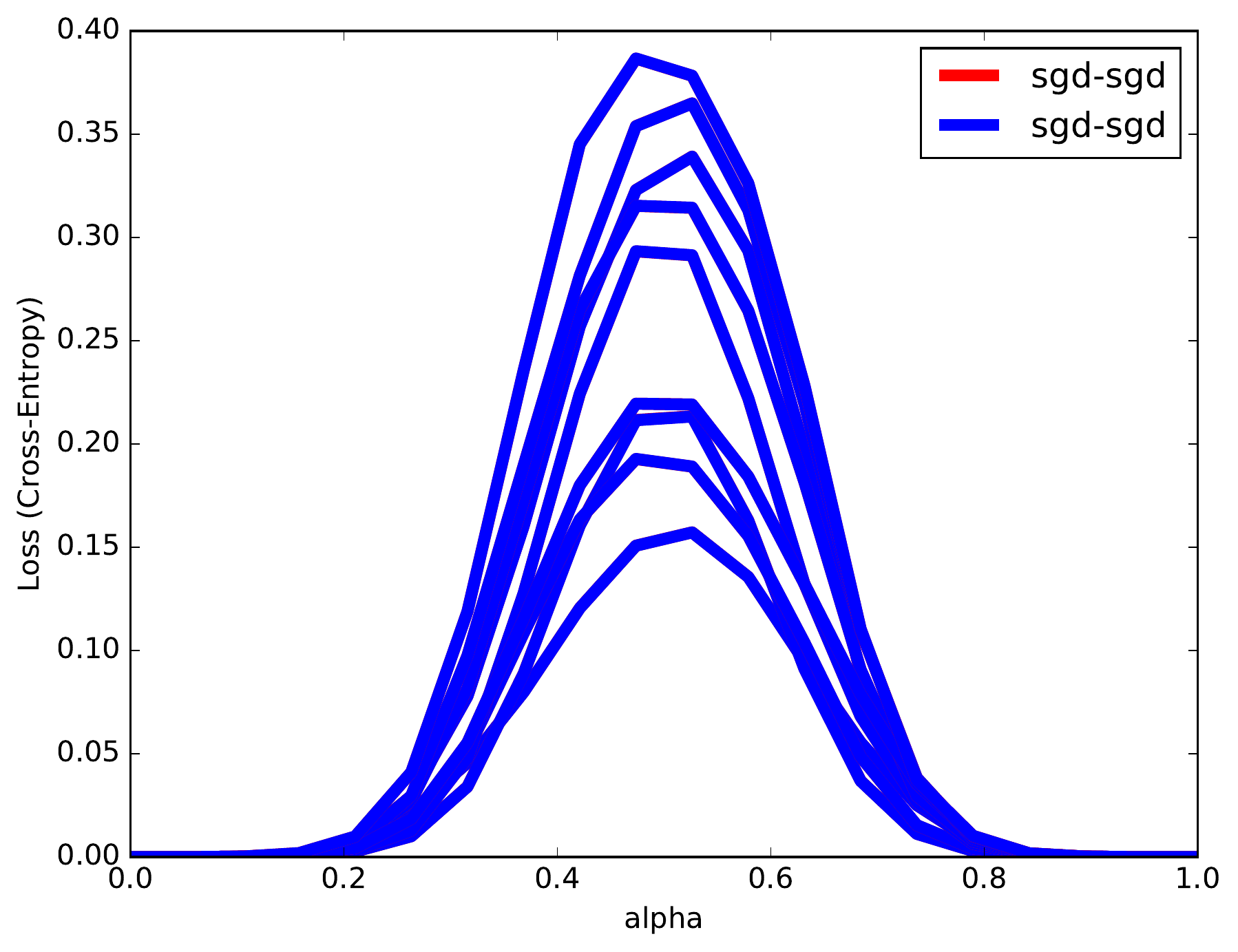}
        \centering \mysubfigurecaption{Functional Difference between - SGD Final to SGD Final.}
    \end{minipage}
    \begin{minipage}{0.245\textwidth}
        \includegraphics[width=\linewidth]{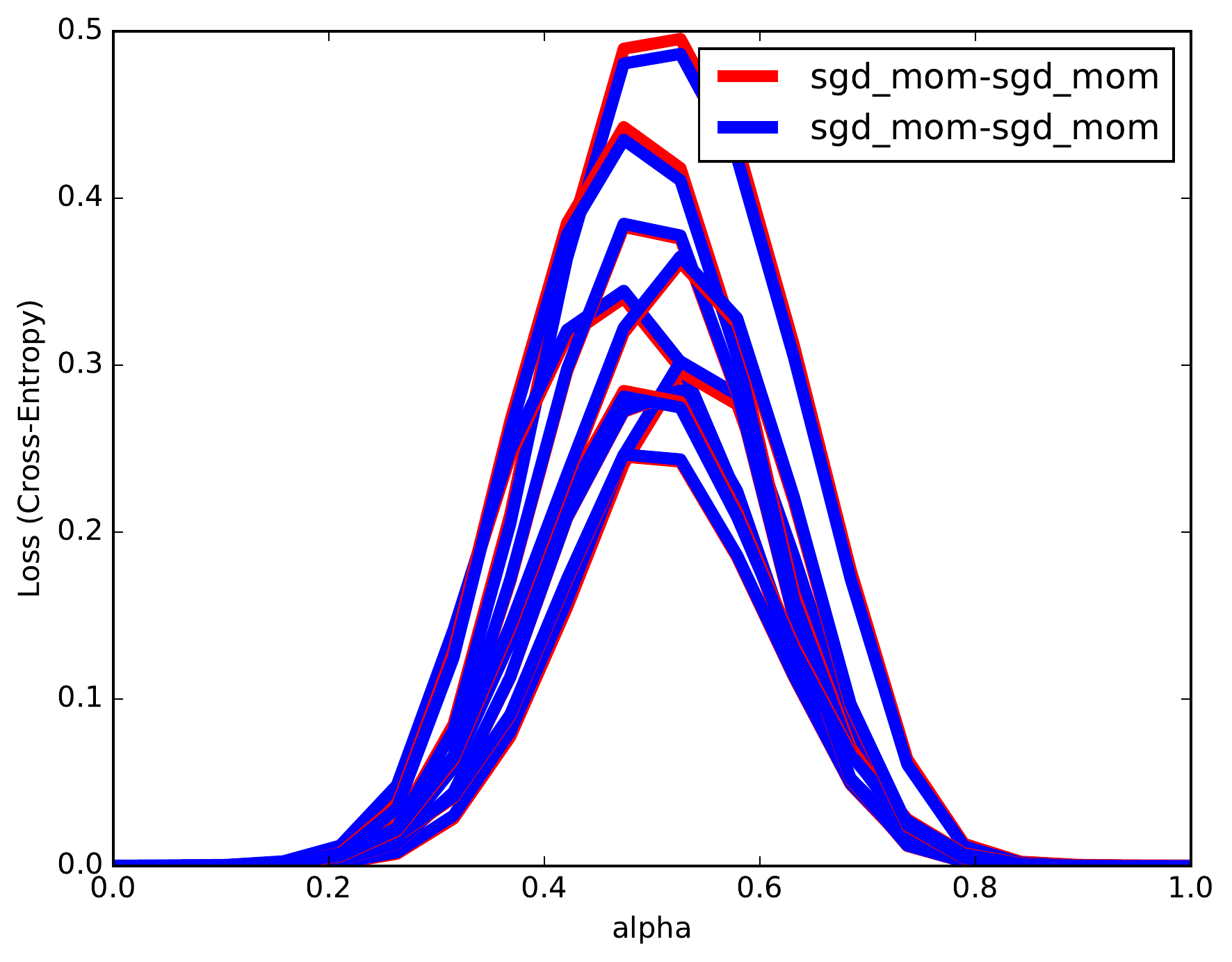}
        \centering \mysubfigurecaption{Functional Difference between - SGDM Final to SGDM Final.}
    \end{minipage}
    \begin{minipage}{0.245\textwidth}
        \includegraphics[width=\linewidth]{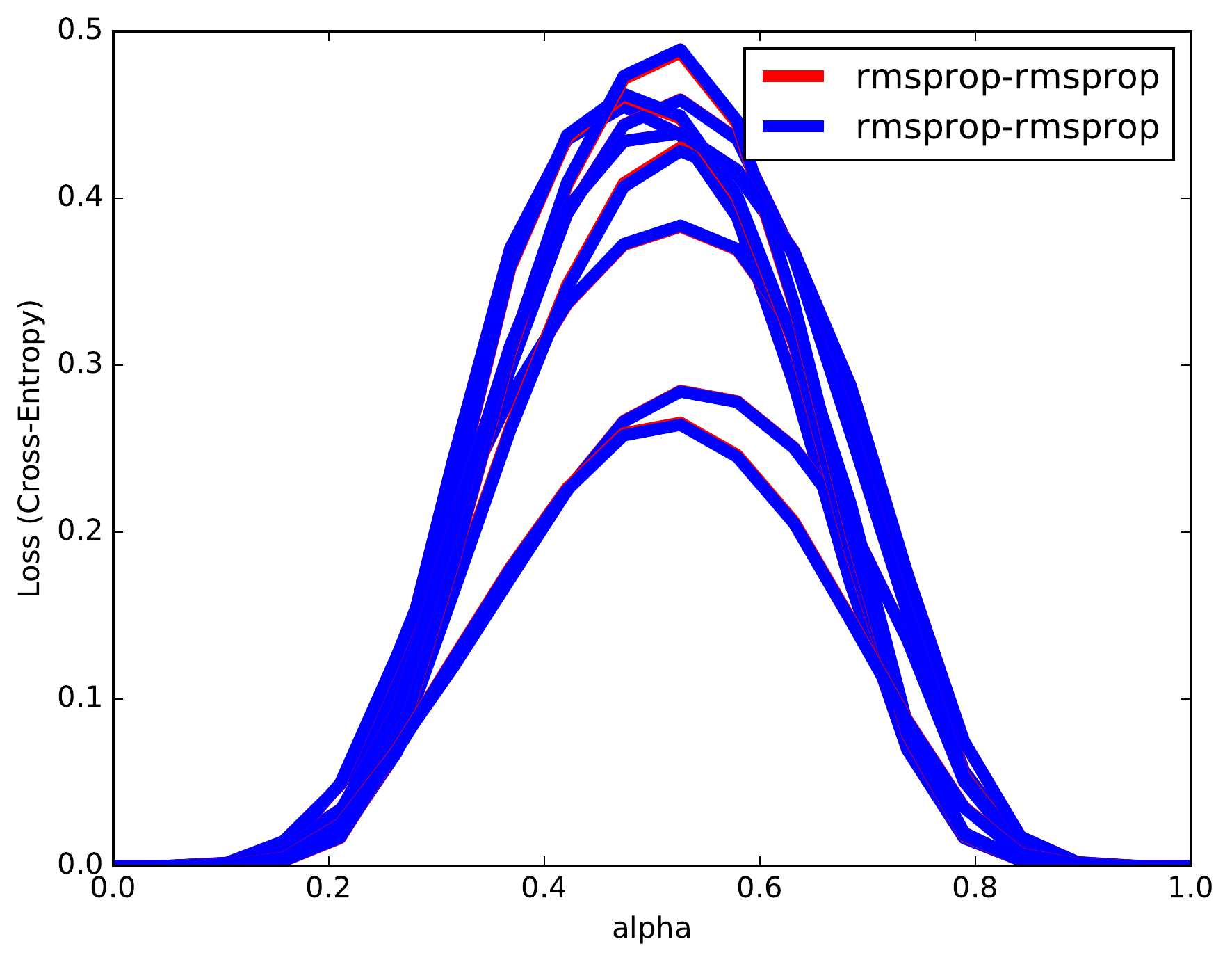}
        \centering \mysubfigurecaption{Functional Difference between - RMSprop Final to RMSprop Final.}
    \end{minipage}
    \begin{minipage}{0.245\textwidth}
        \includegraphics[width=\linewidth]{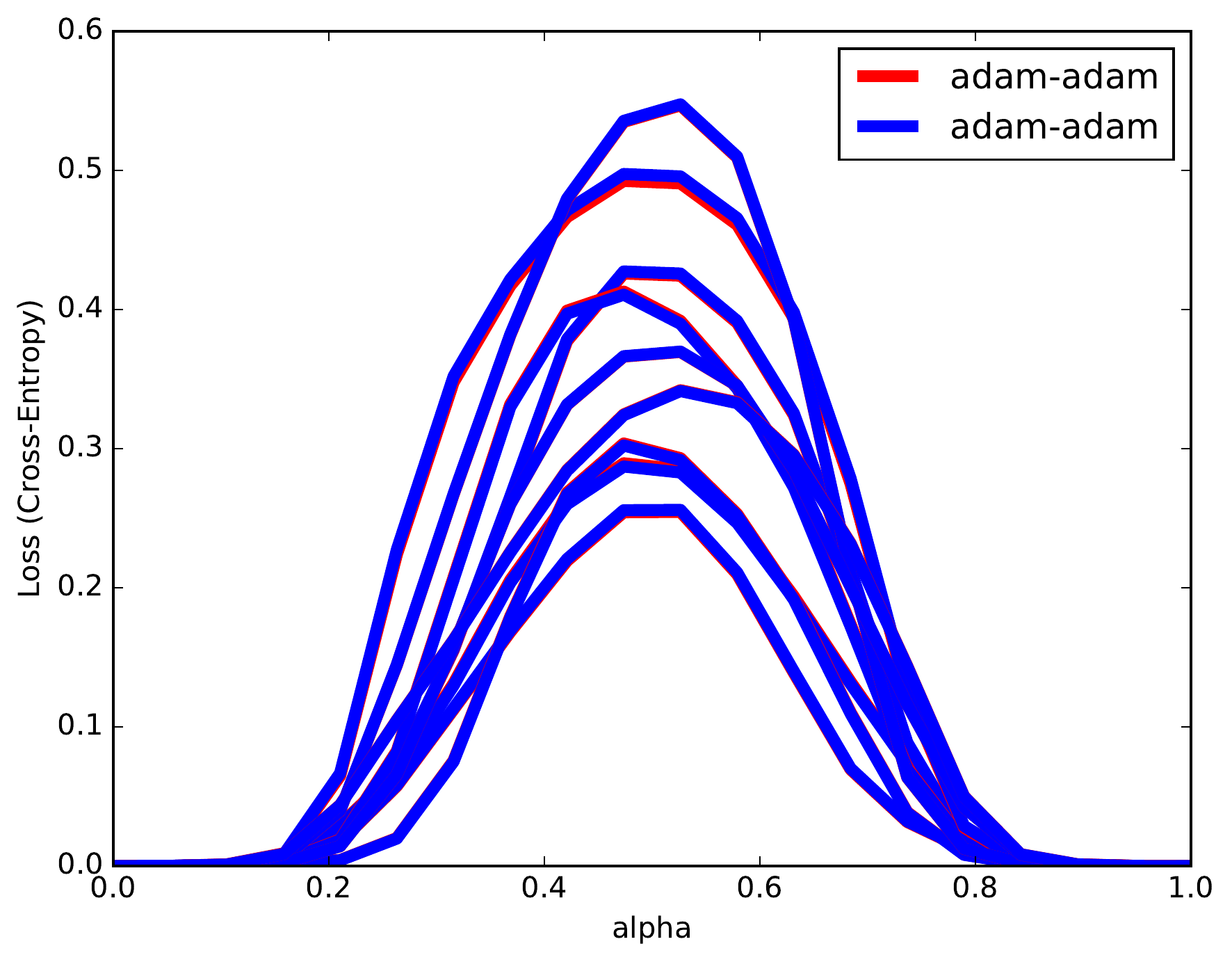}
        \centering \mysubfigurecaption{Functional Difference between - ADAM Final to ADAM Final.}
    \end{minipage}\\
    \begin{minipage}{0.32\textwidth}
        \includegraphics[width=\linewidth]{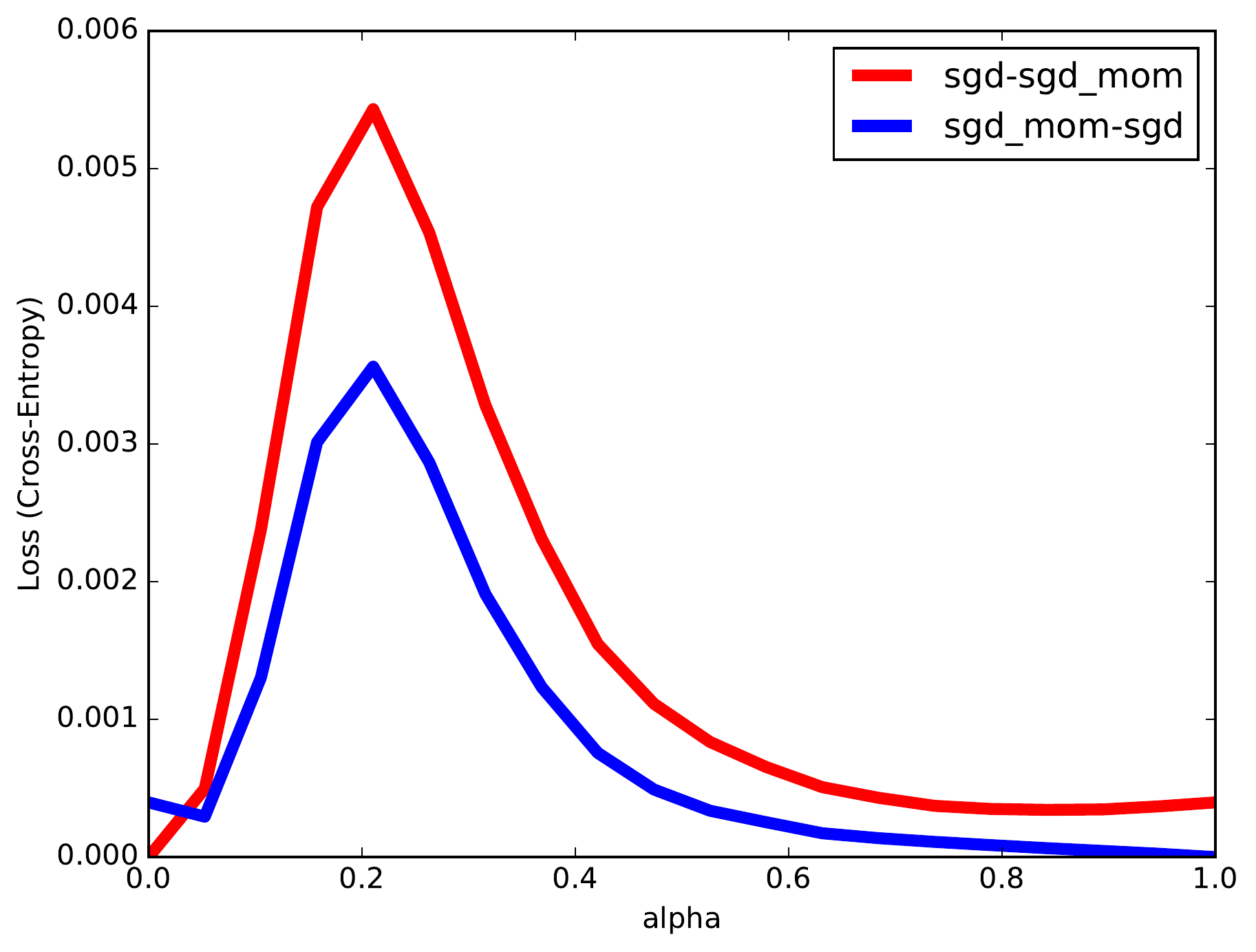}
        \centering \mysubfigurecaption{SGD to SGDM }
    \end{minipage}
    \begin{minipage}{0.32\textwidth}
        \includegraphics[width=\linewidth]{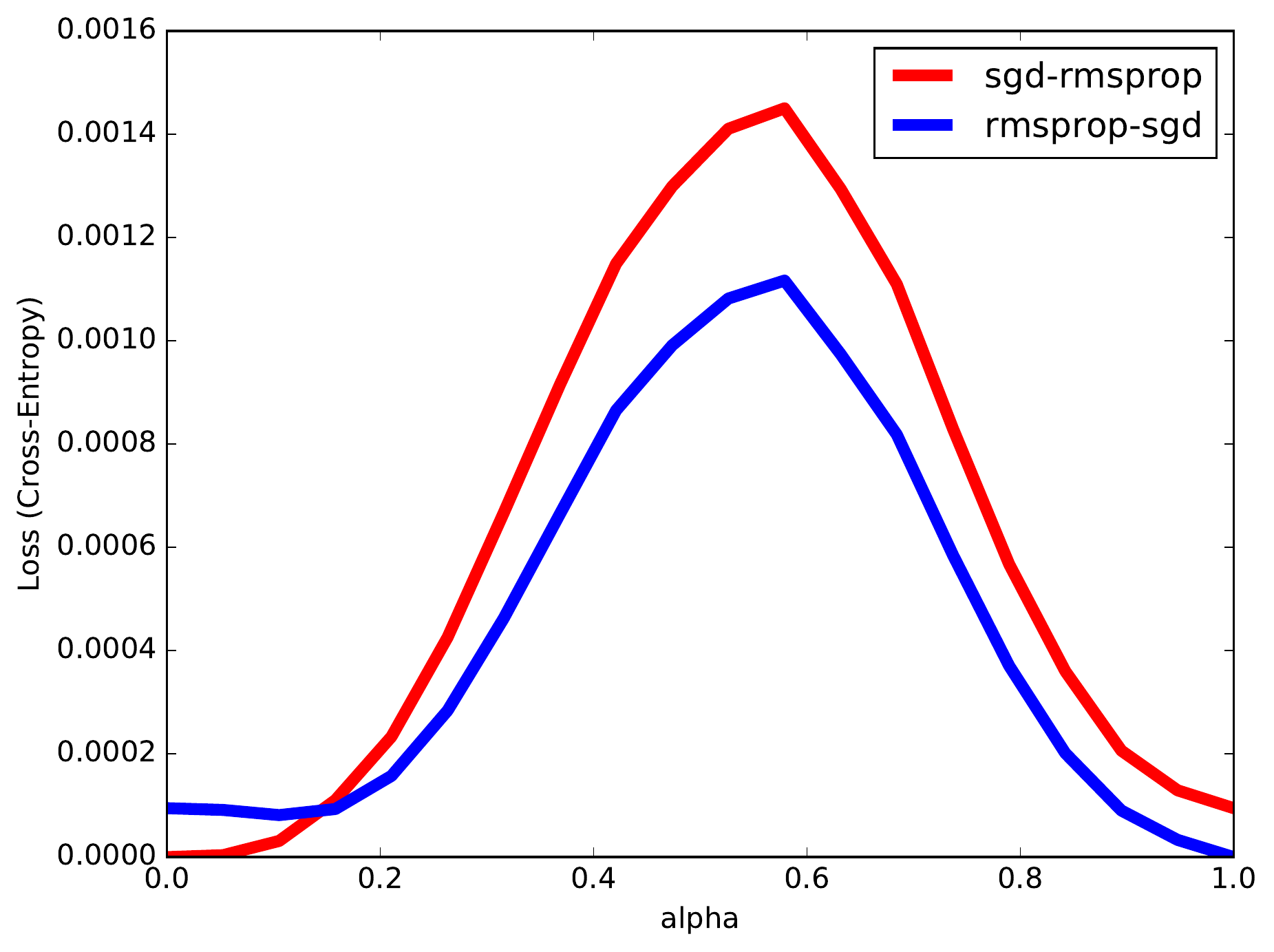}
        \centering \mysubfigurecaption{SGD to RMSprop }
    \end{minipage}
    \begin{minipage}{0.32\textwidth}
        \includegraphics[width=\linewidth]{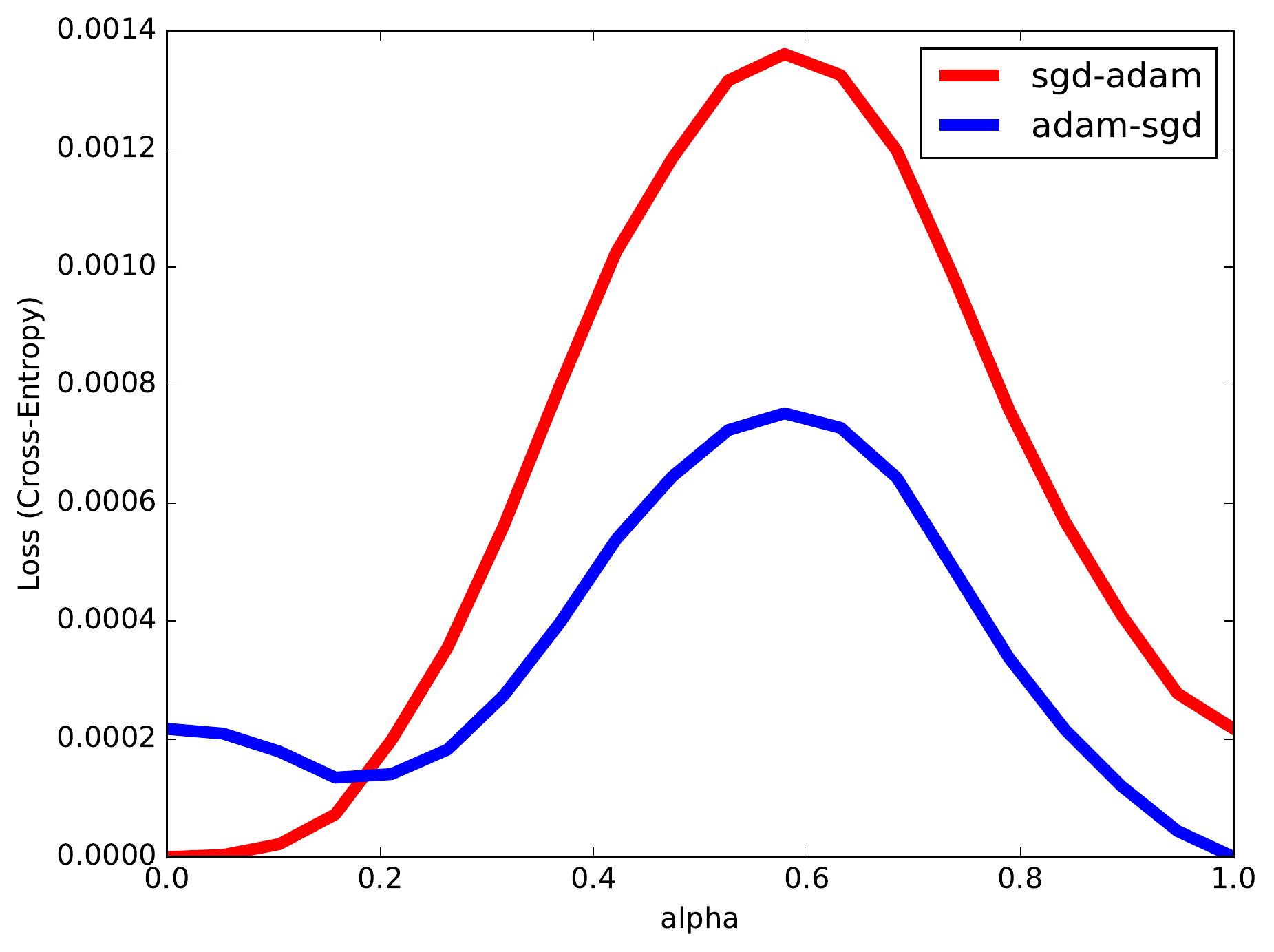}
        \centering \mysubfigurecaption{SGD to ADAM }
    \end{minipage}\\
    \begin{minipage}{0.32\textwidth}
        \includegraphics[width=\linewidth]{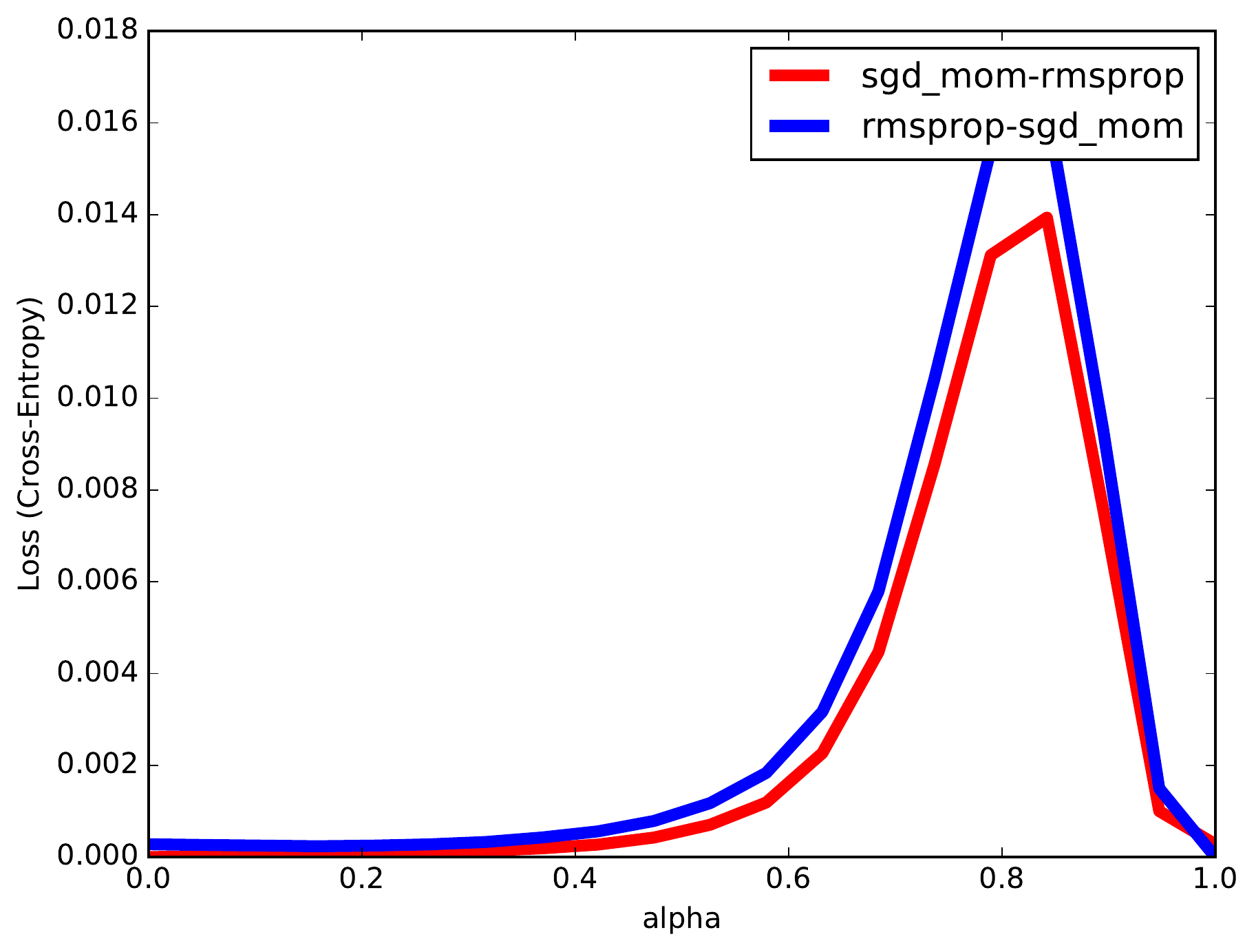}
        \centering \mysubfigurecaption{SGDM to RMSprop}
    \end{minipage}
    \begin{minipage}{0.32\textwidth}
        \includegraphics[width=\linewidth]{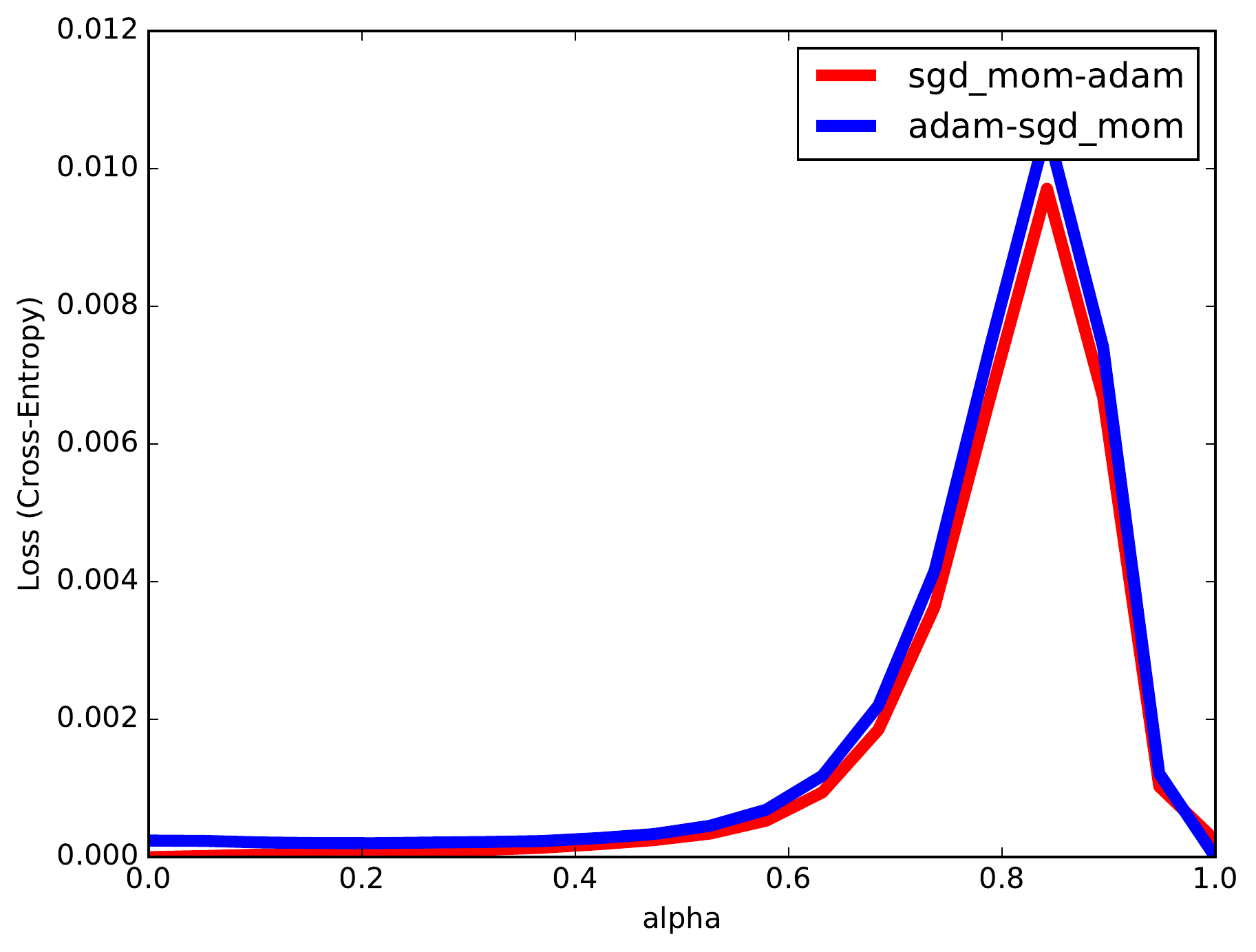}
        \centering \mysubfigurecaption{SGDM to ADAM}
    \end{minipage}
    \begin{minipage}{0.32\textwidth}
        \includegraphics[width=\linewidth]{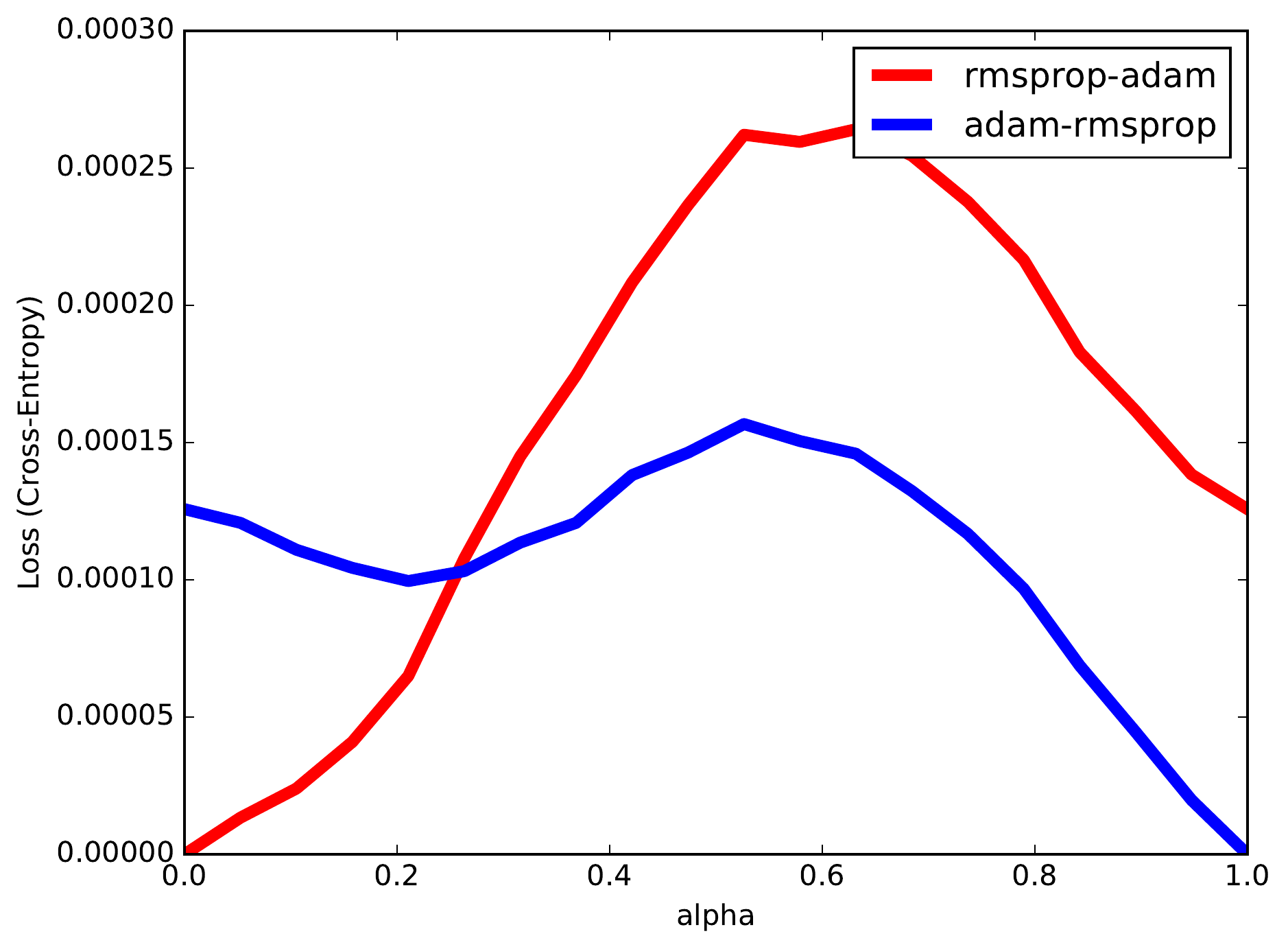}
        \centering \mysubfigurecaption{RMSprop to ADAM}
    \end{minipage}\\
    \caption{The projection of the loss surface at weight vectors between the initial weight and the learned weights found by the two optimization methods. Color as well as height of the surface indicate the loss function value. In the upper triangle, we plot the functional difference between the network corresponding to the learned weights for the first algorithm and networks corresponding to weights linearly interpolated between the first and second algorithm's learned weights.}
    \label{fig:function_diff}
\end{figure}

\begin{figure}[htp]
    \centering
    \begin{minipage}{0.325\textwidth}
        \includegraphics[width=\linewidth]{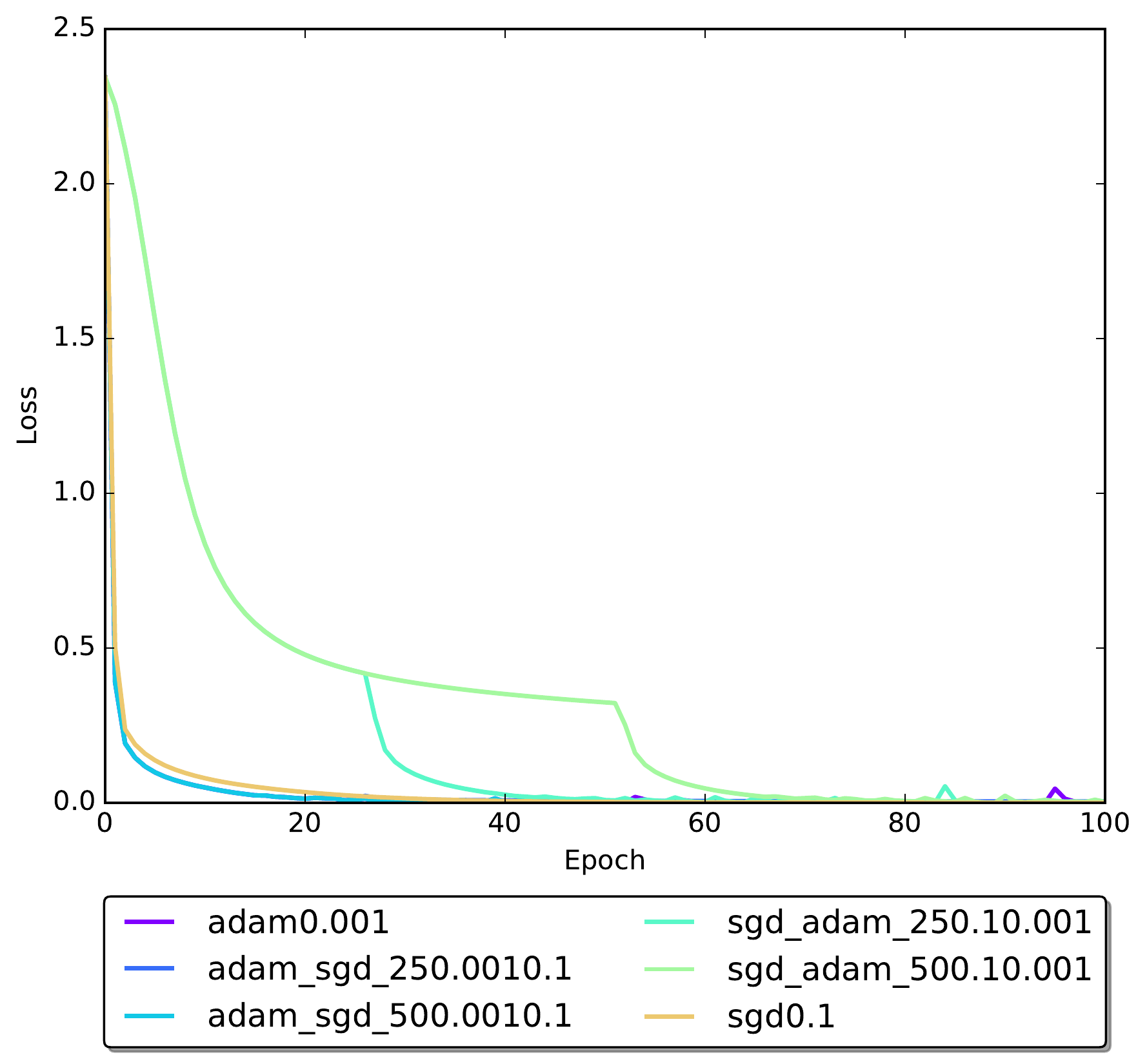}
        \mysubfigurecaption{NN - Loss Curve.}
    \end{minipage}
    \begin{minipage}{0.325\textwidth}
        \includegraphics[width=\linewidth]{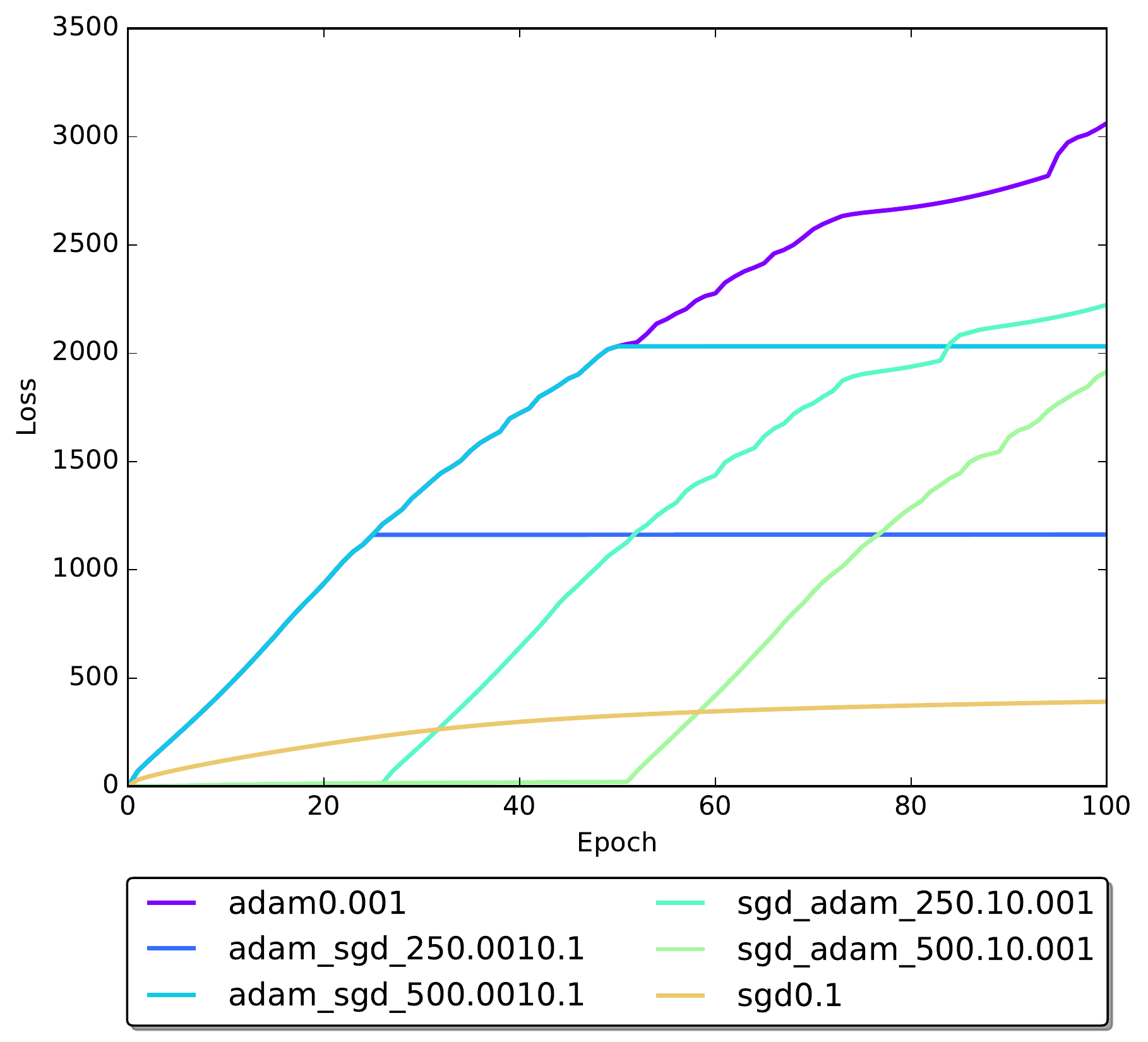}
        \mysubfigurecaption{NN - Traveled distance from initial parameter.}
    \end{minipage}
    \begin{minipage}{0.325\textwidth}
        \includegraphics[width=\linewidth]{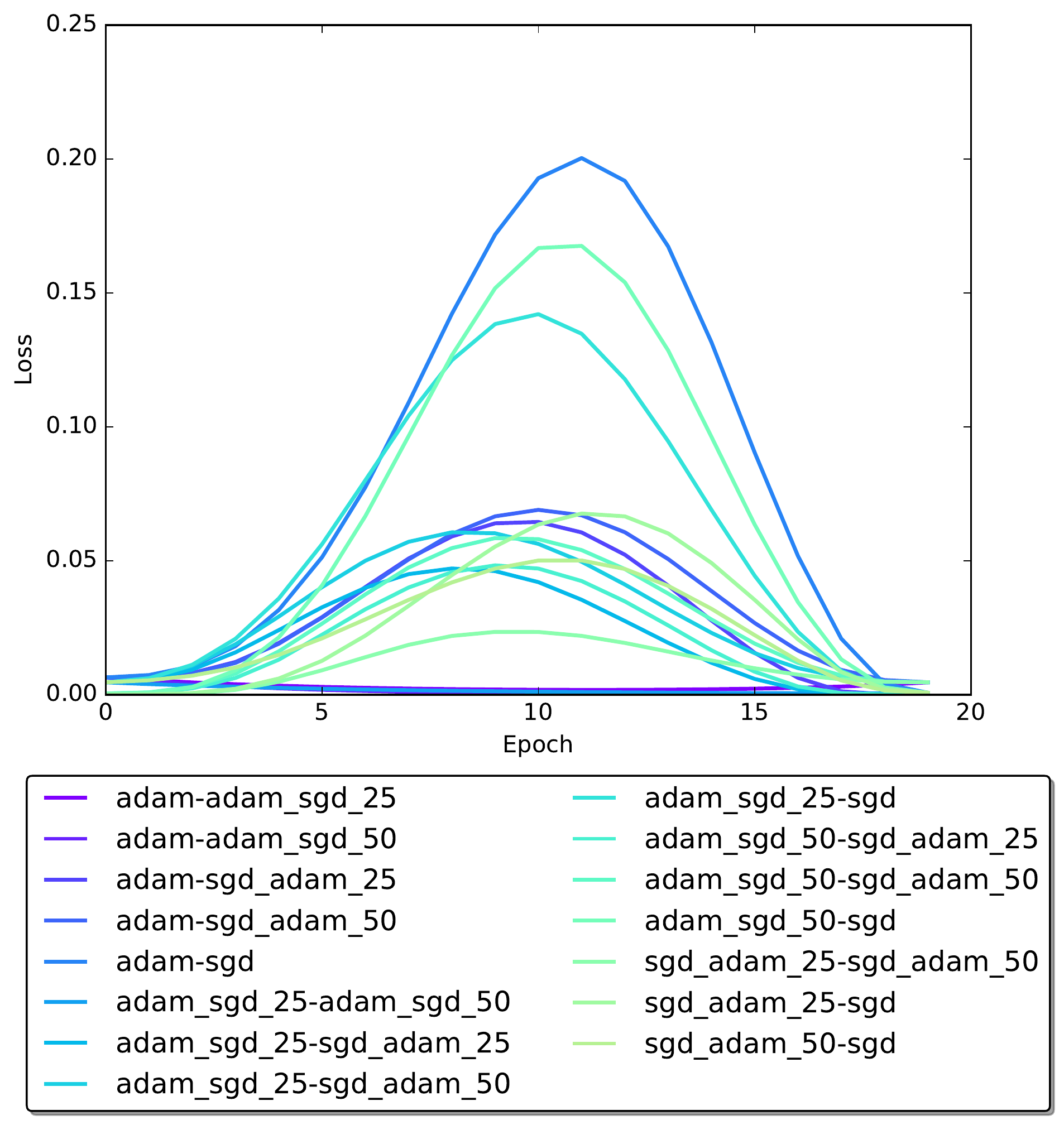}
        \mysubfigurecaption{NN - Final to Final.}
    \end{minipage}

    \caption{Switch algorithms}
    \label{fig:opt_switch_mnist}
\end{figure}
\pagebreak

\subsection{Effects of Batch-Normalization and Extreme Initializations}
\label{app:initializations}
\begin{figure}[htp]
    \centering
    \begin{minipage}{0.32\textwidth}
        \includegraphics[width=\linewidth]{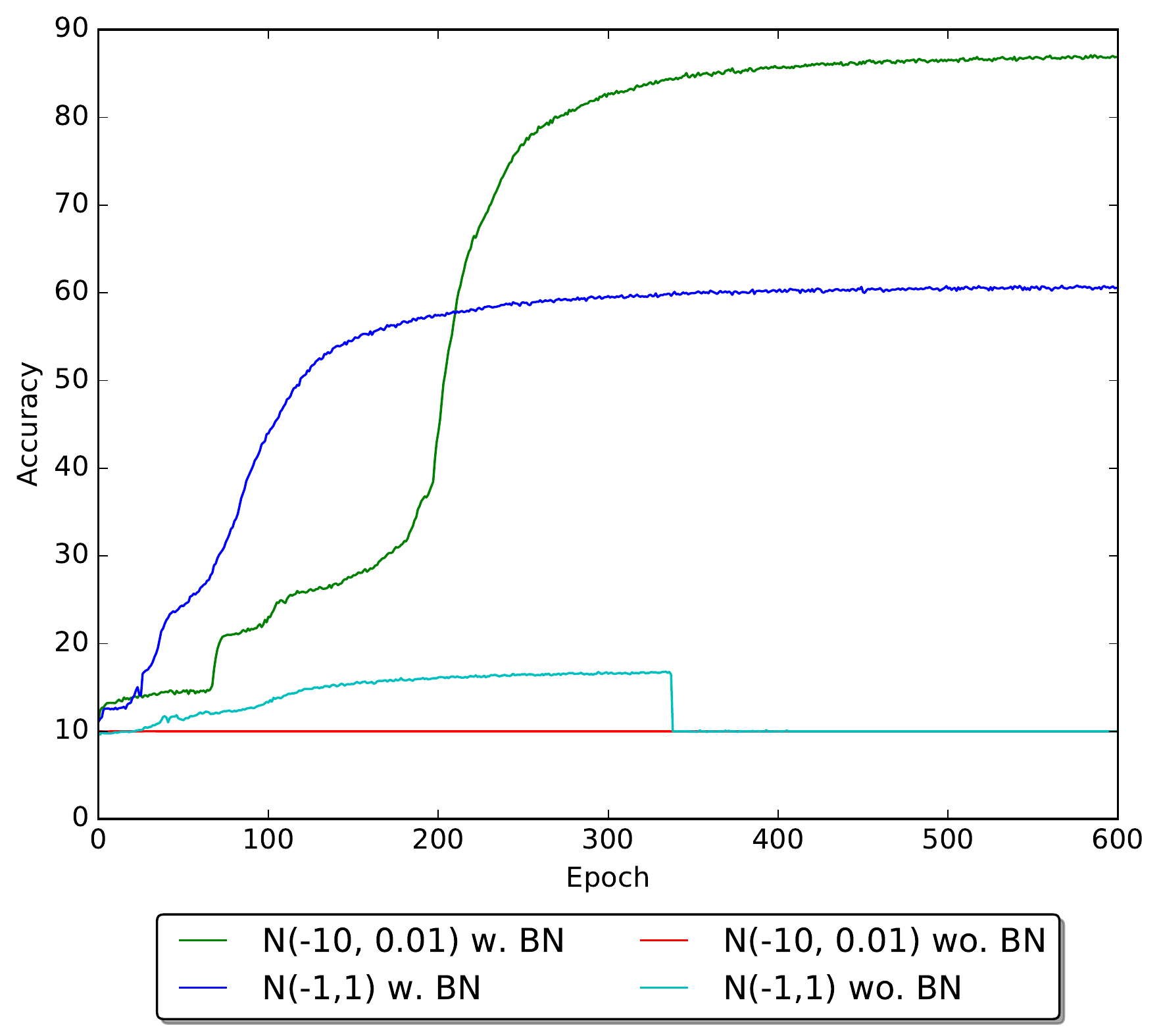}
        \mysubfigurecaption{(a) NIN - Learning curve}
    \end{minipage}
    \begin{minipage}{0.32\textwidth}
        \includegraphics[width=\linewidth]{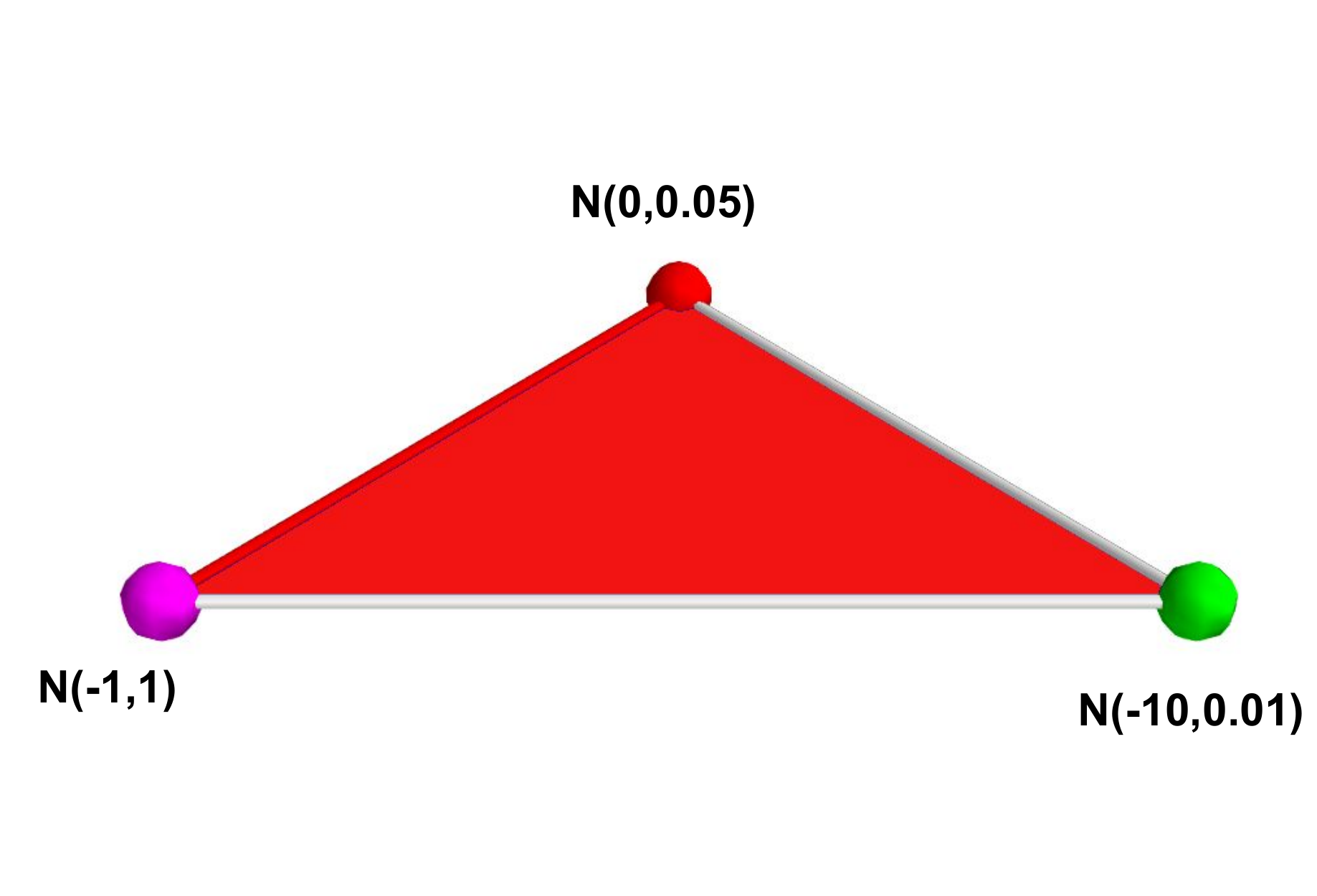}
        \mysubfigurecaption{(b) NIN without batch normalization}
        %\label{fig:nin_nb_weird_init}
    \end{minipage}
    \begin{minipage}{0.32\textwidth}
        \includegraphics[width=\linewidth]{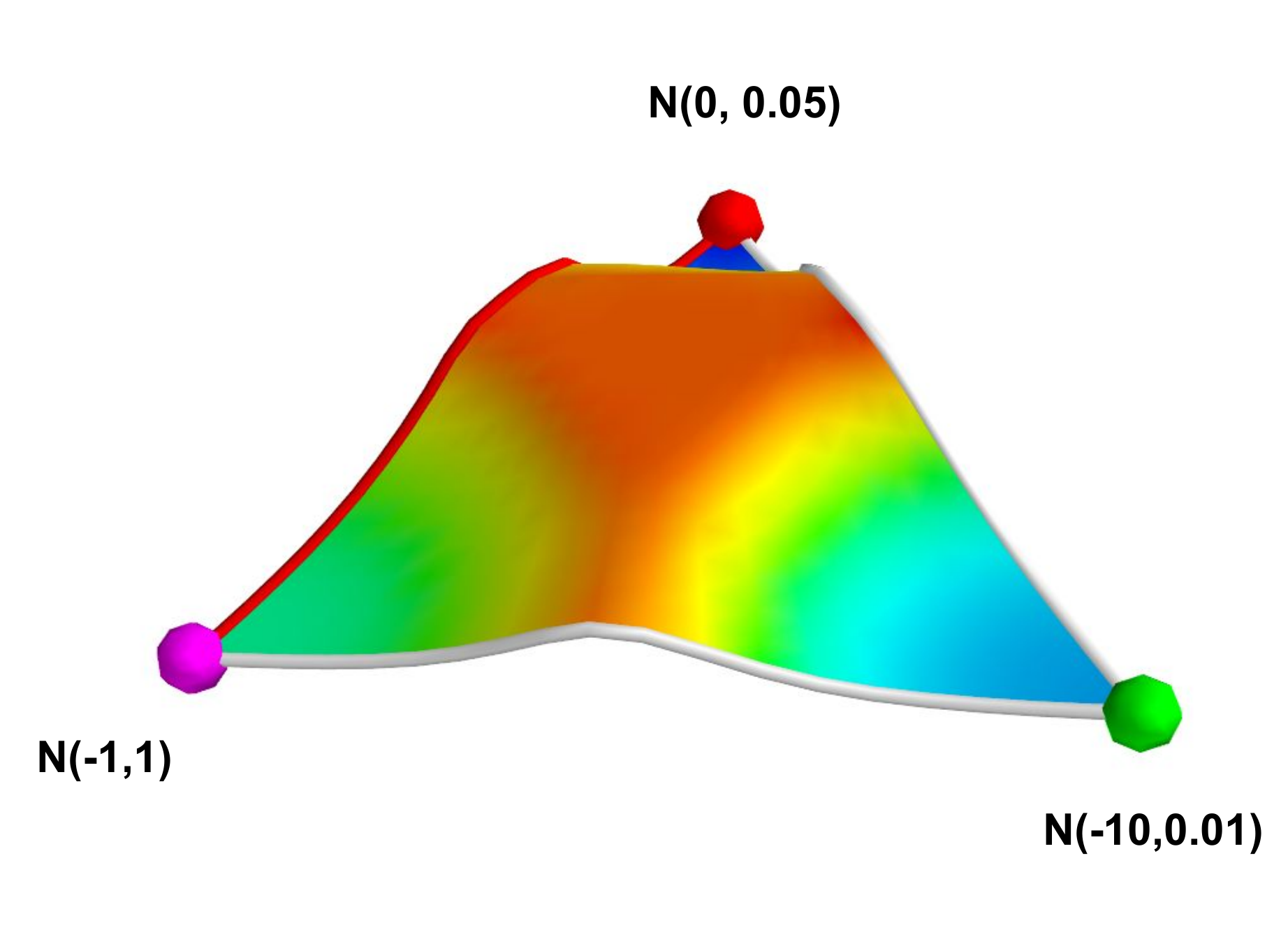}
        \mysubfigurecaption{(c) NIN with batch normalization}
    \end{minipage}
    \caption{NIN trained from different initializaitons.}
    \label{fig:exotic_init}
\end{figure}

The neural networks are typically initialized with very small parameter 
values \citep{Glorot2010, He2015}. Instead, we trained NIN with exotic 
intializations such as initial parameters drawn from $\mathcal{N}(-10.0, 0.01)$ 
or $\mathcal{N}(-1.0, 1.0)$ and observe the loss surface behaviours.
The results are shown in Figure~\ref{fig:exotic_init}.
We can see that NIN without BN does not train at all with any of these initializations.
{\em Swirszcz2016 et al.} mentioned that bad performance of neural networks trained with 
these initializations are due to finding a bad local minima.
However, we see that loss surface region around these initializations are plateau 
\footnote{We used same initializations as \citep{Swirszcz2016}
but we trained different neural networks with SGD on a different dataset. We used NIN and CIFAR10 
and {\em Swirszcz2016 et al.} used smaller neural network and MNIST.} 
rather than a bad local minima as shown in Figure~\ref{fig:exotic_init}(b).
On the other hand, NIN with BN does train slowly over time but finds a local minima.
This implies that BN redeems the ill-posed loss surface (plateau region).
Nevertheless, the local minima it found was not good as when the parameters were 
initialized with small values. However, it is not totally clear whether this is due to
difficulty of training or due to falling in a bad local minima.

\subsection{Switching optimization methods}
\begin{figure}[htp]
    \centering
    \begin{minipage}{\textwidth}
        \includegraphics[width=0.33\linewidth]{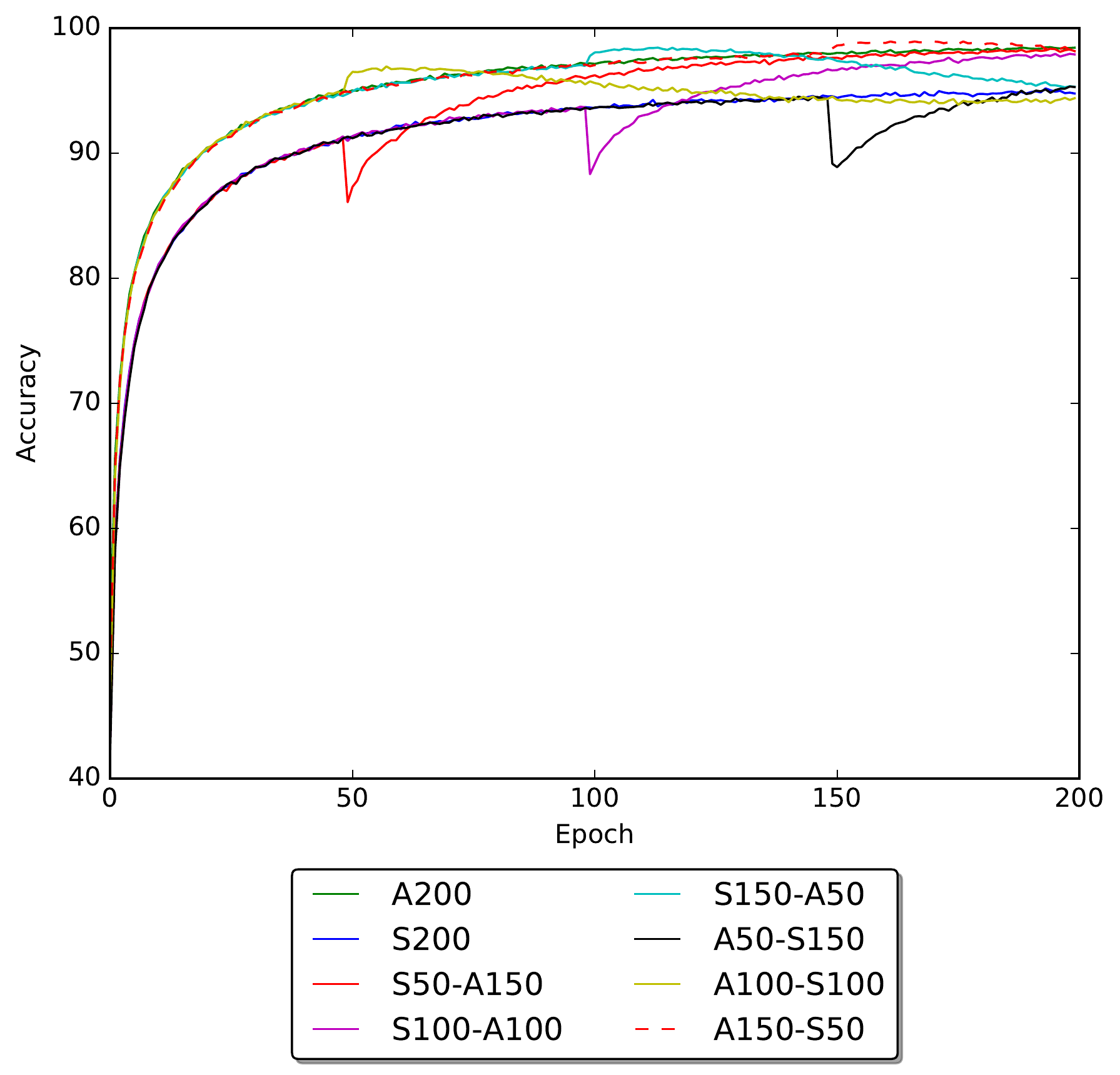}
        \includegraphics[width=0.33\linewidth]{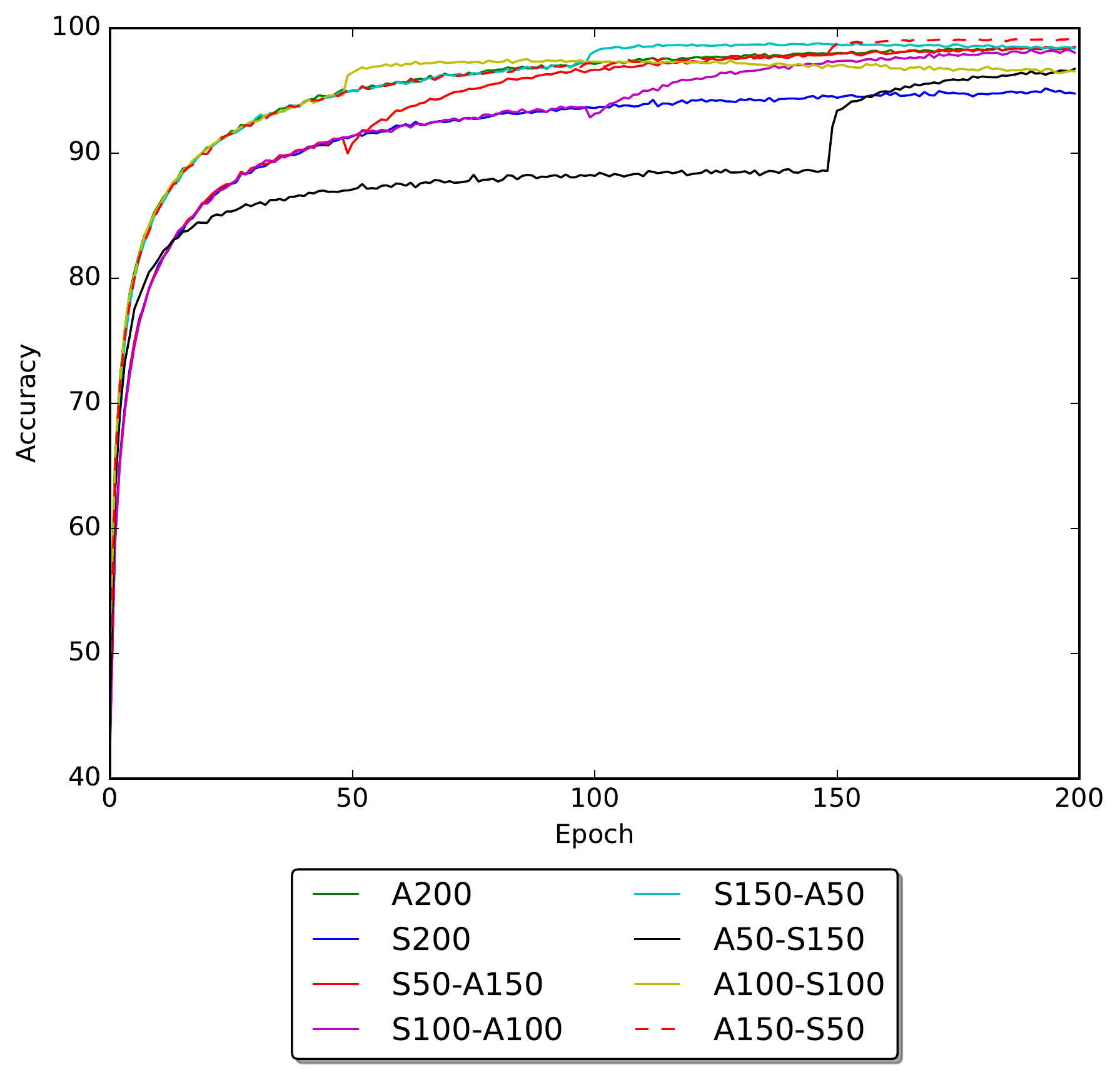}
        \includegraphics[width=0.32\linewidth]{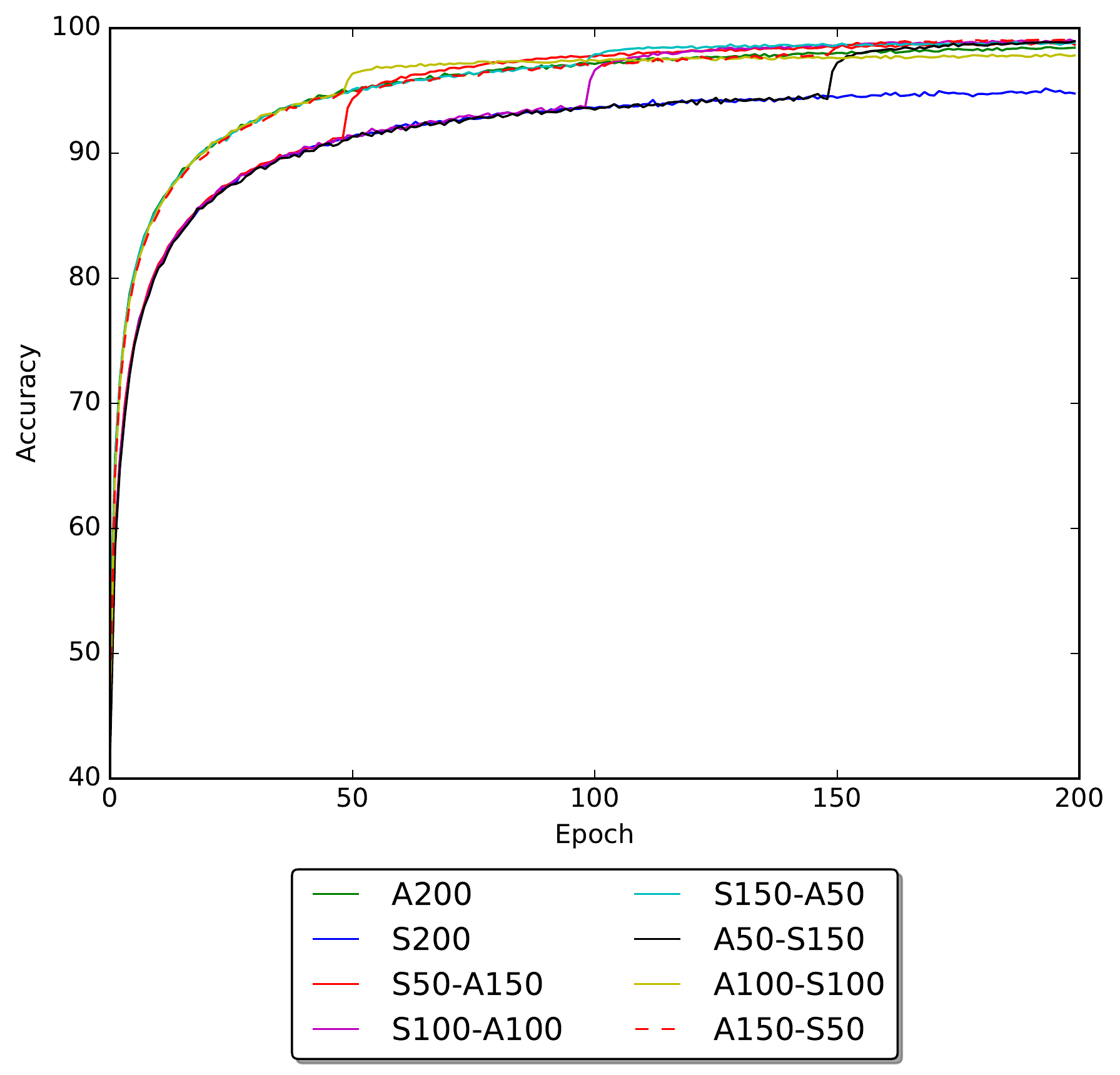}
    \end{minipage}\\
    \caption{NIN - Learning curve when switching methods from 
    SGD to ADAM and visa versa at
    epoch 50 and 100. Learning rate switched from SGD (ADAM) to ADAM (SGD) at
    (left) 0.001 (0.1) to 0.1 (0.001), 
    (middle) 0.001 (0.1) to 0.05, (0.001), and
    (right) 0.001 (0.1) to 0.01 (0.001).}
    \label{fig:nin_switch_full}
\end{figure}

\begin{figure}[htp]
    \centering
    \begin{minipage}{\textwidth}
        \includegraphics[width=0.33\linewidth]{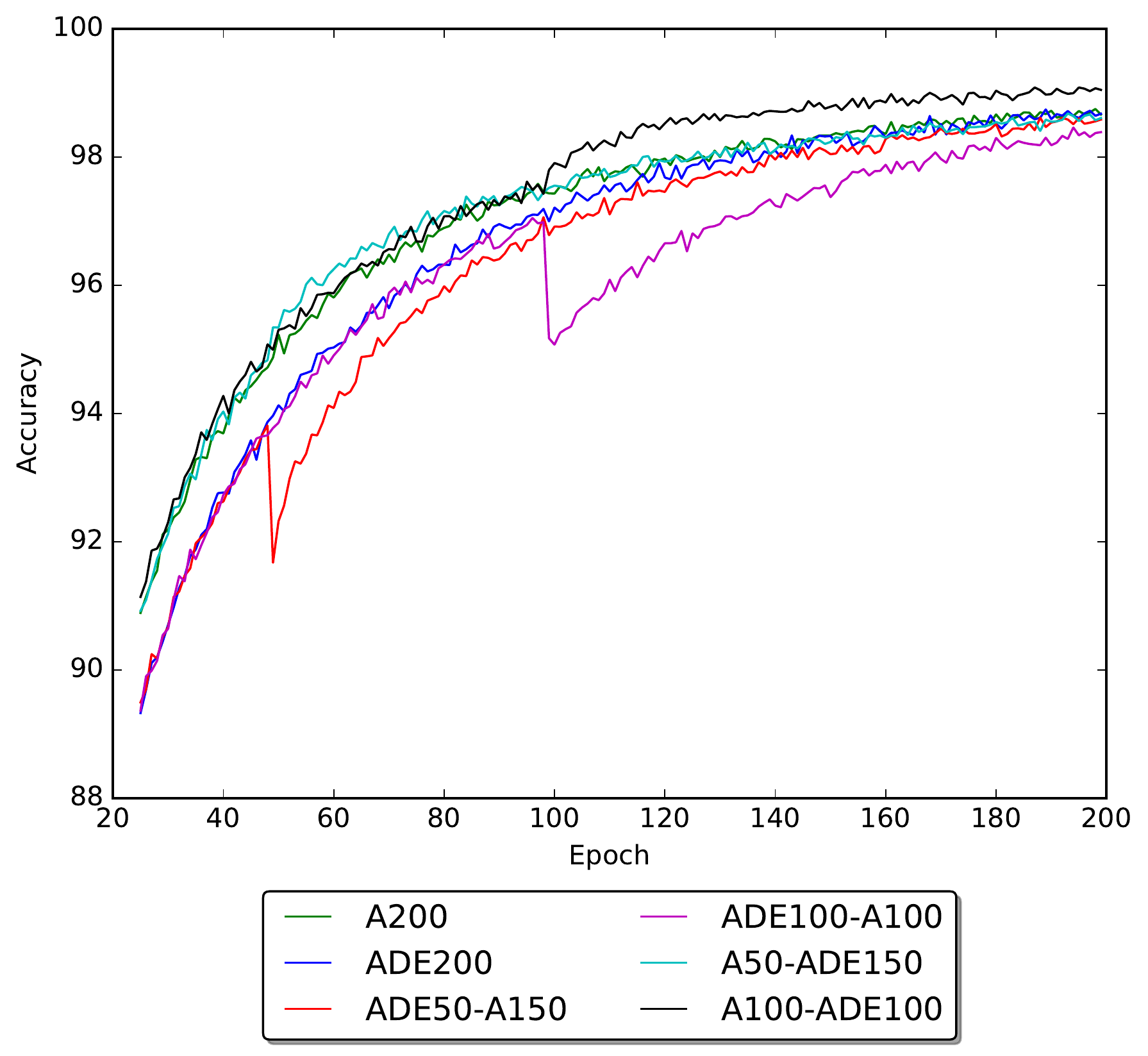}
        \includegraphics[width=0.33\linewidth]{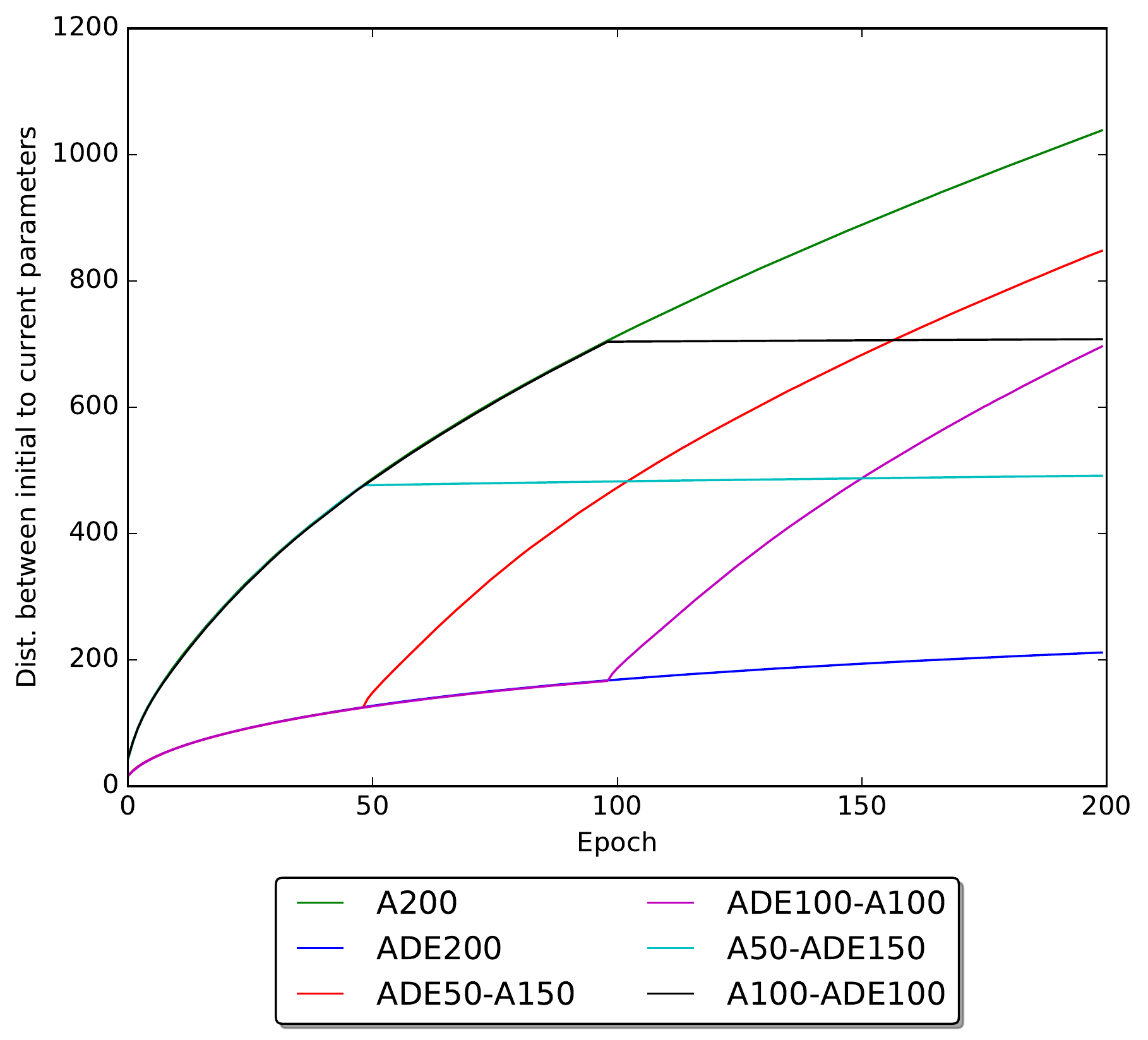}
        \includegraphics[width=0.32\linewidth]{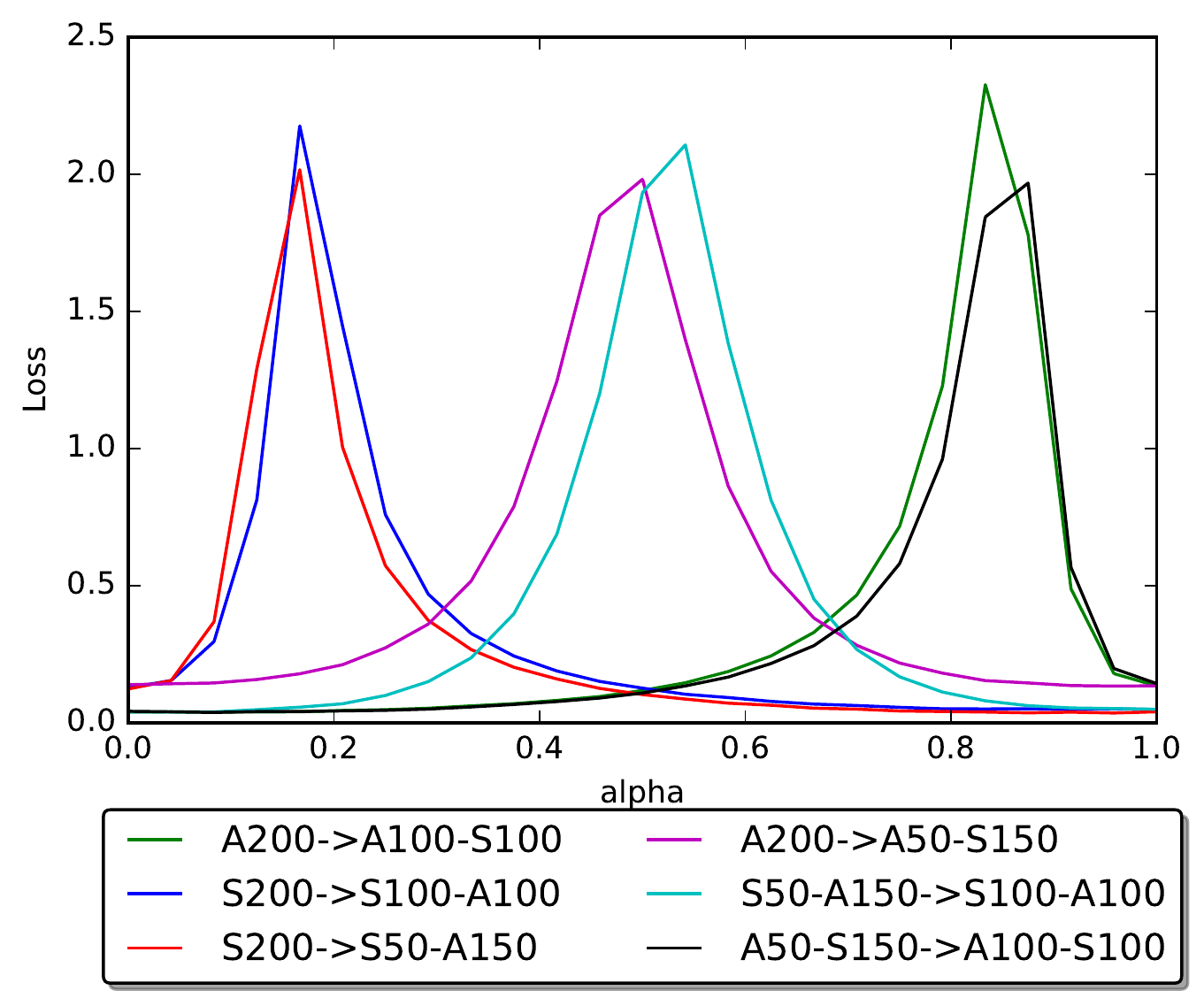}
        %\caption{Learning rate is not required for Adadelta. 
        %Learning rate is set to 0.001 for ADAM in the beginning,
        %and then switched it to 0.1. }
        %\label{fig:vgg_switch_adam_adadelta_large}
    \end{minipage}
    \caption{VGG - Switching methods from ADAM to Adadelta and Adadelta to ADAM at
    epoch 50 and 100. Zoomed in version (Left). Distance between initial
    weights to weights at each epoch (Middle). The interpolation between
    different convergence parameters (Right).
    Each figure shows the results of switching methods at different 
    learning rate. 
    We label the switch of methods in terms of ratio. 
    For instance, ADE50-A50 as trained with ADAM in the first 100 epoch 
    and swithced to Adadelta for the rest of the epoch.}
    \label{fig:vgg_switch_adam_adadelta}
\end{figure}

\begin{figure}[htp]
    \centering
    \begin{minipage}{\textwidth}
        \includegraphics[width=0.33\linewidth]{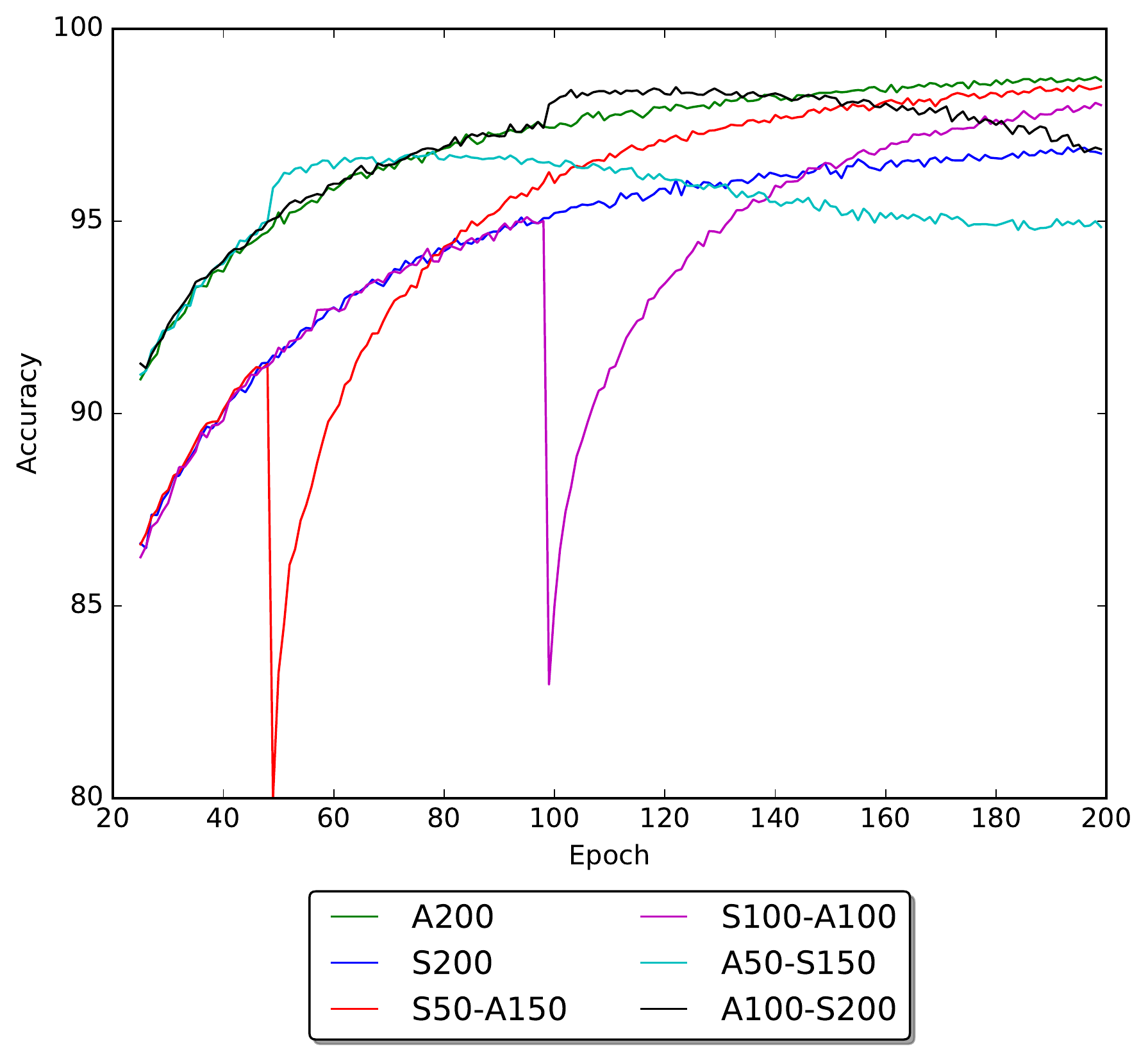}
        \includegraphics[width=0.33\linewidth]{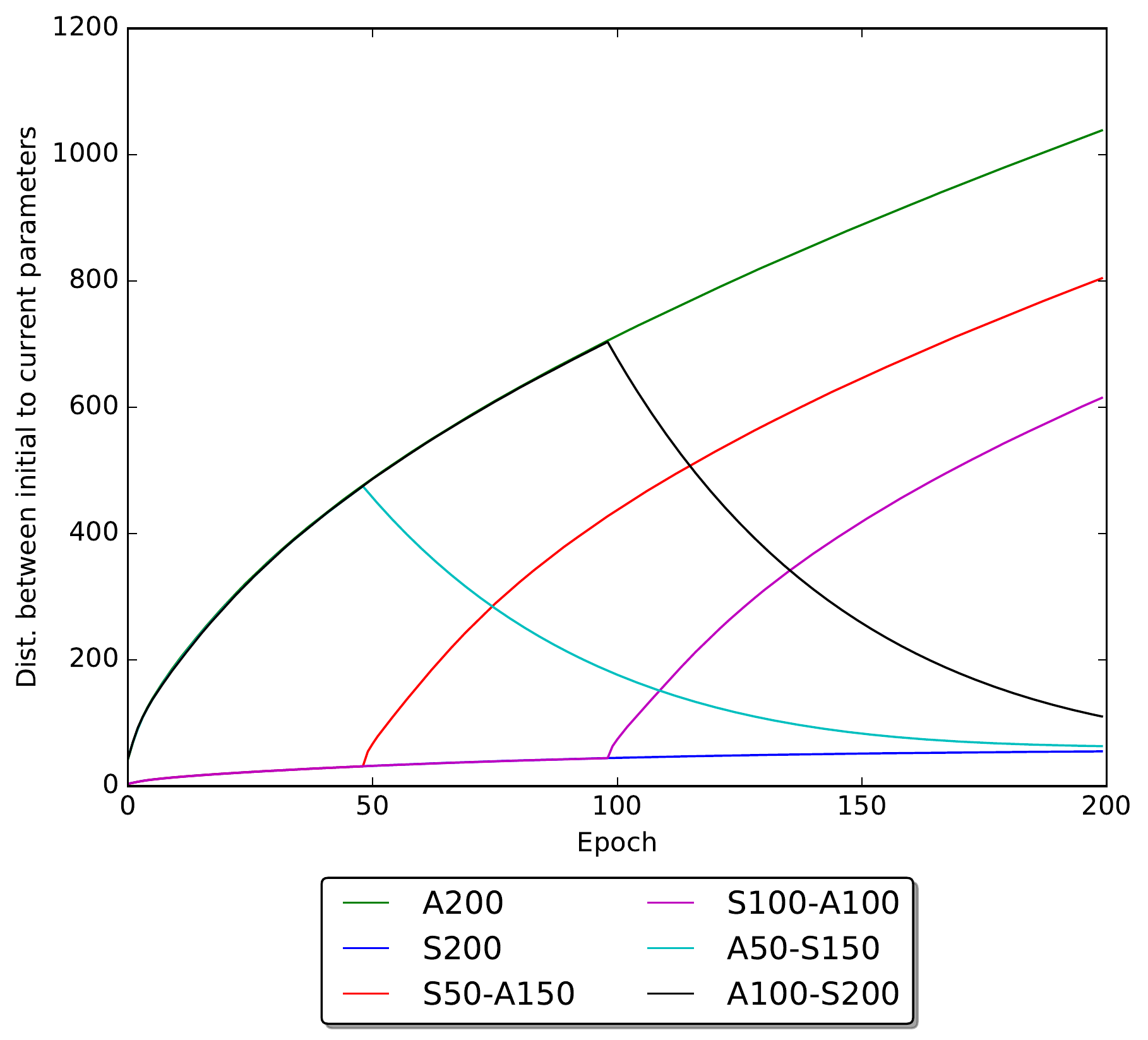}
        \includegraphics[width=0.32\linewidth]{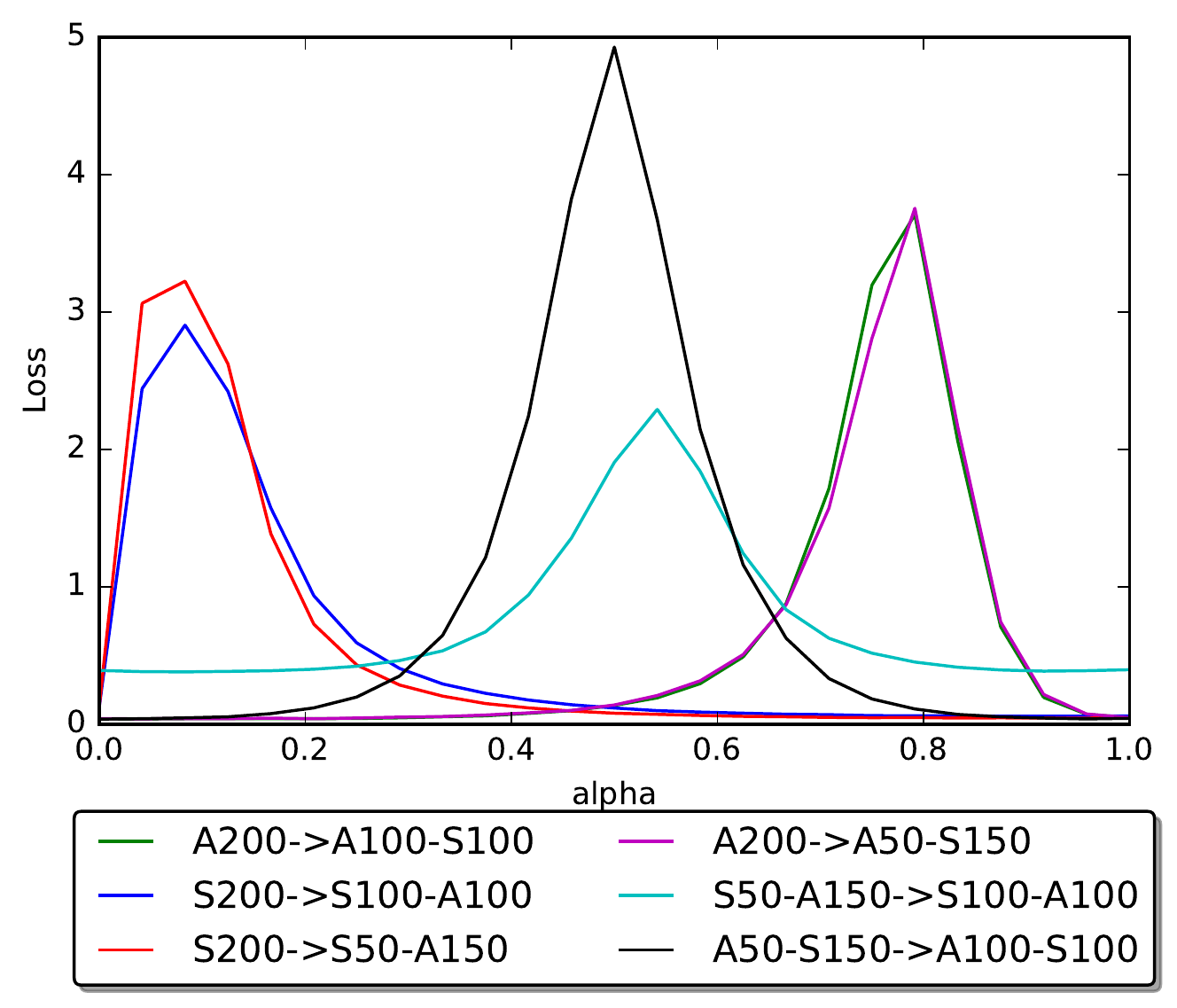}
        \mysubfigurecaption{(a) The learning rates is set to 0.001 and 0.05  for ADAM and SGD
        in the beginning, and then switched it to 0.05, 0.001 for 
        SGD and ADAM. }
        %\label{fig:vgg_switch_large}
    \end{minipage}\\
    \begin{minipage}{\textwidth}
        \includegraphics[width=0.33\linewidth]{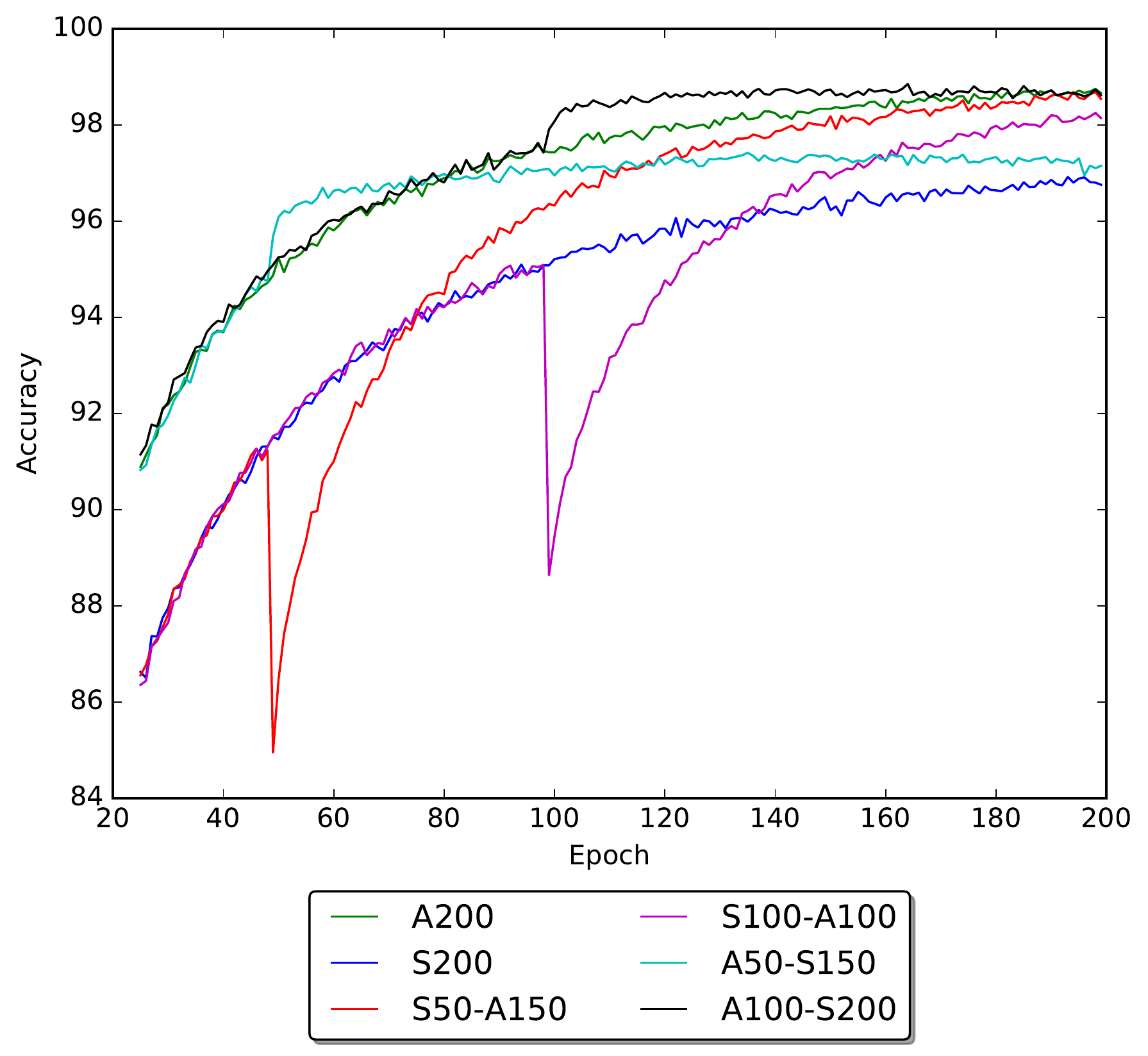}
        \includegraphics[width=0.33\linewidth]{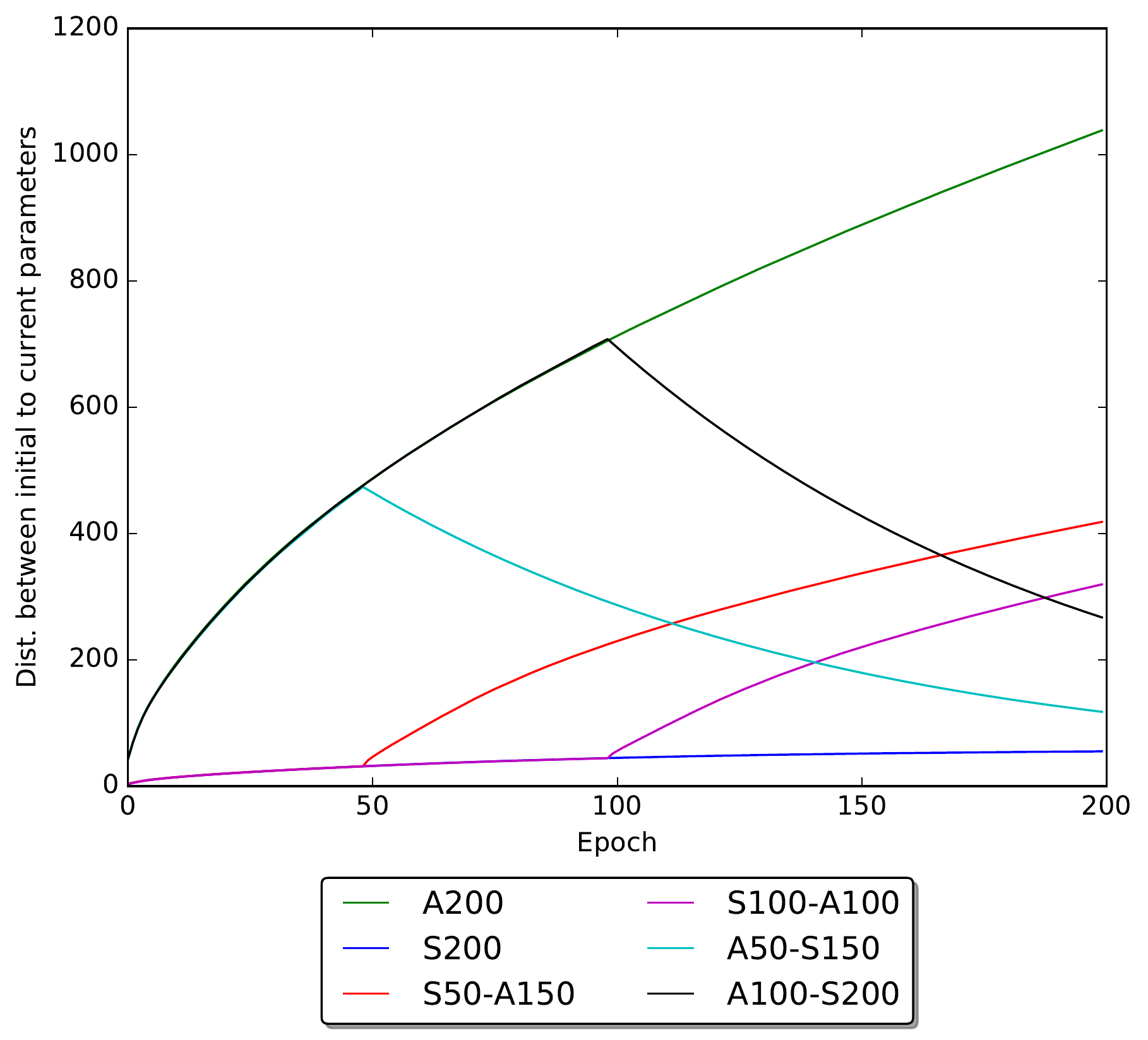}
        \includegraphics[width=0.32\linewidth]{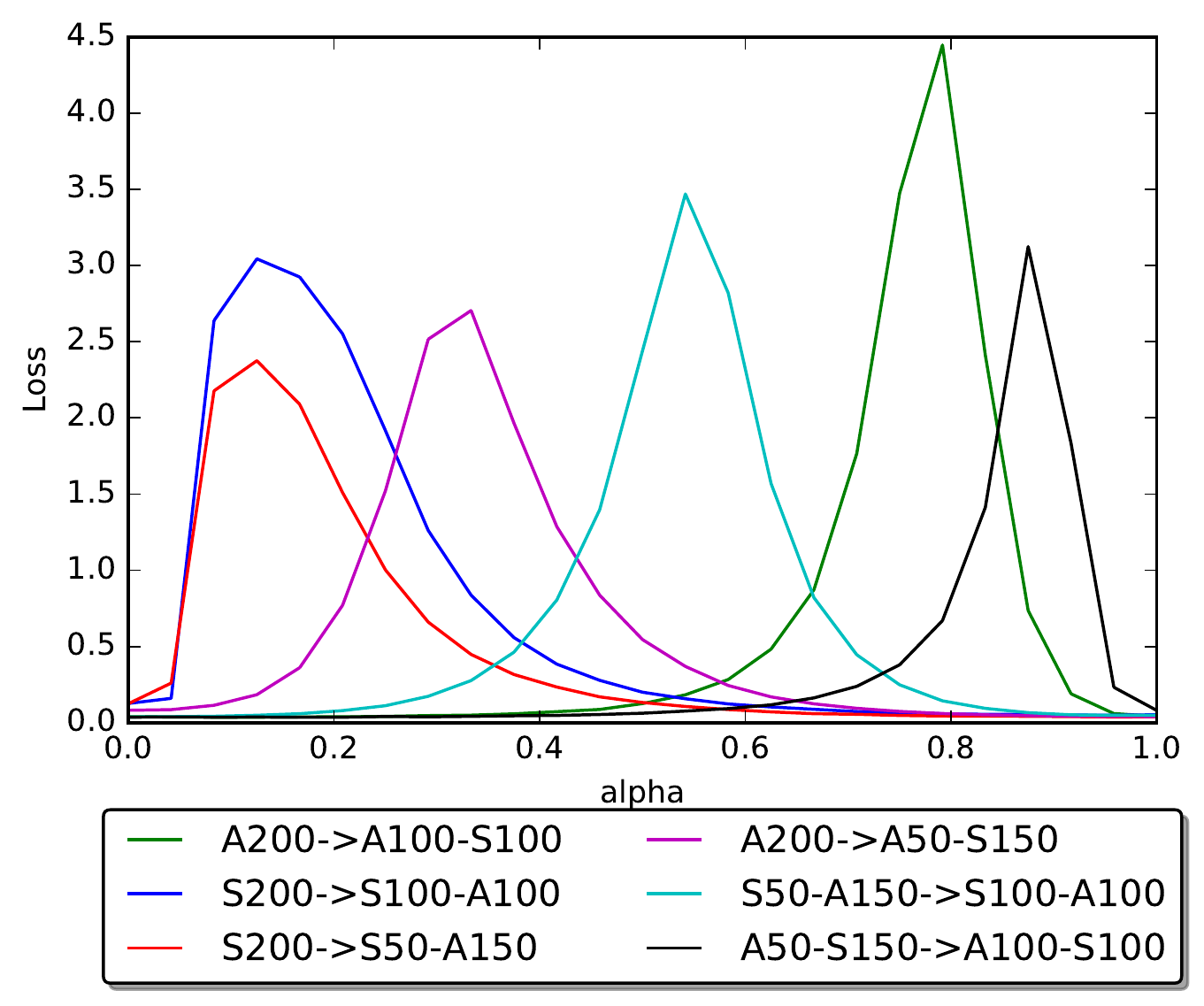}
        \mysubfigurecaption{(b) The learning rates is set to 0.001 and 0.05 for ADAM and SGD 
        in the beginning, and then switched it to 0.05, 0.0005 for 
        SGD and ADAM . }
        %\label{fig:vgg_switch_medium}
    \end{minipage}\\
    \begin{minipage}{\textwidth}
        \includegraphics[width=0.33\linewidth]{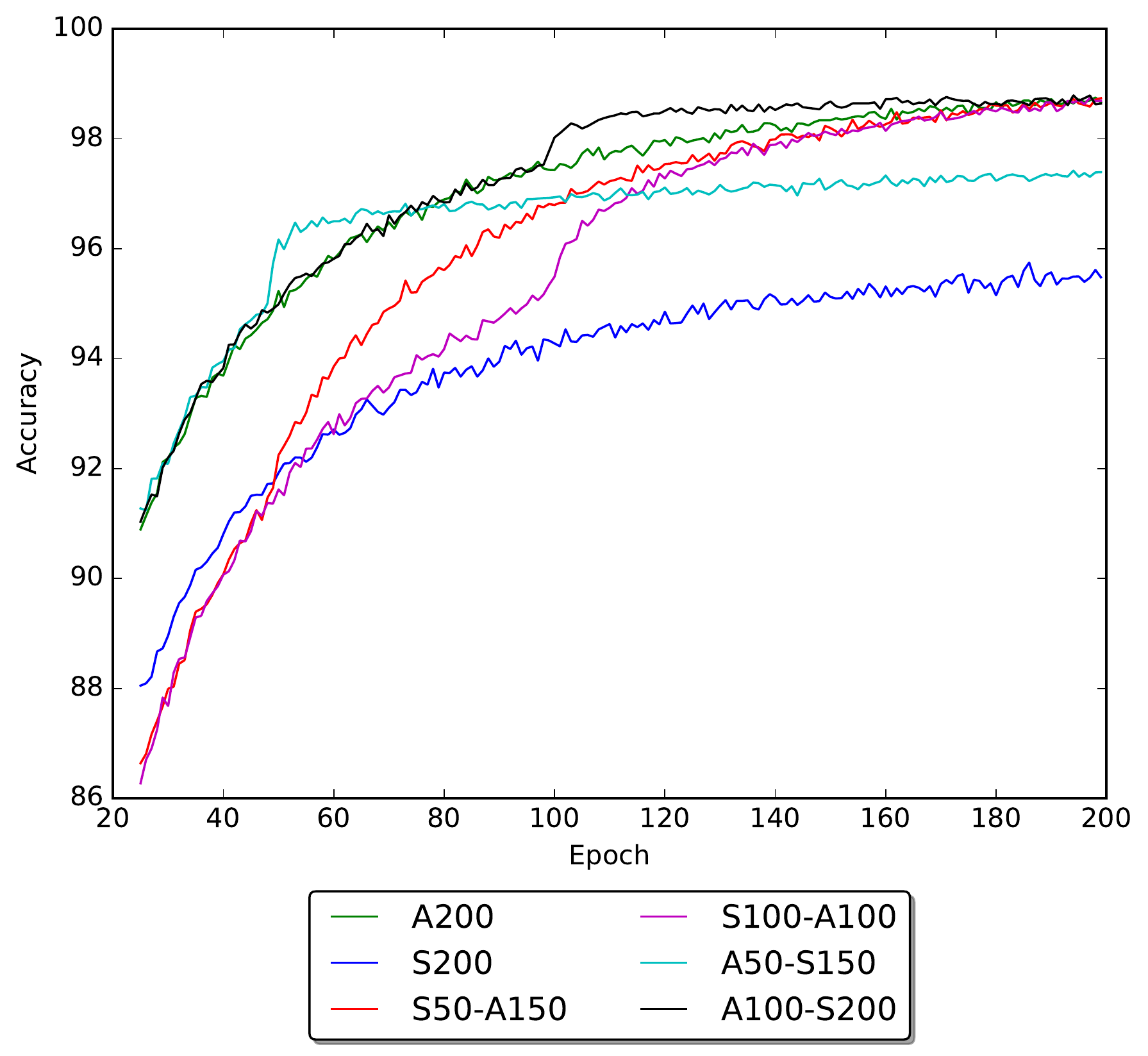}
        \includegraphics[width=0.33\linewidth]{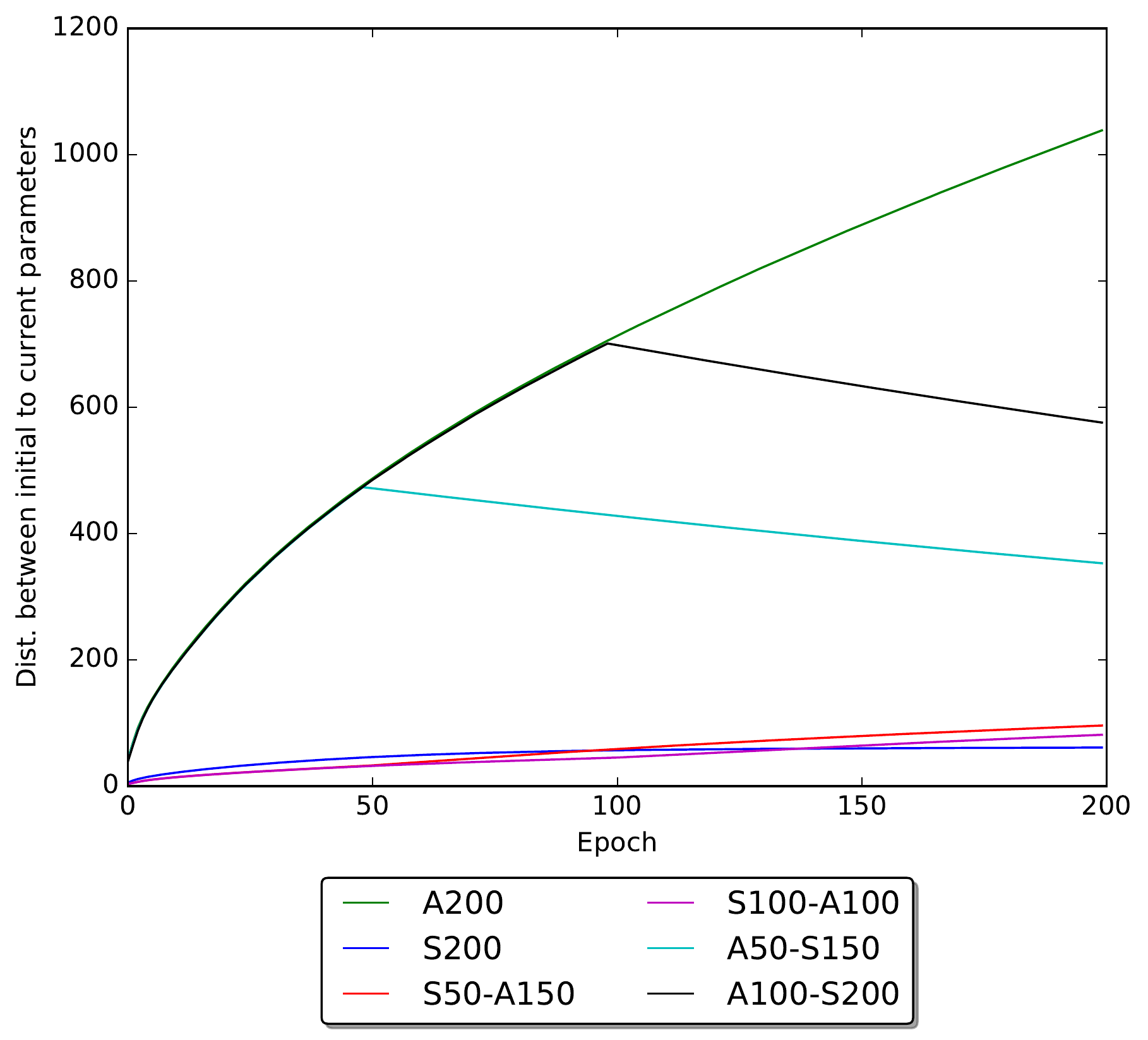}
        \includegraphics[width=0.32\linewidth]{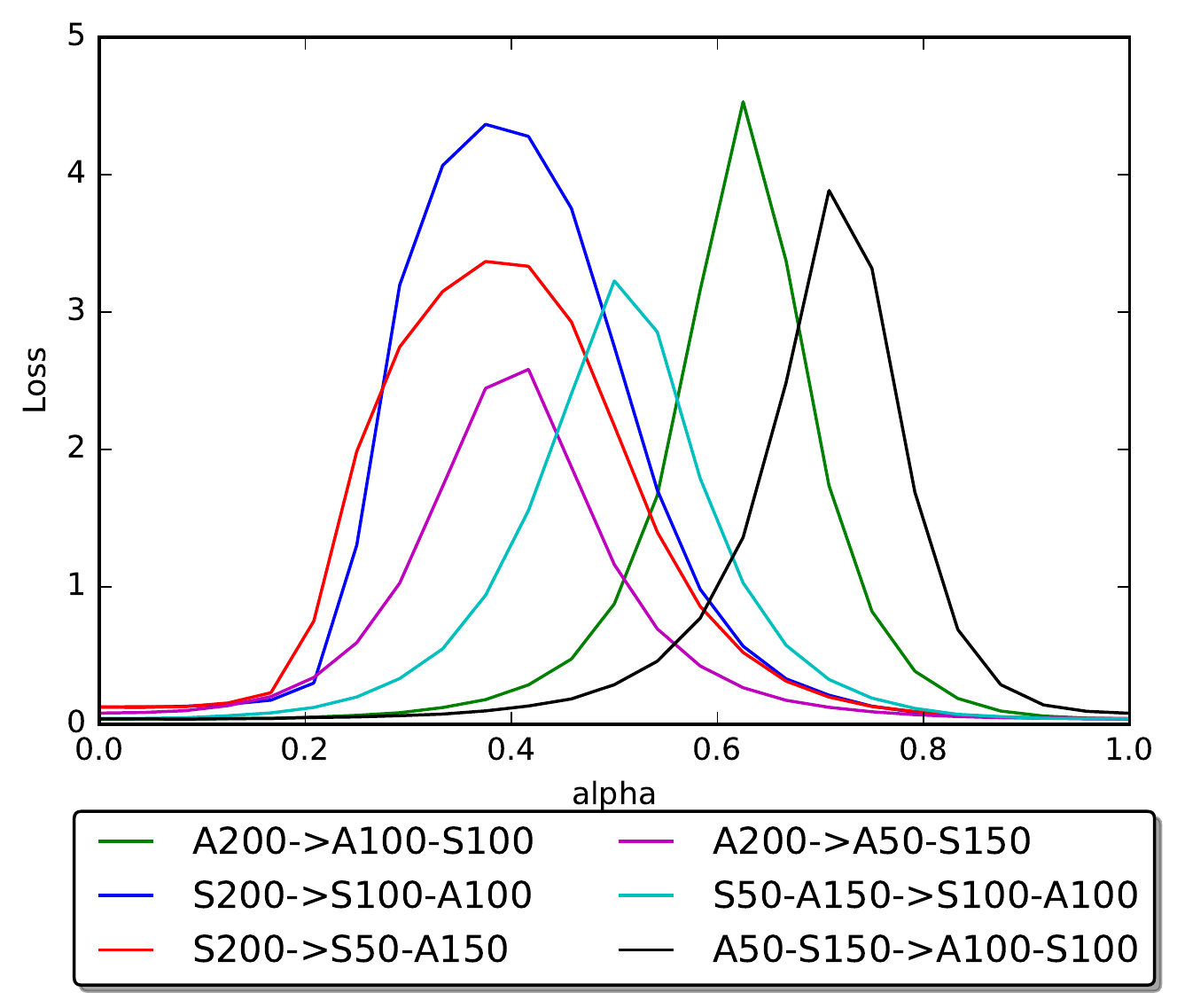}
        \mysubfigurecaption{(c) The learning rates is set to 0.001 and 0.05 for ADAM and SGD
        in the beginning, and then switched it to 0.01, 0.0001 for 
        SGD and ADAM. }
        %\label{fig:vgg_switch_small}
    \end{minipage}
    \caption{VGG - Switching methods from SGD to ADAM and ADAM to SGD at
    epoch 50 and 100. Zoomed in version (Left). Distance between initial
    weights to weights at each epoch (Middle). The interpolation between
    different convergence parameters (Right).
    Each figure shows the results of switching methods at different 
    learning rate.
    We label the switch of methods in terms of ratio. 
    For instance, S100-A100 as trained with SGD in the first 100 epoch 
    and swithced to ADAM for the rest of the epoch.}
    \label{fig:vgg_switch_sgd_adam}
\end{figure}

\begin{figure}[htp]
    \centering
    \begin{minipage}{\textwidth}
        \includegraphics[width=0.33\linewidth]{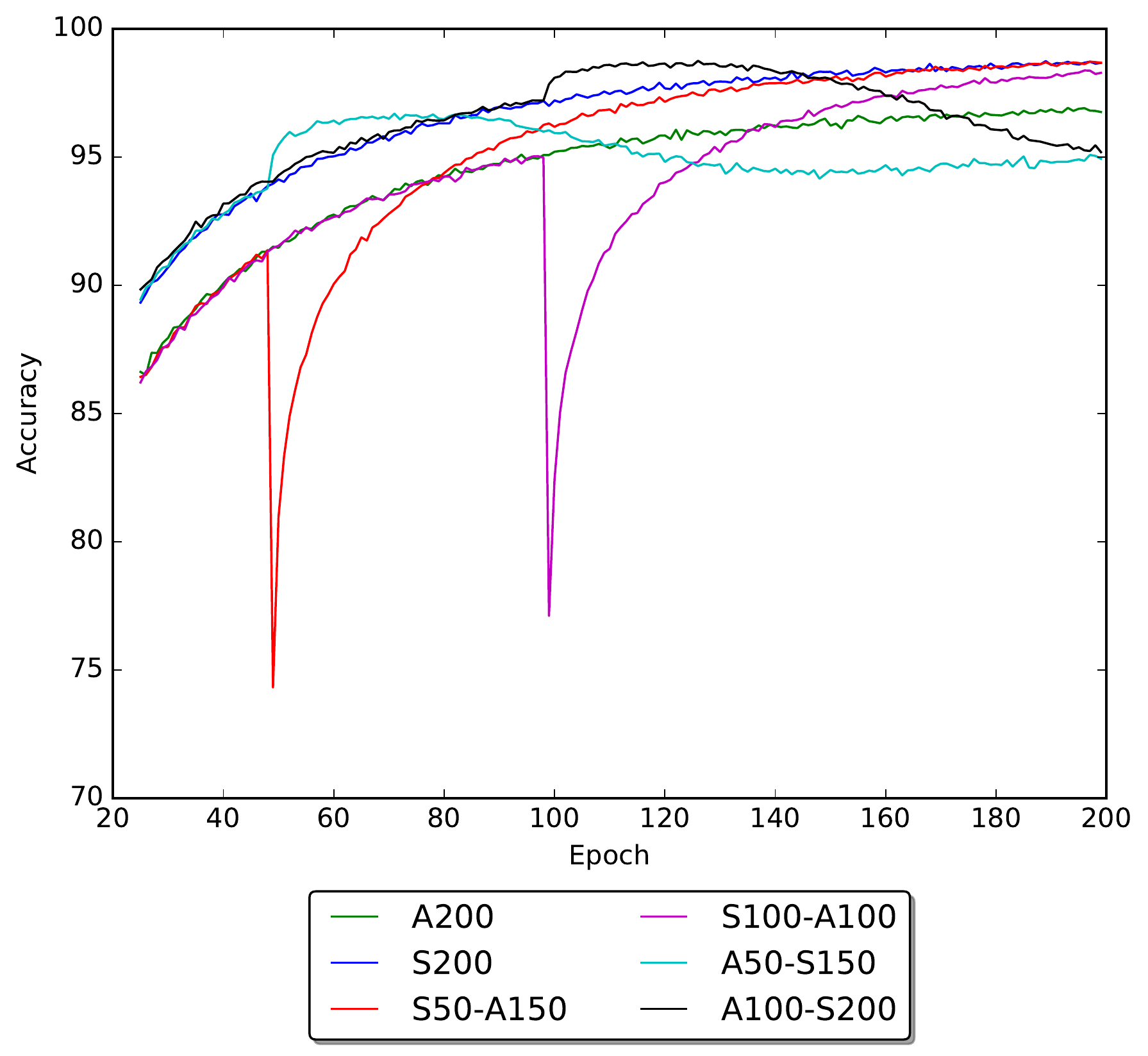}
        \includegraphics[width=0.33\linewidth]{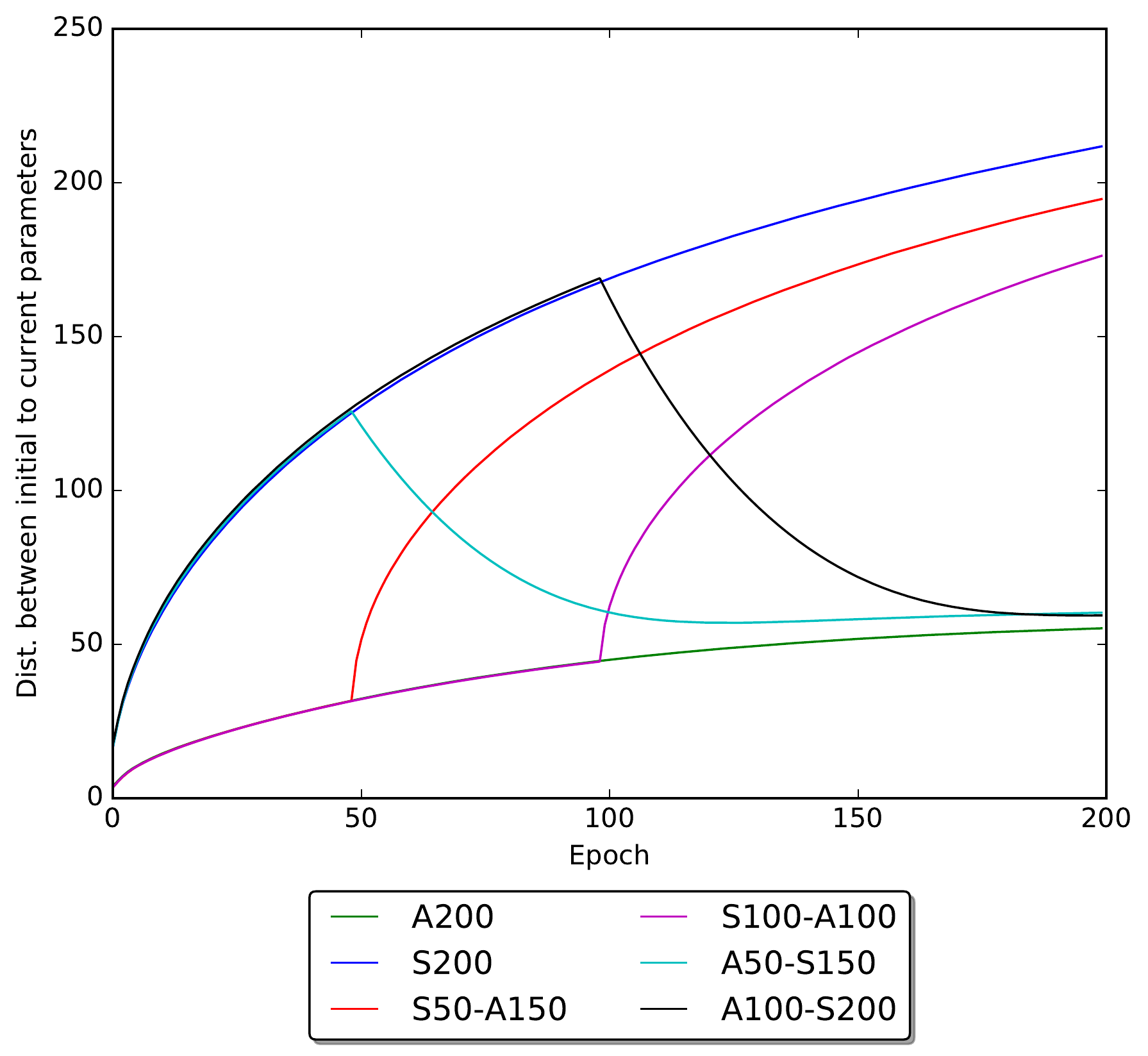}
        \includegraphics[width=0.32\linewidth]{vgg_v3_switch_sgdAdelta_inter_large.pdf}
        \mysubfigurecaption{Learning rate is not required for Adadelta. 
        Learning rate is set to 0.05 for SGD in the beginning,
        and then switched it to 0.1. }
        \label{fig:vgg_switch_adadelta_large}
    \end{minipage}\\
    \begin{minipage}{\textwidth}
        \includegraphics[width=0.33\linewidth]{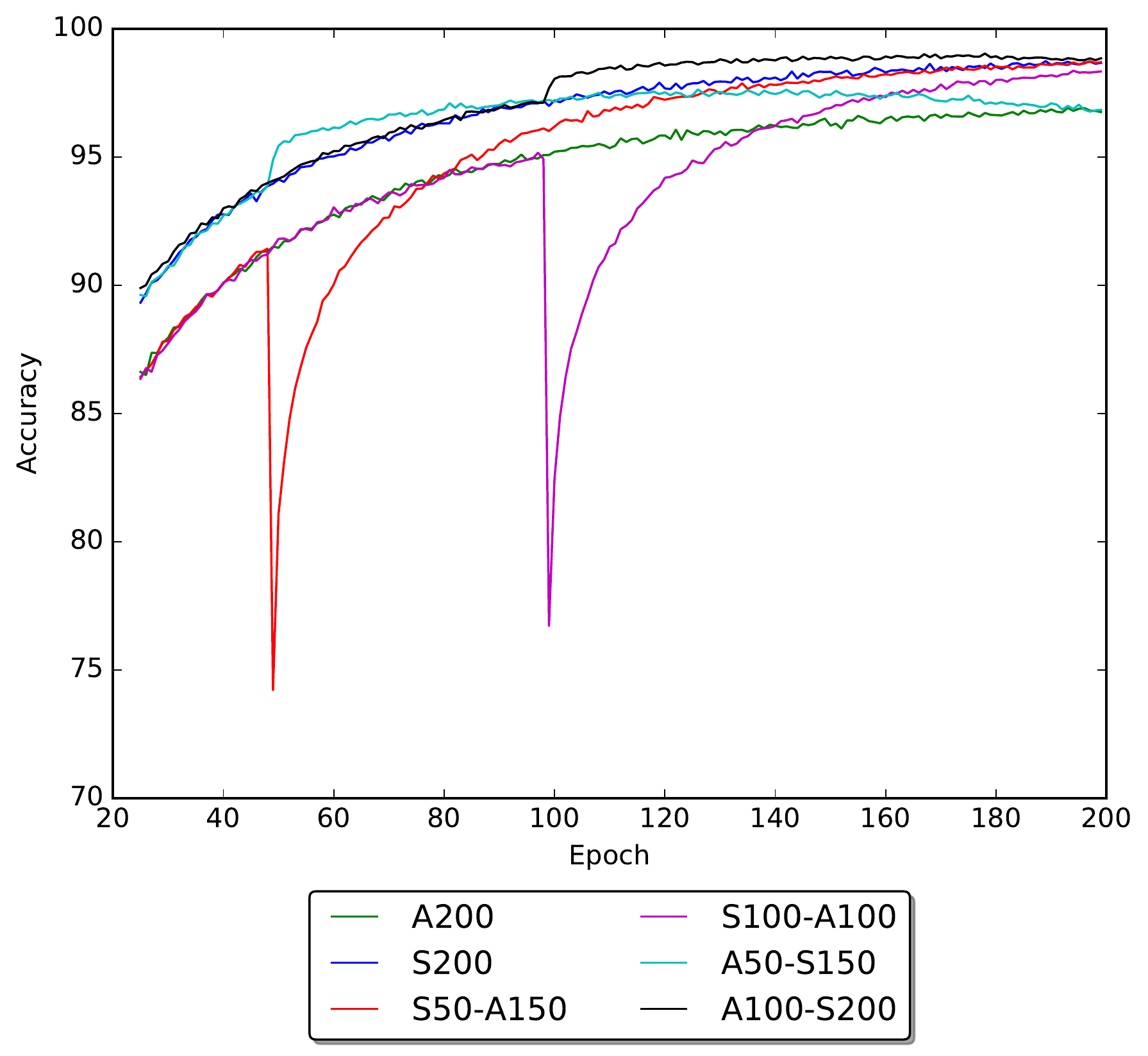}
        \includegraphics[width=0.33\linewidth]{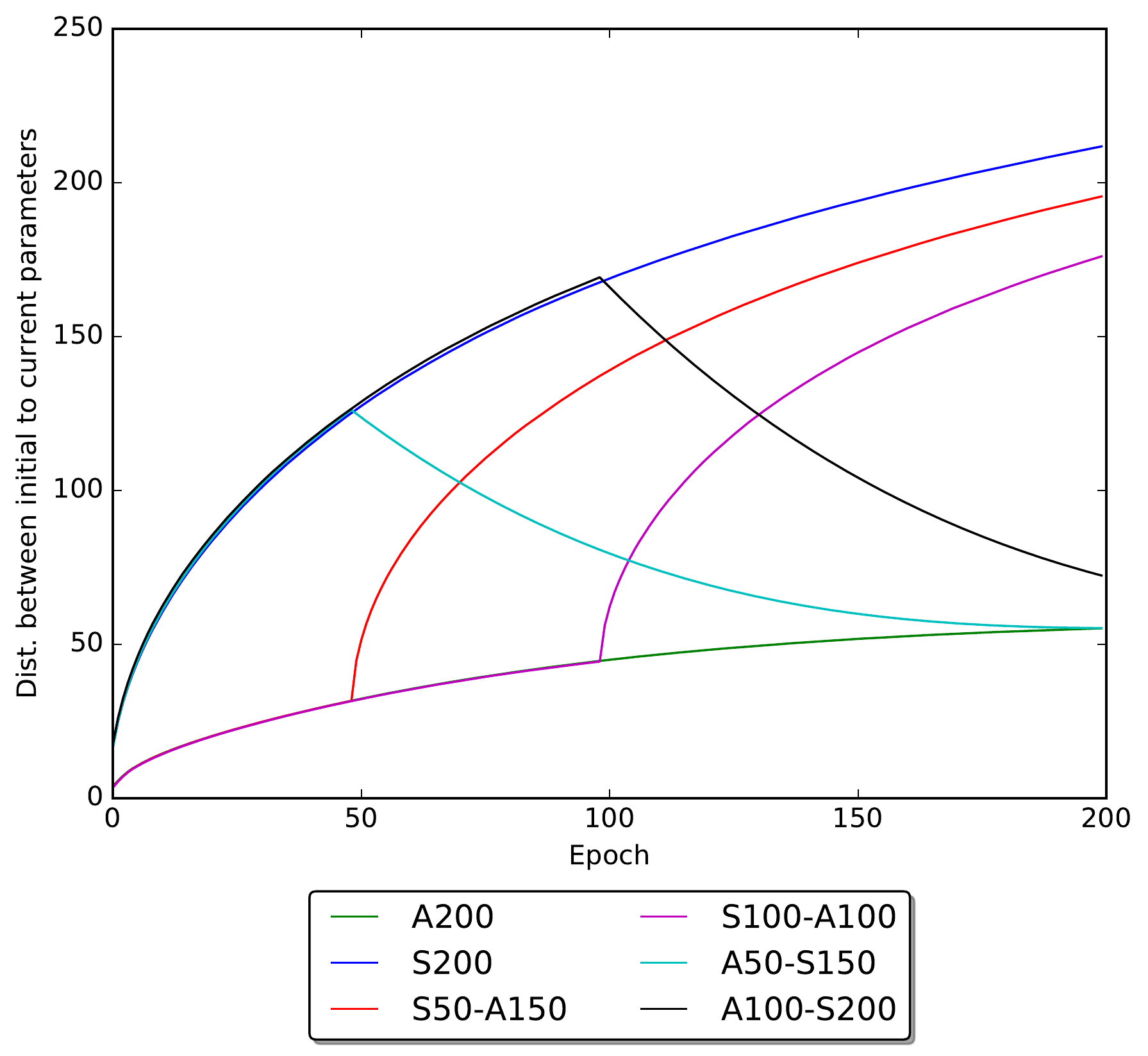}
        \includegraphics[width=0.32\linewidth]{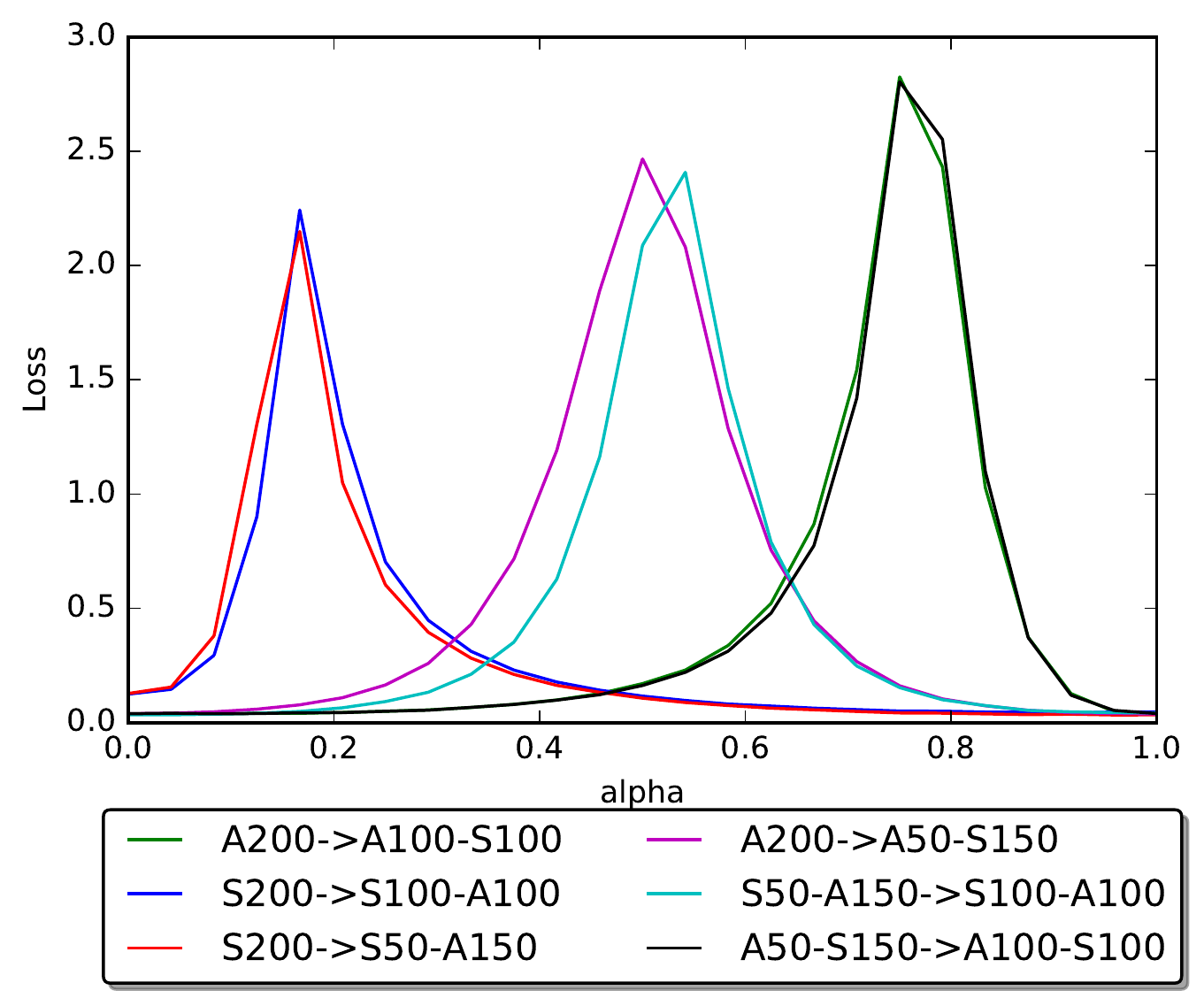}
        \mysubfigurecaption{
        Learning rate is set to 0.05 for SGD. }
        \label{fig:vgg_switch_adadelta_medium}
    \end{minipage}\\
    \begin{minipage}{\textwidth}
        \includegraphics[width=0.33\linewidth]{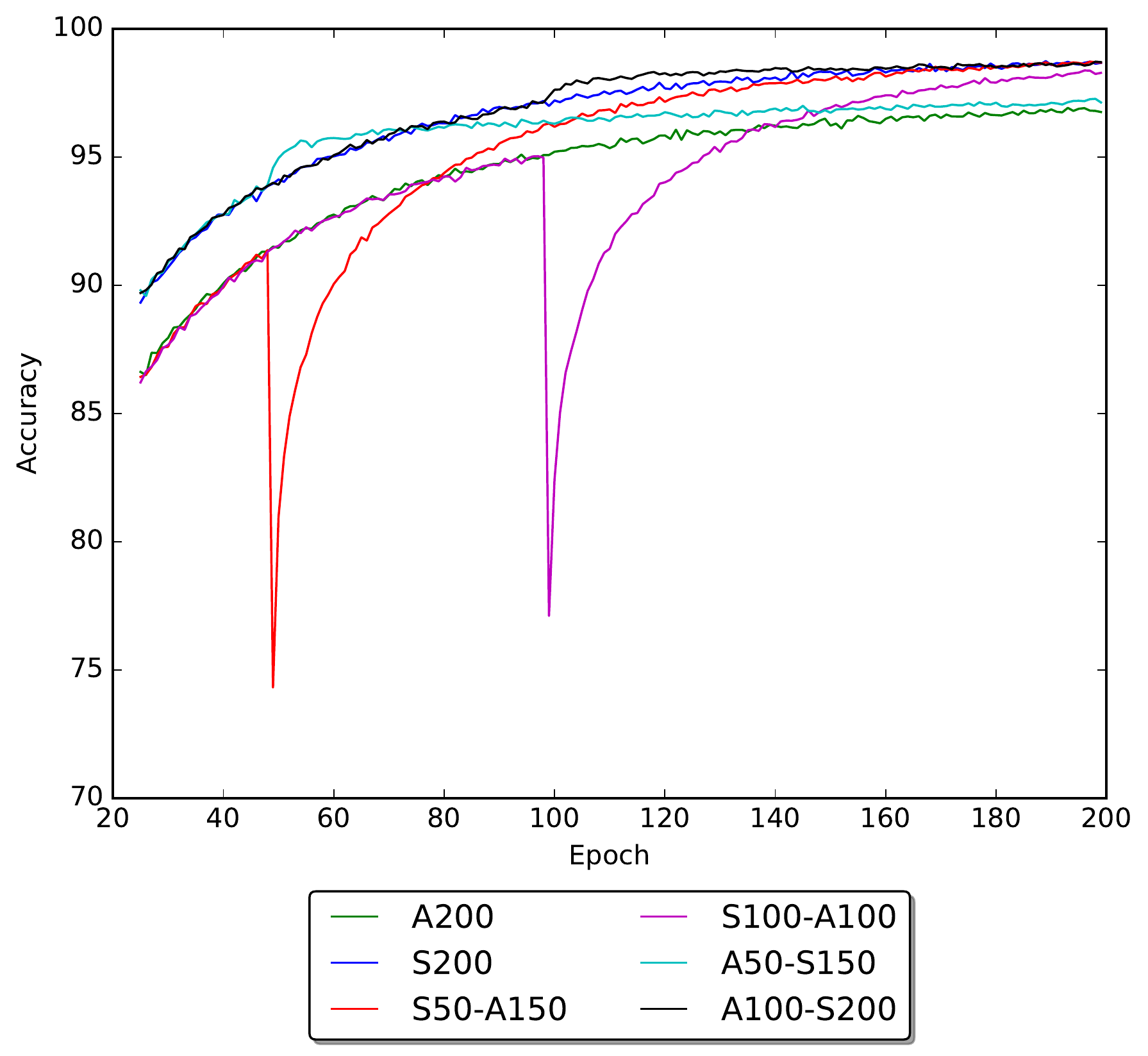}
        \includegraphics[width=0.33\linewidth]{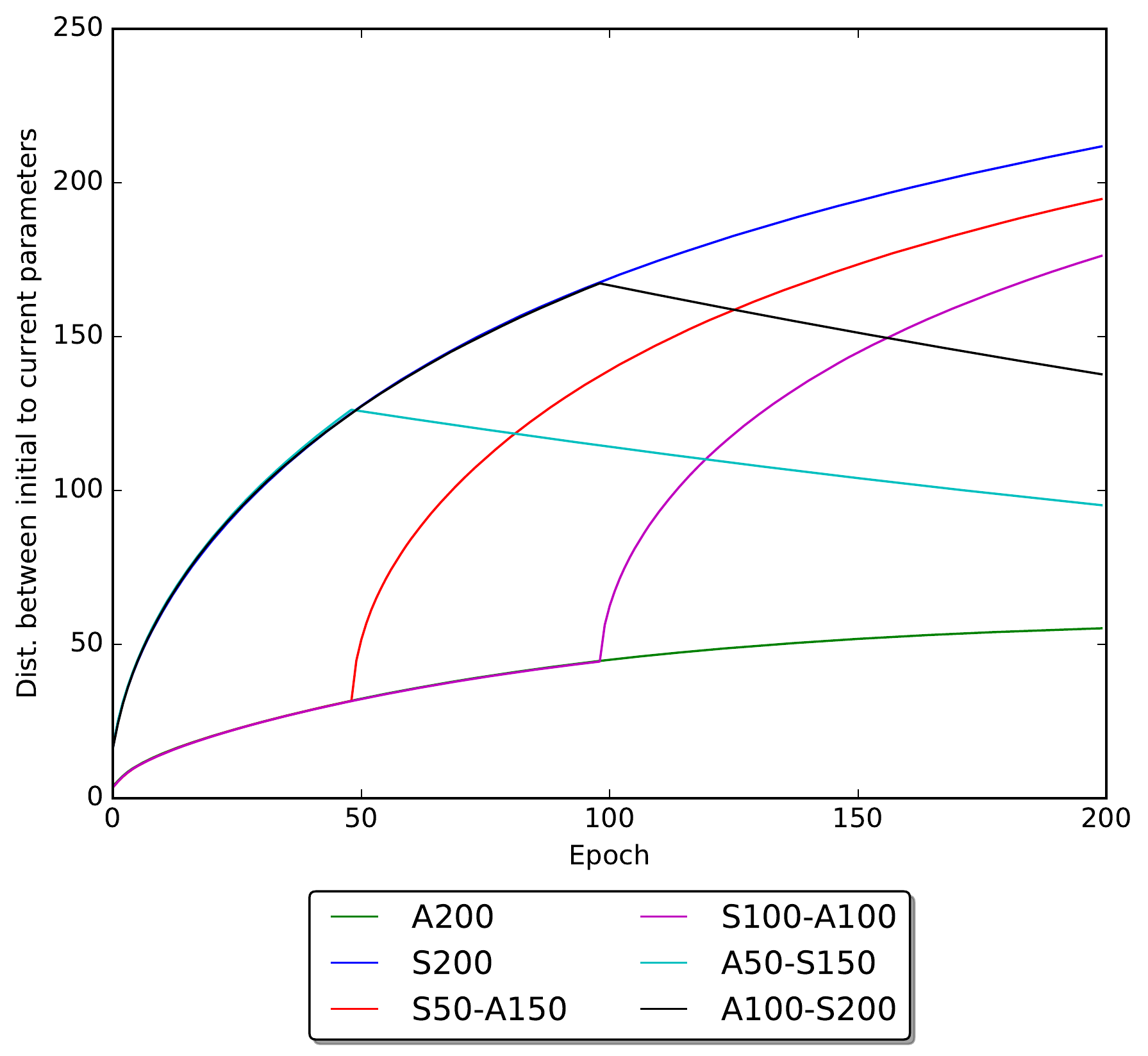}
        \includegraphics[width=0.32\linewidth]{vgg_v3_switch_sgdAdelta_inter_large.pdf}
        \mysubfigurecaption{Learning rate is set to 0.05 for SGD in the beginning,
        and then switched it to 0.01.}
        \label{fig:vgg_switch_adadelta_small}
    \end{minipage}
    \caption{VGG - Switching methods from SGD to Adadelta and Adadelta to SGD at
    epoch 50 and 100. Zoomed in version (Left). Distance between initial
    weights to weights at each epoch (Middle). The interpolation between
    different convergence parameters (Right).
    Each figure shows the results of switching methods at different 
    learning rate. 
    We label the switch of methods in terms of ratio. 
    For instance, S50-A50 as trained with SGD in the first 100 epoch 
    and swithced to Adadelta for the rest of the epoch.}
    \label{fig:vgg_switch_sgd_adadelta}
\end{figure}

%\pagebreak

%\subsection{Repeated Experiments}
%Here, we present more figures with different initializations 
%(different random seeds but using MSR initialization)
%on NIN and VGG models to show that we observe similar results 
%to our analysis.

\end{document}